\documentclass[pdflatex,sn-mathphys-num]{sn-jnl}% Math and Physical Sciences Numbered Reference Style 
\usepackage{graphicx}%
\usepackage{multirow}%
\usepackage{amsmath,amssymb,amsfonts}%
\usepackage{amsthm}%
\usepackage{mathrsfs}%
\usepackage[title]{appendix}%
\usepackage{xcolor}%
\usepackage{textcomp}%
\usepackage{manyfoot}%
\usepackage{booktabs}%
\usepackage{algorithm}%
\usepackage{algorithmicx}%
\usepackage{algpseudocode}%
\usepackage{listings}%
\usepackage{float}
\usepackage{booktabs}
\usepackage{makecell}
\usepackage{adjustbox}
\usepackage{caption} 
\usepackage{pifont}
\usepackage[normalem]{ulem}
\useunder{\uline}{\ul}{}
\usepackage{array}
\usepackage{colortbl}
\usepackage[bottom]{footmisc}
\usepackage{tablefootnote} % Include the tablefootnote package
\usepackage{tabularx}
\usepackage{makecell}
\usepackage{ltablex,multirow,makecell,caption}
\usepackage{lscape}

\usepackage{xcolor}  
\usepackage{mdframed} 

\definecolor{Reviewer1}{RGB}{173, 255, 168}  % Light green for Reviewer 1  
\definecolor{Reviewer2}{RGB}{255, 255, 102}      % Yellow for Reviewer 2 
\definecolor{Reviewer3}{RGB}{173, 216, 230}   % Light blue for Reviewer 3 

\theoremstyle{thmstyleone}
\theoremstyle{thmstyletwo}%

\theoremstyle{thmstylethree}%

\begin{document}

\title[Article Title]{An Evaluation of Large Language Models on Text Summarization Tasks Using Prompt Engineering Techniques.}

\author[1]{Walid Mohamed Aly}
\author[1]{Taysir Hassan A. Soliman}
\author[2]{Amr Mohamed AbdelAziz}

\affil[1]{Information Systems Department, Faculty of Computers and Information, Assiut University, Assiut, Egypt, 71515\\}

\affil[2]{Information Systems Department, Faculty of Computers and Artificial Intelligence, Beni-Suef University, Beni-Suef, Egypt, 62111\\}


\abstract{
Large Language Models (LLMs) continue to advance natural language processing with their ability to generate human-like text across a range of tasks. Despite the remarkable success of LLMs in Natural Language Processing (NLP), their performance in text summarization across various domains and datasets has not been comprehensively evaluated. At the same time, the ability to summarize text effectively without relying on extensive training data has become a crucial bottleneck. To address these issues, we present a systematic evaluation of six LLMs across four datasets: CNN/Daily Mail and NewsRoom (news), SAMSum (dialog), and ArXiv (scientific). By leveraging prompt engineering techniques including zero-shot and in-context learning, our study evaluates the performance using the ROUGE and BERTScore metrics. In addition, a detailed analysis of inference times is conducted to better understand the trade-off between summarization quality and computational efficiency. For Long documents, introduce a sentence-based chunking strategy that enables LLMs with shorter context windows to summarize extended inputs in multiple stages. The findings reveal that while LLMs perform competitively on news and dialog tasks, their performance on long scientific documents improves significantly when aided by chunking strategies. In addition, notable performance variations were observed based on model parameters, dataset properties, and prompt design. These results offer actionable insights into how different LLMs behave across task types, contributing to ongoing research in efficient, instruction-based NLP systems.
\footnote{This manuscript is an extended version of the work accepted for publication in the International Journal of Advanced Computer Science and Applications (IJACSA), Volume 16, Issue 6, June 2025.}  
}




\maketitle

\section{Introduction}\label{sec1}

In today's world, the rapid expansion of the internet has resulted in an exponential increase in the volume of data available to us like articles, books, documents, news,…, etc. However, most of the data is unstructured and redundant, and increasing every moment. As a result, there is a growing need to condense this vast amount of content and summarize this huge amount of data into shorthand and meaningful text to make it more accessible and easier to understand. By summarizing data, the time needed to process large data volumes can be significantly reduced while enabling us to quickly identify and extract the most pertinent information\citep{bib1}.

Automatic Text Summarization (\textbf{ATS}) \cite{bib3}  is a sub-domain of NLP, that deals with distilling key concepts and information from lengthy documents into concise summaries while preserving coherence and consistency. In NLP, the ability to condense text into a concise summary has become a major area of focus. Due to the fact that creating a summary that preserves the coherence and consistency of the document necessitates having a thorough understanding of both the input document’s structure and key details. In recent years, there has been a notable advancement in this field, which has been greatly enhanced by the active contributions of numerous researchers \cite{bib4}.

Efforts in text summarization have evolved from traditional heuristic-based methods \cite{bib5} to more sophisticated approaches to cover different types of text summarization including generic \cite{bib6}, domain-aware \cite{bib8}, multi-document \cite{bib10,bib9}, multimodal \cite{bib7}, extractive \cite{bib11,bib12} and abstractive text summarization \cite{bib13,bib14}. Various techniques are used in text summarization to enhance the produced summary including machine learning approaches \cite{bib15}, deep learning-based approaches\cite{bib16,bib17}. Recent breakthroughs in deep learning have improved NLP tasks including text summarization, especially with transformer methods \cite{bib18,bib19}. With the emergence of LLMs, NLP tasks including text summarization are about to have significant improvements and revolutionize the way in which machines understand and interact with human language.

LLMs \cite{bib22} represent a type of Artificial Intelligence (AI) that possesses significant power in understanding complex patterns and responding with well-reasoned contextually appropriate responses.
Researchers have demonstrated exceptional competence in a variety of NLP tasks, including machine translation \cite{bib39}, question answering \cite{bib40,bib75}, and text summarization \cite{bib11}.
LLMs \cite{bib20} have demonstrated unparalleled capabilities in understanding and generating human-like text, thereby holding tremendous promise for enhancing the quality and efficiency of summarization tasks.
Despite recent studies try to benchmark LLMs on text summarization tasks, there is a remarkable absence of comprehensive, large-scale, and multi-dimensional analysis focusing on benchmarking and analyzing the models on the different text summarization tasks. This leads to a significant gap in our understanding of these models’ capabilities.

Prompt Engineering is a sophisticated AI engineering methodology \cite{bib60}. It entails augmenting LLMs by giving them customized cues and modifying the source text to produce the intended result. Prompt engineering is crucial in LLMs as it quickly becomes essential for unlocking the full potential of such models. Also, prompt engineering utilizes prior knowledge and logical reasoning of the input to influence the outputs generated by the Models \cite{bib63}. Recently, Wide-ranging techniques have been developed as a result of recent notable advancements in the field of prompt engineering \cite{bib68}. The range of these developments includes basic techniques as well as more advanced strategies intended to manage challenging jobs. Furthermore, prompt engineering advances show a deeper comprehension of utilizing LLMs improving their functionality and usefulness in a range of applications and domains \cite{bib62}.

% To explore LLMs' capabilities for text summarization, six language models and four datasets are used in our thorough analysis. LLMs are tested using different distinct prompting techniques to assess their text-summarizing proficiency. We investigate the effects of different prompts, numbers of in-context demonstrations, and parameters on multiple LLMs.  

% We specifically address the following research inquiries:
% \begin{itemize}
%     \item	What is the general performance of LLMs in text summarization?
%     \item Do the models demonstrate proficiency across various types of text summarization tasks from different domains, parameters, and prompts?
%     \item How do models perform when they need to summarize long documents and scientific papers?
%     \item What is the impact of prompt engineering techniques, such as Zero-shot Learning (ZSL) and In-Context Learning (ICL), on the performance of LLMs in summarization tasks?
% \end{itemize}

To explore LLMs' capabilities for text summarization, In this study, we present a comprehensive evaluation of six LLMs, targeting a diverse range of domains including news articles, dialogues, and scientific publications. Our contributions can be summarized as follows:

\begin{itemize}
    \item An evaluation of six open-source LLMs across four benchmark datasets: CNN/DailyMail, NewsRoom, SAMSum, and ArXiv, providing insights into their cross-domain summarization abilities.

    \item A comparative assessment of prompting strategies, including Zero-Shot Learning (ZSL) and In-Context Learning (ICL), to evaluate the effect of prompt design, context length, and number of demonstrations on model performance.

    \item An empirical analysis of model scale and architecture, examining the impact of parameter size and design on summarization quality across different domains.

    \item A dedicated investigation of long-document summarization, introducing a chunking strategy to mitigate the limitations imposed by context window size, and evaluating its influence on summary coherence and quality.

    \item A detailed efficiency and cost analysis, measuring inference time for each model, highlighting trade-offs between performance and deployment feasibility.

    % \item A study of failure cases and limitations, particularly in scientific domains, to identify common weaknesses and inform future improvements in LLM-based summarization systems.

\end{itemize}

The remainder of this paper is organized as follows: Section \ref{sec2} introduces related works on text summarization and LLMs. Section \ref{sec3} describes in detail the experimental setup and methods, Section \ref{sec4} interprets the results, Section \ref{sec5} provides Discussion, Finally, Section \ref{sec6} provides a concise conclusion and future work.

\section{Related Work} \label{sec2}
% Lierature Review
The extractive and abstractive methods are two main popular approaches for text summarization \cite{bib5}. In the extractive approach, the models directly select or rearrange words from the source text to create the summary \cite{bib11}. In contrast, in abstractive summarization, the model utilizes a vocabulary spanning the corpus and generates new sentences to form the summary \cite{bib13}.

The Lead system is a highly reliable base for newspaper summarization, which involves extracting the first k sentences from documents to create a robust summary, particularly effective for newspaper articles \cite{bib23}. Researchers in \cite{bib24} presented a model to select sentences from the documents as a binary classification problem through a two-layer RNN, the first one works at the word level to get a representation of each word in terms of the position of the word in a sentence. The other one works at the sentence level to finally decide which sentence should be selected in the final summary and which is not. Authors in \cite{bib25} introduced an end-to-end framework based on a neural network for extracting sentences for the final summary. The framework employs a hierarchical encoder with two components: one for sentence representation and another for sentence extraction, which selects sentences based on model score and their importance relative to previously selected sentences. The great impact of the transformer architecture \cite{bib26} in NLP tasks was boosting the results in ATS by fine-tuning powerful pre-trained language models. Authors in \cite{bib27} performed a fine-tuning of pre-trained Bidirectional Encoder Representations from Transformers (BERT) \cite{bib28} model for extractive and abstractive text summarization to extract syntactic and semantic relationships from the text to form the summary. Authors in \cite{bib29} used a word-level attention mechanism that selects relevant content to be in the final summary. 

Researchers in \cite{bib30} introduced an abstractive text summarization technique improves abstract text summarization by utilizing a Transformer-based strategy with a self-attention mechanism. Their model attempts to enhance text comprehension by resolving coreference issues which will result in more accurate summaries. Another abstractive text summarization system proposed in \cite{bib31} based on a sequence-to-sequence deep learning model that combines the semantic-based data transformations and the encoder-decoder deep learning models through three components: the first was a model for text generalization, the second model which was used to generate the summary, and the third one marked the summary human-readable according to the salient information from the original document. BART (Bidirectional and Auto-Regressive Transformers) \cite{bib32} can be seen as one of the state-of-the-art effective models in text generation tasks. Fine-tuning BART model for abstractive text summarization achieves good results because of its huge capabilities in text generation and its ability to handle long documents and capture complex dependencies between sentences. Furthermore, researchers in \cite{bib33} used the hierarchical approach over word level and then sentence level to introduce an improved model over BART for text summarization. Another cutting-edge model for text summarization is PEGASUS \cite{bib34} which uses a pre-trained method to create summaries from gap sentences taken out of the original document \cite{bib35}. Using this technique the model can extract the structure and main ideas from the input text, which helps it generate summaries that are both coherent and informative.

LLMs have demonstrated notable effectiveness across a spectrum of tasks, though their precise capabilities and limitations remain somewhat unclear. We will explore recent studies that made progress in examining the performance of LLMs in text summarization tasks. \cite{bib36} introduce an exploration of text summarization with a diverse set of LLMs, including MPT-7b-instruct, falcon-7b-instruct, and OpenAI ChatGPT text davinci-003 models across two standard text summarization datasets CNN/Daily Mail \cite{bib3} and XSum\cite{bib37} (25 test sample of each dataset) and evaluate the results using BLUE\cite{bib38}, ROUGE and BERTScore\cite{bib39}. Results showed that text-davinci-003 outperformed other models on the two datasets. Authors in \cite{bib40} introduced a systematic study for assessing the performance of zero-shot prompting performance on medical evidence. They used two LLMs in their study GPT-3.5\cite{bib42} and ChatGPT \cite{bib41}. Results were evaluated using multiple evaluation metrics ROUGE \cite{bib2}, METEOR \cite{bib43}, and BLUE \cite{bib38}. In addition, the authors performed an extensive human evaluation of the model-generated summaries to attain a thorough comprehension of the summarization capabilities exhibited by LLMs. The authors in \cite{bib44} applied an adaptation of eight open-source and proprietary LLMs for four distinct summarization tasks comprising six datasets in the clinical domain. The data used in the evaluation are radiology reports, patient questions, progress notes, and dialogue. The study includes multiple automatic and human evaluations of the results across the different datasets.

Researchers in \cite{bib45} proposed to use In-Context Learning \cite{bib46,bib47} with LLMs in performing multidimensional evaluation including relevance, factual consistency, coherence, and fluency, then study the effectiveness of this technique to evaluate summaries generated by GPT-3, and concluded that the proposed evaluation aligned well with human judgments when evaluating summaries written by GPT-3.Researchers in \cite{bib48} performed benchmarking evaluation of the news summarization dataset by performing a comprehensive human evaluation of ten LLMs across two news datasets \cite{bib3,bib37}. They found that the state-of-the-art LLM performed comparably to with summaries written by freelance writers, with instruction tuning being the key factor for success. The emergence of LLMs with text summarization resulted in generating more coherent and factual consistency summaries.

\section{Methodology and Experimental Setup}\label{sec3}

In this section, we provide the details of the datasets, tasks, prompting techniques, and language models used in our research study. We begin by describing the characteristics of the datasets utilized as text summarization tasks. Following this, the prompting techniques employed are elaborated upon, showcasing their role in guiding model responses. Lastly, we delve into the selection and configurations of the used LLMs, emphasizing their capabilities and the rationale behind their use.

\subsection{Datasets}
In this work, the following datasets are used to serve as the fundamental basis for our evaluation and comparative analysis of summaries produced by LLMs. 

Table \textbf{\ref{tab:table1}} provides the statistics of the used datasets: 

\subsubsection{CNN/DailyMail } 
CNN/DailyMail\cite{bib3} is a benchmark dataset for training and evaluating text summarization models. It contains just over 300k unique news articles written by journalists at CNN and the Daily Mail. Both extractive and abstractive summarization are supported by the dataset’s current version. However, machine reading comprehension and abstractive question answering were the main goals of the original version.
\subsubsection{CORNELL NEWSROOM }
CORNELL NEWSROOM dataset \cite{bib67} contains summaries from 38 major news sources that were compiled by contributors from search and social media metadata between 1998 and 2017. These summaries present a wide range of summarization strategies combining aspects of extractive and abstractive methods.

\subsubsection{SAMSum Corpus }
SAMSum \cite{bib65}  is a dataset of abstractive dialogue summaries and contains over 16k chat conversations with summaries that have been manually annotated. In the increasingly fascinating field of summarization research the authors of the dataset examine the problem of abstractive dialogue summarization. The increasing ubiquity of online discussions made possible by apps like WeChat WhatsApp and Messenger lends significance to this direction. In dialogue summarization models capture the evolving nature of digital communication by distilling talks between several participants into brief summaries.

\subsubsection{ArXiv dataset }
ArXiv dataset \cite{bib66} provides access to a selection of research papers from the ArXiv preprint server, which hosts articles from a variety of fields including biology physics computer science and mathematics and more. Usually, the dataset comprises research paper full texts accompanied by corresponding human-generated summaries. This dataset provides a challenge for text summarization as the research articles present long documents compared to news articles or dialogues. So, the performance of the LLMs in summarizing these articles are explored in detail.

In our work, 2K random samples used from the test set of each dataset except the SAMSum dataset which we used all test samples in our analysis and used samples from the training set for the few-shot prompting method.

% Redefine \thetable to use Roman numerals
%\renewcommand{\thetable}{\Roman{table}}

\begin{table}[h]

    \centering
    \begin{adjustbox}{max width=\textwidth}

        \begin{tabular}{|c|c|c|c|c|c|c|}
        \hline
        \multirow{2}{*}{\textbf{Dataset}}& 
        \multirow{2}{*}{\textbf{Domain}} &
    
        \multicolumn{3}{|c|}{\textbf{No. of Documents}}&
        
       \multirow{2}{2cm}{\textbf{\makecell{Docs Len\\(\#Words)}}} & 
       \multirow{2}{2cm}{\textbf{\makecell{Sum Len\\(\#Words)}}} \\
         & & Train & Valid & Test & &  \\
         \hline
        \textbf{CNN/DailyMail} & 
        \multirow{2}{*}{\textbf{News Summarization}} &  
        287K & 13K & \textbf{11K} & 
        656 & 
        52 \\
        \textbf{NEWSROOM} & 
         & 
        995K & 108K & \textbf{108K} & 
        658 & 
        26 \\
        \hline
        \textbf{SAMSum} & \textbf{Dialogue Summarization} & 
        14K & 818 & \textbf{819} & 
         - & -
         \\
        \hline
         \textbf{ArXiv} & \textbf{Scientific Paper Summarization} & 
        203K & 6K & \textbf{6K} & 
        4938 & 
        220 \\
        \hline
        \end{tabular}  
        \end{adjustbox}
    \caption{\textbf{Statistics of the datasets.}}
        \label{tab:table1}    
\end{table}

\subsection{Prompt Engineering Techniques}

As demonstrated in Section \ref{sec1}, Prompt Engineering encompasses various techniques utilized to elicit desired outputs from LLMs. In the scope of our study, two primary methods are employed: zero-shot learning and in-context learning. By examining these techniques, our study aims to explore the effectiveness and nuances of each approach in improving the performance and reliability of LLMs in text summarization across the used models and datasets. 

\subsubsection{Zero-shot Learning (ZSL) \cite{bib69}}

ZSL represents the fundamental form of prompting, wherein a model is given a prompt or directive without any explicit examples or training data regarding the task. The model is then required to produce a response based only on the prompt that was supplied and its own internal knowledge base. ZSL format is expressed as P = f prompt (TD; xtest), where TD denotes the task description, x test signifies the test example, and f prompt represents a function that converts the data into a natural language prompt.

\subsubsection{In-context Learning (ICL) \cite{bib46}}

ICL gained popularity by using language models to learn tasks given only a few examples. Through ICL, LLMs are provided with a prompt containing a series of input-output pairs to accomplish a specific task. At the end of the prompt, a test input is appended and allows LM to make a prediction just by conditioning the prompt and predicting the next tokens. Recent works have attempted to explain ICL \cite{bib45} and drawn connections between ICL mechanisms and other standard learning algorithms such as regression and gradient descent \cite{bib49,bib50}.

These prompt engineering techniques are applied in our study to take advantage of the versatility and generalization abilities of LLMs across various datasets in various domains. ZSL is employed in our work since it is a basic method in prompt engineering \cite{bib68} and gained popularity when improved by instruction-tuning over language models \cite{bib70}. On the other hand, ICL, which uses a few examples within the prompt, enables quick adaptation to specific tasks such as text summarization. By using these methods, our research objectives to evaluate LLMs' performance in text summarization over different datasets are achieved.

\subsection{Large Language Models}
 
In this work, LLMs are investigated and assessed for different text summarization tasks. LLMs are categorized according to architecture into two categories: sequence-to-sequence and autoregressive models. The sequence-to-sequence models were introduced in the original transformer architecture \cite{bib26}. They implement a relatively direct method of embedding an entire sequence into a higher-dimensional space, followed by decoding by a decoder. These were primarily designed for translation tasks \cite{bib53}, due to their excellence at mapping sequences between languages and later adapted into text summarization tasks \cite{bib54}. The other architecture is the Autoregressive Model, which have popularity for unsupervised learning and are popularized by GPT (Generative Pre-trained Transformer) \cite{bib55}. Autoregressive models predict subsequent tokens by sequentially leveraging preceding tokens. They employ probabilistic inference to generate text, heavily relying on the decoder component of the transformer. Unlike sequence-to-sequence models, autoregressive models do not require a predefined input sequence and are adapted to text generation tasks.

The models used in our study include three variants from the Llama-2 family \cite{bib52} specifically  Llama-2-7B-chat, Llama-2-13B-chat, and Llama-2-70B-chat, selected to analyze how model scale impacts summarization performance across domains. Llama-2’s open-source design ensures reproducibility, and its decoder-only architecture is widely adopted in NLP research \cite{bib51}, making it a robust baseline for benchmarking. Mistral-7B-Instruct-v0.1 \cite{bib56}, a 7B-parameter model released under the Apache 2.0 license, provides an instruction-tuned alternative for efficiency-focused comparisons. Gemma-7B-it \cite{bib58}, like the Llama-2 family and Mistral-7B , employs a decoder-only transformer architecture and shares a comparable parameter count (7B). Gemma also share with Mixtral-8x7B-Instruct-v0.1 as instruction-tuning framework for refined comparison in that framework.

Mixtral-8x7B-Instruct-v0.1 , a fine-tuned variant of Mixtral-8x7B, employs eight specialized sub-networks that activate dynamically per task, optimizing computational efficiency while maintaining high performance\cite{bib57}. Its sparse architecture reduces inference costs by leveraging only relevant experts, avoiding full-network activation\cite{bib78}. Mixtral-8x7B-Instruct-v0.1 has 47B parameters and a 32k-token context window. Its powerful architecture is essential for processing lengthy inputs and maintaining coherence.

The inclusion of multiple 7B-parameters models and  enables a controlled comparison of architectural and training variations within a fixed parameter scale. Meanwhile, models like Llama-2-13b-chat, Mixtral-8x7B-Instruct-v0.1 (47B parameters, sparse MoE), and Llama-2-70B-chat  provide cross-scale comparisons, evaluating performance trends across model sizes and configurations. This dual approach facilitates systematic analysis of how architectural innovations (e.g., MoE), parameter scaling, and instruction-tuning impact summarization quality, scalability, and computational efficiency.

Proprietary models were excluded to ensure reproducibility and mitigate potential biases from opaque training data or closed-source updates. Table \ref{tab:table2} provides a comparison of the used models.
All experiments were conducted on two NVIDIA A100 GPUs to ensure computational robustness with mixed-precision floating-point (fp16) support, model parallelism, and quantization to reduce memory footprint.
% Quantization-aware optimizations and model parallelism were applied to reduce memory usage, allowing deployment of like Llama-2-70B-chat and Mixtral-8x7B effectively. 

% Redefine \thetable to use Roman numerals
%\renewcommand{\thetable}{\Roman{table}}
\begin{table}[h]
    \centering
    \begin{adjustbox}{max width=\textwidth}  
    \begin{tabular}{|c|c|c|}
        \hline
        \textbf{Model Name} & \textbf{\# Parameters}  & \textbf{Context-Length}  \\
        \hline
        Llama-2-7b-chat & 7B & 4K \\
        \hline
        gemma-7b-it & 7B  & 8K  \\
        \hline
        Mistral-7B-Instruct-v0.1 & 7B  & 8K  \\
        \hline
        Llama-2-13b-chat & 13B  & 4K  \\
        \hline

        Mixtral-8x7B-Instruct-v0.1 & 47B  & 32K \\
        \hline
        Llama-2-70b-chat & 70B  & 4K  \\
        \hline
        
    \end{tabular}
    \end{adjustbox}
    \caption{Characteristics of Different LLMs Utilized in our work.}
    \label{tab:table2}
\end{table}

\subsection{Long Document Processing via Chunking Strategy}
\label{sec:chunking}

Summarizing long documents, such as scientific articles from the ArXiv dataset, is challenging due to the inherent context length limitations of LLMs. To work within these constraints, our initial experiments employ a preprocessing step that trims each article to fit the models' maximum allowable context. This approach is designed to strike a balance between adhering to context length restrictions and preserving the essential details necessary for an accurate summary. However, we recognize that this trimming process may lead to the loss of some contextual nuances and finer details.

Although newer models with extended context windows offer a promising solution, they are not always accessible or computationally efficient for all users. To address these limitations, we employ chunking strategy, which involve dividing documents into semantically coherent units (e.g., paragraphs, pages, or even using fixed length chunks), summarize each independently, and iteratively refine the results to maintain global coherence \cite{bib59}. To create more meaningful chunks, we utilized the Natural Language Toolkit (NLTK)\footnote{\url{https://www.nltk.org/}} to segment each document into sentences. This method ensures that each chunk retains its semantic integrity, which is crucial for preserving key details during summarization. Independent summaries for each chunk are generated. These intermediate summaries are subsequently combined and reprocessed by the model, yielding a unified summary that effectively encapsulates the essential information from each segment.

\subsection{Evaluation Metrics}

Using robust metrics that accurately reflect the quality of the generated summaries is essential when assessing the performance of text summarization models. The metrics for this study are divided into two main categories: \textbf{Word Overlap } and \textbf{Semantic Similarity Metrics}. A thorough assessment of the quality of the generated summaries is provided by employing both categories of metrics to capture the lexical and semantic fidelity of the summaries. Afterwards each category is given a detailed explanation.

\subsubsection{Word Overlap Metrics:}
These metrics evaluate how similar the text generated by the model to the reference summary by comparing the overlap of their tokens (words or subworlds). In our study, ROUGE \cite{bib2} is applied for word overlap metrics

    \textbf{ROUGE} (Recall-Oriented Understudy for Gisting Evaluation) \cite{bib2}: is a set of metrics used to automatically assess the quality of summaries generated by machines. ROUGE-N, and ROUGE-L, are used to compare generated summaries to gold standard summaries of each dataset for quality assessment.

    \textbf{ROUGE-N} counts the number of n-grams (sequences of n words) that overlap between the machine-generated summary and the gold standard summary. Higher ROUGE-N scores mean the output summary of the model shares more common n-grams with the human-written one, suggesting better content overlap.

    \textbf{ROUGE-L} focuses on the Longest Common Subsequence (LCS) of words between the two summaries. LCS is the longest sequence of words that appears in the same order in both summaries. A high ROUGE-L score means the output summary of the model captures the overall flow of ideas and sentence structure like the gold standard.

\subsubsection{Semantic Similarity Metrics}
By comparing the model's output to the reference summary, semantic similarity metrics assess how closely the generated summary adheres to the original text's meaning and content. BERT Score \cite{bib39} is a metric used to assess the text quality produced by NLP models in a variety of contexts including text summarization. Using contextual embeddings derived from a BERT\cite{bib28} model, BERTScore calculates the degree of similarity between the generated text and a reference text. This metric addresses common challenges encountered by n-gram-based metrics. n-gram-based metrics might misidentify paraphrases because of differences between semantically correct expressions and the reference texts surface form which could distort performance assessments. In contrast, BERTScore utilize contextualized token embeddings which are used to compute similarity accurately. Moreover, BERTScore avoids penalties for significant semantic rearrangements and outperforms n-gram models in capturing long-range dependencies.

\subsection{Prompt building and model parameters}

In this section the prompt building and model parameters utilized in the experiments are detailed, including the specific prompts employed for each dataset in ZSL, illustrating how the model was instructed to perform the summarization task. Additionally, the structure of the ICL prompts, which incorporate a few example summaries to guide the model's output generation, is described. Through these experiments, the effectiveness of different prompting techniques and model configurations in producing high-quality text summaries is evaluated.

A prompt refers to user-provided text input that directs models to produce desired results. Prompt design and selection have become increasingly important for optimizing LLMs \cite{bib60}. In addition, adapting prompts to LLMs is crucial because it directly affects the quality and relevance of the model's outputs \cite{bib62,bib72}. Furthermore, the way a prompt is structured, and words included can significantly influence the model's interpretation and response \cite{bib73}.

Temperature is one of the crucial hyperparameters in LLMs' settings since it controls the randomness of the model's output \cite{bib64}. A lower temperature value (closer to 0) makes the model more deterministic, leading to more focused and predictable responses by favoring higher probability words. Conversely, a higher temperature value increases randomness, enabling a greater variety of words to be explored by the model leading to more imaginative or varied results.

In our study, temperature values (0.1, 0.5, 0.9) were systematically tested to evaluate their impact on summary quality. Since lower temperatures (e.g., 0.1) reduce output randomness by favoring high-probability tokens, prioritizing factual accuracy. This make the summary aligned with reference texts—critical for summarization tasks where precision is paramount. Conversely, higher temperatures (e.g., 0.9) increase diversity but risk introducing hallucinations or incoherent phrasing\cite{bib64}.

Our analysis of temperature settings Table \ref{tab:table3} reveals that temperature 0.1 consistently maximizes ROUGE metrics across datasets, particularly ROUGE-L, which reflects structural alignment with reference summaries. Specifically, lower temperature settings constrain the randomness in token sampling, which is advantageous for summarization tasks that prioritize factual accuracy and consistency.
By contrast, higher temperatures (0.5 or 0.9) sometimes introduced more creative outputs but at the risk of producing less relevant or hallucinated terms. Hence, the temperature value is set to be 0.1 for the final evaluations. By doing this we target generating high-quality and accurate summaries that focus on the input documents and closely match the reference texts.

\begin{table}[ht]
\centering
\begin{adjustbox}{max width=\textwidth}  

%\resizebox{\columnwidth}{!}{%
%\renewcommand{\arraystretch}{1.5}
\begin{tabular}{|
>{\columncolor[HTML]{FFFFFF}}c 
>{\columncolor[HTML]{FFFFFF}}c |
>{\columncolor[HTML]{FFFFFF}}c 
>{\columncolor[HTML]{FFFFFF}}c 
>{\columncolor[HTML]{FFFFFF}}c 
>{\columncolor[HTML]{FFFFFF}}c 
>{\columncolor[HTML]{FFFFFF}}c 
>{\columncolor[HTML]{FFFFFF}}c 
>{\columncolor[HTML]{FFFFFF}}c 
>{\columncolor[HTML]{FFFFFF}}c 
>{\columncolor[HTML]{FFFFFF}}c 
>{\columncolor[HTML]{FFFFFF}}c 
>{\columncolor[HTML]{FFFFFF}}c 
>{\columncolor[HTML]{FFFFFF}}c |}
\hline
 &
   &
  \multicolumn{12}{c|}{\cellcolor[HTML]{FFFFFF}\textbf{Datasets}} \\ 
 &
   &
  \multicolumn{3}{c|}{\cellcolor[HTML]{FFFFFF}\textbf{CNNDM}} &
  \multicolumn{3}{c|}{\cellcolor[HTML]{FFFFFF}\textbf{NewsRoom}} &
  \multicolumn{3}{c|}{\cellcolor[HTML]{FFFFFF}\textbf{SAMSum}} &
  \multicolumn{3}{c|}{\cellcolor[HTML]{FFFFFF}\textbf{ArXiv}} \\ \hline
\begin{tabular}
[c]{@{}c@{}}Temp\\  Value\end{tabular} &
  Metrics &
  \multicolumn{1}{c|}{\cellcolor[HTML]{FFFFFF}\begin{tabular}[c]{@{}c@{}}P/\\    R1\end{tabular}} &
  \multicolumn{1}{c|}{\cellcolor[HTML]{FFFFFF}\begin{tabular}[c]{@{}c@{}}R/\\     R2\end{tabular}} &
  \multicolumn{1}{c|}{\cellcolor[HTML]{FFFFFF}\begin{tabular}[c]{@{}c@{}}F1/\\     RL\end{tabular}} &
  \multicolumn{1}{c|}{\cellcolor[HTML]{FFFFFF}\begin{tabular}[c]{@{}c@{}}P/\\     R1\end{tabular}} &
  \multicolumn{1}{c|}{\cellcolor[HTML]{FFFFFF}\begin{tabular}[c]{@{}c@{}}R/\\     R2\end{tabular}} &
  \multicolumn{1}{c|}{\cellcolor[HTML]{FFFFFF}\begin{tabular}[c]{@{}c@{}}F1/\\     RL\end{tabular}} &
  \multicolumn{1}{c|}{\cellcolor[HTML]{FFFFFF}\begin{tabular}[c]{@{}c@{}}P/\\    R1\end{tabular}} &
  \multicolumn{1}{c|}{\cellcolor[HTML]{FFFFFF}\begin{tabular}[c]{@{}c@{}}R/\\     R2\end{tabular}} &
  \multicolumn{1}{c|}{\cellcolor[HTML]{FFFFFF}\begin{tabular}[c]{@{}c@{}}F1/\\     RL\end{tabular}} &
  \multicolumn{1}{c|}{\cellcolor[HTML]{FFFFFF}\begin{tabular}[c]{@{}c@{}}P/\\     R1\end{tabular}} &
  \multicolumn{1}{c|}{\cellcolor[HTML]{FFFFFF}\begin{tabular}[c]{@{}c@{}}R/\\     R2\end{tabular}} &
  \begin{tabular}[c]{@{}c@{}}F1/\\     RL\end{tabular} \\ \hline
\cellcolor[HTML]{FFFFFF} &
  BERT &
  \multicolumn{1}{c|}{\cellcolor[HTML]{FFFFFF}\textbf{88.76}} &
  \multicolumn{1}{c|}{\cellcolor[HTML]{FFFFFF}86.13} &
  \multicolumn{1}{c|}{\cellcolor[HTML]{FFFFFF}87.41} &
  \multicolumn{1}{c|}{\cellcolor[HTML]{FFFFFF}86.99} &
  \multicolumn{1}{c|}{\cellcolor[HTML]{FFFFFF}\textbf{87.67}} &
  \multicolumn{1}{c|}{\cellcolor[HTML]{FFFFFF}\textbf{87.31}} &
  \multicolumn{1}{c|}{\cellcolor[HTML]{FFFFFF}89.31} &
  \multicolumn{1}{c|}{\cellcolor[HTML]{FFFFFF}\textbf{91.48}} &
  \multicolumn{1}{c|}{\cellcolor[HTML]{FFFFFF}\textbf{90.37}} &
  \multicolumn{1}{c|}{\cellcolor[HTML]{FFFFFF}\textbf{86.86}} &
  \multicolumn{1}{c|}{\cellcolor[HTML]{FFFFFF}80.09} &
  83.33 \\ %\cline{2-14} 
\multirow{-2}{*}{\cellcolor[HTML]{FFFFFF}0.1} &
  Rouge &
  \multicolumn{1}{c|}{\cellcolor[HTML]{FFFFFF}37.44} &
  \multicolumn{1}{c|}{\cellcolor[HTML]{FFFFFF}\textbf{16.42}} &
  \multicolumn{1}{c|}{\cellcolor[HTML]{FFFFFF}\textbf{24.53}} &
  \multicolumn{1}{c|}{\cellcolor[HTML]{FFFFFF}\textbf{25.38}} &
  \multicolumn{1}{c|}{\cellcolor[HTML]{FFFFFF}\textbf{8.82}} &
  \multicolumn{1}{c|}{\cellcolor[HTML]{FFFFFF}\textbf{19.96}} &
  \multicolumn{1}{c|}{\cellcolor[HTML]{FFFFFF}\textbf{39.35}} &
  \multicolumn{1}{c|}{\cellcolor[HTML]{FFFFFF}\textbf{14.61}} &
  \multicolumn{1}{c|}{\cellcolor[HTML]{FFFFFF}\textbf{30.7}} &
  \multicolumn{1}{c|}{\cellcolor[HTML]{FFFFFF}\textbf{49.74}} &
  \multicolumn{1}{c|}{\cellcolor[HTML]{FFFFFF}\textbf{32.47}} &
  \textbf{33.98} \\ \hline
\cellcolor[HTML]{FFFFFF} &
  BERT &
  \multicolumn{1}{c|}{\cellcolor[HTML]{FFFFFF}86.35} &
  \multicolumn{1}{c|}{\cellcolor[HTML]{FFFFFF}\textbf{88.03}} &
  \multicolumn{1}{c|}{\cellcolor[HTML]{FFFFFF}87.16} &
  \multicolumn{1}{c|}{\cellcolor[HTML]{FFFFFF}\textbf{87.11}} &
  \multicolumn{1}{c|}{\cellcolor[HTML]{FFFFFF}87.30} &
  \multicolumn{1}{c|}{\cellcolor[HTML]{FFFFFF}87.28} &
  \multicolumn{1}{c|}{\cellcolor[HTML]{FFFFFF}\textbf{90.54}} &
  \multicolumn{1}{c|}{\cellcolor[HTML]{FFFFFF}88.76} &
  \multicolumn{1}{c|}{\cellcolor[HTML]{FFFFFF}89.62} &
  \multicolumn{1}{c|}{\cellcolor[HTML]{FFFFFF}85.67} &
  \multicolumn{1}{c|}{\cellcolor[HTML]{FFFFFF}81.85} &
  83.70 \\ %\cline{2-14} 
\multirow{-2}{*}{\cellcolor[HTML]{FFFFFF}0.5} &
  Rouge &
  \multicolumn{1}{c|}{\cellcolor[HTML]{FFFFFF}\textbf{38.38}} &
  \multicolumn{1}{c|}{\cellcolor[HTML]{FFFFFF}14.73} &
  \multicolumn{1}{c|}{\cellcolor[HTML]{FFFFFF}24.49} &
  \multicolumn{1}{c|}{\cellcolor[HTML]{FFFFFF}24.42} &
  \multicolumn{1}{c|}{\cellcolor[HTML]{FFFFFF}8.19} &
  \multicolumn{1}{c|}{\cellcolor[HTML]{FFFFFF}18.41} &
  \multicolumn{1}{c|}{\cellcolor[HTML]{FFFFFF}39.08} &
  \multicolumn{1}{c|}{\cellcolor[HTML]{FFFFFF}14.13} &
  \multicolumn{1}{c|}{\cellcolor[HTML]{FFFFFF}30.25} &
  \multicolumn{1}{c|}{\cellcolor[HTML]{FFFFFF}40.20} &
  \multicolumn{1}{c|}{\cellcolor[HTML]{FFFFFF}12.92} &
  22.09 \\ \hline
\cellcolor[HTML]{FFFFFF} &
  BERT &
  \multicolumn{1}{c|}{\cellcolor[HTML]{FFFFFF}88.40} &
  \multicolumn{1}{c|}{\cellcolor[HTML]{FFFFFF}86.13} &
  \multicolumn{1}{c|}{\cellcolor[HTML]{FFFFFF}87.24} &
  \multicolumn{1}{c|}{\cellcolor[HTML]{FFFFFF}87.02} &
  \multicolumn{1}{c|}{\cellcolor[HTML]{FFFFFF}87.40} &
  \multicolumn{1}{c|}{\cellcolor[HTML]{FFFFFF}87.19} &
  \multicolumn{1}{c|}{\cellcolor[HTML]{FFFFFF}90.30} &
  \multicolumn{1}{c|}{\cellcolor[HTML]{FFFFFF}88.51} &
  \multicolumn{1}{c|}{\cellcolor[HTML]{FFFFFF}89.38} &
  \multicolumn{1}{c|}{\cellcolor[HTML]{FFFFFF}85.67} &
  \multicolumn{1}{c|}{\cellcolor[HTML]{FFFFFF}\textbf{82.03}} &
  \textbf{83.80} \\ %\cline{2-14} 
\multirow{-2}{*}{\cellcolor[HTML]{FFFFFF}0.9} &
  Rouge &
  \multicolumn{1}{c|}{\cellcolor[HTML]{FFFFFF}36.87} &
  \multicolumn{1}{c|}{\cellcolor[HTML]{FFFFFF}13.03} &
  \multicolumn{1}{c|}{\cellcolor[HTML]{FFFFFF}23.64} &
  \multicolumn{1}{c|}{\cellcolor[HTML]{FFFFFF}24.28} &
  \multicolumn{1}{c|}{\cellcolor[HTML]{FFFFFF}80.01} &
  \multicolumn{1}{c|}{\cellcolor[HTML]{FFFFFF}18.91} &
  \multicolumn{1}{c|}{\cellcolor[HTML]{FFFFFF}39.34} &
  \multicolumn{1}{c|}{\cellcolor[HTML]{FFFFFF}14.36} &
  \multicolumn{1}{c|}{\cellcolor[HTML]{FFFFFF}30.53} &
  \multicolumn{1}{c|}{\cellcolor[HTML]{FFFFFF}39.29} &
  \multicolumn{1}{c|}{\cellcolor[HTML]{FFFFFF}12.19} &
  21.61 \\ \hline
  
\end{tabular}%
\end{adjustbox}
%}%
\caption{Performance at different temperature values.}
\label{tab:table3}
\end{table}

To leverage the potential of LLMs with ZSL, the different models are guided using a variety of prompts, aiming to assess the results across different prompts to fully utilize the models' capabilities. The prompts are customized based on the diversity of each dataset to provide summaries and assess the output of each model against the reference summaries. The prompts for zero-shot prompting are available in Table \ref{tab:table4}.

\begin{figure}[ht]
    \centering
    \includegraphics[width=1\textwidth]{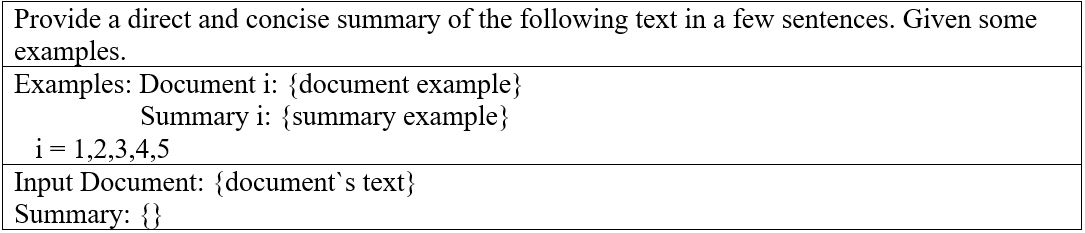}
    \caption{Structure of In-context Learning's Prompt.}
    \label{fig:fig1}
\end{figure}

For working with In-Context learning, the structure of the prompt is shown in Figure \ref{fig:fig1}. Including demonstrations of the prompt in in-context learning is crucial but overwhelming models with excessive examples could generate undesired output \cite{bib46,bib74}. Hence one, three, five and seven in-context demonstrations are used to guide LLMs in our experiments. This approach aims to strike a balance between providing sufficient guidance for LLMs to learn from the context of diverse examples without hindering their ability to generalize effectively.

% Redefine \thetable to use Roman numerals
%\renewcommand{\thetable}{\Roman{table}}

\begin{table}[ht]
    \centering
    \begin{adjustbox}{max width=\textwidth} 
    \renewcommand{\arraystretch}{1.25}
    \begin{tabular}{|m{2cm}|m{12cm}|}
        \hline
        \textbf{Prompt ID} & \textbf{Prompt Text} \\
        \hline
        \multicolumn{2}{|c|}{\textbf{CNN/DM Prompts}} \\
        \hline
        Prompt\#1 & Write a concise and comprehensive summary of this news article. \\
        \hline
        Prompt\#2 & Provide an abstract of this news article in a direct and concise summary. \\
        \hline
        Prompt\#3 & I want you to act as a text summarizer to help me create a concise summary of the provided text. The summary can be up to 3 sentences in length, expressing the key points and concepts written in the original text without adding your interpretations. \\
        \hline
        Prompt\#4 & I want you to act as a text summarizer to help me create a concise summary of the provided text. The summary can be up to 2 sentences in length, expressing the key points and concepts written in the original text without adding your interpretations. \\
        \hline
        Prompt\#5 & Please summarize the following news article in an informative extractive summary with two sentences. \\
        \hline
        \multicolumn{2}{|c|}{\textbf{NewsRoom Prompts}} \\
        \hline
        Prompt\#1 & Summarize this news article in one sentence. \\
        \hline
        Prompt\#2 & Write a concise and comprehensive summary of this news article in one sentence. \\
        \hline
        Prompt\#3 & I want you to act as a text summarizer to help me create a concise summary of the following article. \\
        \hline
        Prompt\#4 & Provide an abstract of this news article in a direct and concise summary in a few words. \\
        \hline
        Prompt\#5 & You are a helpful assistant. Please summarize the following text in one sentence. \\
        \hline
        \multicolumn{2}{|c|}{\textbf{SAMSum Prompts}} \\
        \hline
        Prompt\#1 & Summarize the following dialogue. \\
        \hline
        Prompt\#2 & Summarize the following dialogue in a few words. \\
        \hline
        Prompt\#3 & Summarize the following dialogue into an abstractive summary without any explanation. \\
        \hline
        Prompt\#4 & In short, what’s going on in this conversation? \\
        \hline
        Prompt\#5 & Summarize the following conversation. \\
        \hline
        Prompt\#6 & Summarize this conversation in one or two sentences. \\
        \hline
        \multicolumn{2}{|c|}{\textbf{ArXiv Prompts}} \\
        \hline
        Prompt\#1 & Provide an abstract of the following research article. \\
        \hline
        Prompt\#2 & Summarize the main points and findings of the scientific paper. \\
        \hline
        Prompt\#3 & I want you to act as a research paper summarizer to get an abstract for this research paper. \\
        \hline
        Prompt\#4 & Given this research article, create a TLDR to be used as a formal abstract for this paper. \\
        \hline
    \end{tabular}
    \end{adjustbox}
    \caption{Examples of different prompts provided to the LLMs .}
    \label{tab:table4}
    
\end{table}

\section{Results}\label{sec4}
This section presents an overview of the performance of different LLMs with an emphasis on how well they can produce accurate and coherent summaries. To guarantee correctness and consistency, the generated summaries are carefully cleaned before evaluation. This step was crucial in removing any noise and unimportant data that the models may have introduced. The evaluation includes comparisons of different prompting techniques. 

\subsection{Comparisons of prompting techniques}
Subsequently, the effectiveness of different prompting techniques is compared to evaluate their impact on the performance of LLMs in text summarization. The analysis examines how each prompting method influences the quality and accuracy of the generated summaries. By comparing these techniques across various datasets, insights are provided into the strengths and weaknesses of each approach. These comparisons aims to highlight the most effective strategies for optimizing LLMs for summarization tasks.

\subsubsection{ZSL Results}
\begin{figure}[h]
    \centering
    \includegraphics[width=1\textwidth]{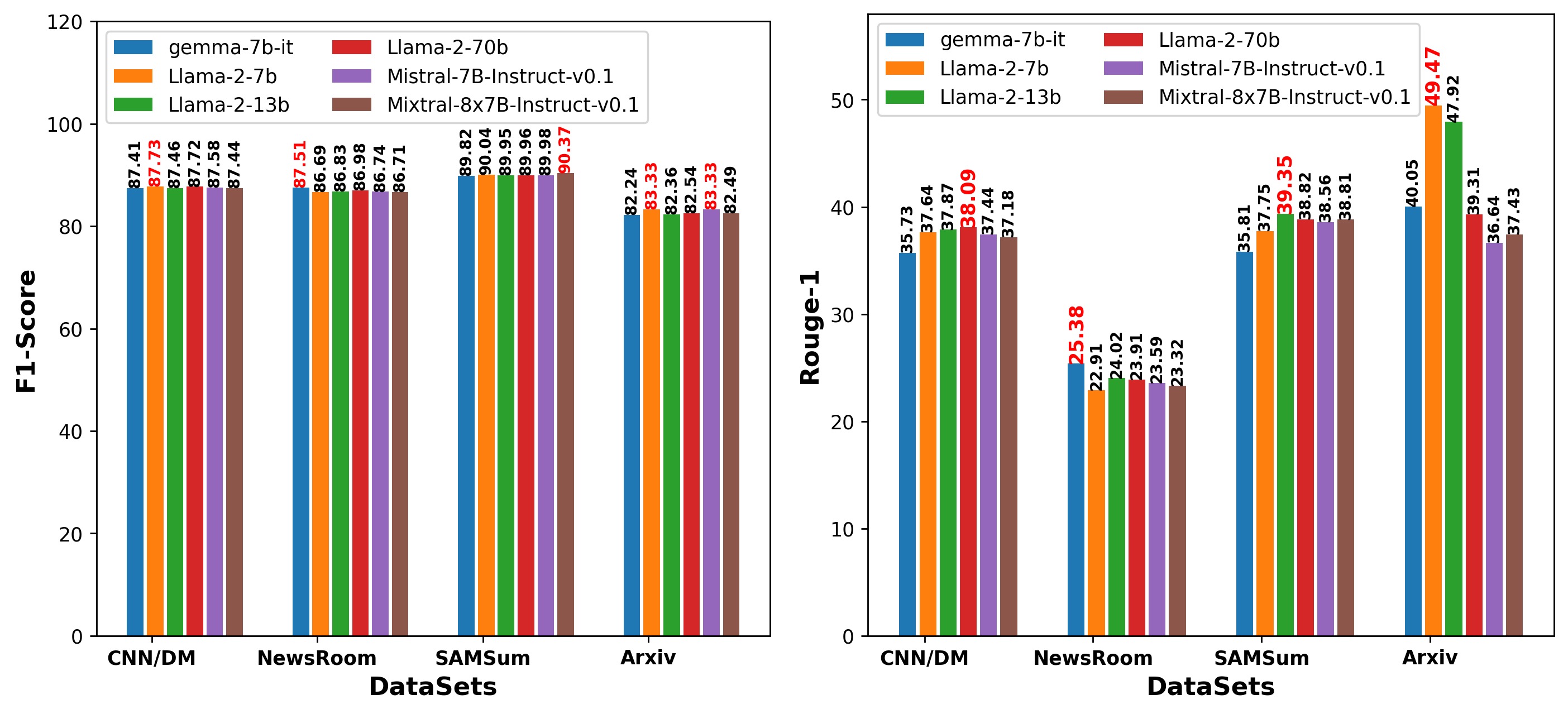}
    \caption{Bar Plots indicating the performance of LLMs across different summarization datasets in zero-shot setting. The X-axis represents the different datasets, and the Y-axis represents the performance of the models in Rouge-1 and F1 Score.}
    \label{fig:fig2}
\end{figure}

In zero-shot prompting, multiple prompts are used to guide the models in generating concise summaries that accurately describe the original articles. These prompts aim to provide clear and specific instructions to help the models understand the summarization task without any prior examples. Figure \ref{fig:fig2} offers a comparative analysis of the different LLMs used in our work across the different datasets. Figure 2.a shows the performance of the models measured by F1-Score as a part of the BERTScore metric, and Figure 2. b shows the results in Rouge-1.

\textbf{CNN/DM results with ZSL:\footnote{Note that only the prompts with the highest result are shown and all prompts can be found at \url{https://github.com/walid798/TextSummarizationCode}}} Table \ref{tab:table3}  illustrates the different prompts used in prompting the models to generate the summaries. Prompt design is an important topic in using LLMs as mentioned before. When designing the prompts, we used prompts that depend on instructing the model to generate \textbf{Length-constrained summaries} as shown in \textbf{prompts 3,4,5.} These prompts specify the maximum number of sentences allowed in the summary. The summary length is specified according to the average summary length of the dataset shown in Table \ref{tab:table1}. Other prompts are \textbf{Structured summarization prompts}, which provide specific instructions or guidelines for the models in the summarization process without necessarily focusing on length constraints.

Another deeper analysis of the result shown in Table \ref{tab:table5} can be discussed in two ways \textbf{\underline{Vertical and Horizontal}} Analysis. \textbf{\underline{Vertical Analysis}} compares the different models using the same prompt regarding Rouge and BERTScore metrics for each prompt. The comparison shows that Mistral-7B-Instruct-v0.1 achieves the best result on prompt\#1 regarding the ROUGE metric. This indicates that this specific prompt formulation optimally guides the model to generate summaries that align closely with the reference texts in lexical. For BERTScore, Llama-2-13b-chat  achieves the best results in prompt\#1 for Precision and Llama-2-70b-chat for Recall, and F1 Score, which closed slightly from the result of Mistral-7B-Instruct-v0.1 with BERTScore. This indicates that while Llama-2-70b-chat excels in lexical similarity and content coverage with this prompt, Mistral-7B-Instruct-v0.1 outperforms it in capturing more comprehensive information or achieving a better balance of BERTScore. For prompt\#2, Llama-2-13b-chat achieves the best performance in both ROUGE and BERTScore metrics except that Llama-2-7b-chat gets the best Recall. This indicates the model's ability to excel in ROUGE and BERTScore metrics, showcasing its proficiency in generating summaries that maintain content overlap and lexical similarity with this prompt. For prompt\#3 and prompt\#4, it's obvious it's a long and descriptive prompt about the desired output. Mistral-7B-Instruct-v0.1 and Llama-2-70b-chat get the best Rouge for prompt\#3 and prompt\#4 respectively. For BERTScore the best results span among several models gemma-7b-it, Llama-2-70b-chat, and Mistral-7B-Instruct-v0.1 for prompt\#3. Furthermore, gemma-7b-it, Llama-2-7b-chat-chat, and Llama-2-70b-chat. This implies that the models' output semantically is relevant to the reference text but not all models generated phrases lexically close to the reference summaries and the input document. 
  
\textbf{\underline{Horizontal Analysis}} compares the performance of the same model across the different prompts. The comparison shows that for gemma-7b-it model, prompt\#3 gets the best Rouge-1, Rouge-L, and F1 Score indicating that this prompt formulated well to get the best output of that model in terms of lexically and semantically. For Llama-2-7b-chat and Llama-2-13b-chat models, prompt\#5 and prompt\#2 get the best result in terms of Rouge-1, Rouge-L, and F1 Score. For Mistral-7B-Instruct-v0.1 and Mixtral-8x7B-Instruct-v0.1, prompt\#3 and prompt\#2 get the best results for both Rouge-1 and F1 Score respectively except that prompt\#5 get the best F1 Score for Mixtral-8x7B-Instruct-v0.1.

\begin{table}[h]
\centering
\begin{adjustbox}{max width=\textwidth} 
\renewcommand{\arraystretch}{1.5}

%\resizebox{\columnwidth}{!}{%
\begin{tabular}{|cc|cc|cc|cc|cc|clc|}
\hline
\multicolumn{2}{|c|}{Prompts} &
  \multicolumn{2}{c|}{\cellcolor[HTML]{C96969}\textbf{prompt\#1}} &
  \multicolumn{2}{c|}{\cellcolor[HTML]{C96969}\textbf{prompt\#2}} &
  \multicolumn{2}{c|}{\cellcolor[HTML]{C96969}\textbf{prompt\#3}} &
  \multicolumn{2}{c|}{\cellcolor[HTML]{C96969}\textbf{prompt\#4}} &
  \multicolumn{3}{c|}{\cellcolor[HTML]{C96969}\textbf{prompt\#5}} \\ \hline
\multicolumn{1}{|c|}{\textbf{Models}} &
  Metris &
  \multicolumn{1}{c|}{\textbf{Bert}} &
  \textbf{Rouge} &
  \multicolumn{1}{c|}{\textbf{Bert}} &
  \textbf{Rouge} &
  \multicolumn{1}{c|}{\textbf{Bert}} &
  \textbf{Rouge} &
  \multicolumn{1}{c|}{\textbf{Bert}} &
  \textbf{Rouge} &
  \multicolumn{2}{c|}{\textbf{Bert}} &
  \textbf{Rouge} \\ \hline
\multicolumn{1}{|c|}{\cellcolor[HTML]{C27BA0}} &
  \cellcolor[HTML]{FFFFFF}\textbf{P/R1} &
  \multicolumn{1}{c|}{\cellcolor[HTML]{FFFFFF}85.73} &
  \cellcolor[HTML]{FFFFFF}31.69 &
  \multicolumn{1}{c|}{\cellcolor[HTML]{FFFFFF}85.99} &
  \cellcolor[HTML]{FFFFFF}34.52 &
  \multicolumn{1}{c|}{\cellcolor[HTML]{FFFFFF}{\ul\textbf{88.76}}} &
  \cellcolor[HTML]{FFFFFF}36.66 &
  \multicolumn{1}{c|}{\cellcolor[HTML]{FFFFFF}\textbf{88.23}} &
  35.26 &
  \multicolumn{2}{c|}{\cellcolor[HTML]{FFFFFF}88.42} &
  \cellcolor[HTML]{FFFFFF}35.73 \\  
\rowcolor[HTML]{E7E6E6} 
\multicolumn{1}{|c|}{\cellcolor[HTML]{C27BA0}} &
  \cellcolor[HTML]{EFEFEF}\textbf{R/R2} &
  \multicolumn{1}{c|}{\cellcolor[HTML]{EFEFEF}88.24} &
  12.37 &
  \multicolumn{1}{c|}{\cellcolor[HTML]{E7E6E6}87.51} &
  13.04 &
  \multicolumn{1}{c|}{\cellcolor[HTML]{E7E6E6}86.13} &
  13.62 &
  \multicolumn{1}{c|}{\cellcolor[HTML]{E7E6E6}\textbf{85.91}} &
  12.38 &
  \multicolumn{2}{c|}{\cellcolor[HTML]{E7E6E6}85.91} &
  13.32 \\  
\rowcolor[HTML]{FFFFFF} 
\multicolumn{1}{|c|}{\multirow{-3}{*}{\cellcolor[HTML]{C27BA0}\textbf{gemma-7b-it}}} &
  \textbf{F1/RL} &
  \multicolumn{1}{c|}{\cellcolor[HTML]{FFFFFF}86.96} &
  20.17 &
  \multicolumn{1}{c|}{\cellcolor[HTML]{FFFFFF}86.72} &
  24.99 &
  \multicolumn{1}{c|}{\cellcolor[HTML]{FFFFFF}87.41} &
  24.48 &
  \multicolumn{1}{c|}{\cellcolor[HTML]{FFFFFF}87.04} &
  23.74 &
  \multicolumn{2}{c|}{\cellcolor[HTML]{FFFFFF}87.13} &
  23.81 \\ \hline
\rowcolor[HTML]{E7E6E6} 
\multicolumn{1}{|c|}{\cellcolor[HTML]{E6B8AF}} &
  \textbf{P/R1} &
  \multicolumn{1}{c|}{\cellcolor[HTML]{E7E6E6}84.84} &
  28.95 &
  \multicolumn{1}{c|}{\cellcolor[HTML]{E7E6E6}86.24} &
  32.29 &
  \multicolumn{1}{c|}{\cellcolor[HTML]{E7E6E6}86.73} &
  36.41 &
  \multicolumn{1}{c|}{\cellcolor[HTML]{E7E6E6}86.98} &
  37.19 &
  \multicolumn{2}{c|}{\cellcolor[HTML]{E7E6E6}88.05} &
  \cellcolor[HTML]{E7E6E6}37.64 \\  
\rowcolor[HTML]{FFFFFF} 
\multicolumn{1}{|c|}{\cellcolor[HTML]{E6B8AF}} &
  \textbf{R/R2} &
  \multicolumn{1}{c|}{\cellcolor[HTML]{FFFFFF}88.28} &
  11.66 &
  \multicolumn{1}{c|}{\cellcolor[HTML]{FFFFFF}\textbf{88.23}} &
  11.13 &
  \multicolumn{1}{c|}{\cellcolor[HTML]{FFFFFF}88.11} &
  14.59 &
  \multicolumn{1}{c|}{\cellcolor[HTML]{FFFFFF}87.88} &
  14.21 &
  \multicolumn{2}{c|}{\cellcolor[HTML]{FFFFFF}87.29} &
  14.06 \\ 
\rowcolor[HTML]{EFEFEF} 
\multicolumn{1}{|c|}{\multirow{-3}{*}{\cellcolor[HTML]{E6B8AF}\textbf{Llama-2-7b-chat}}} &
  \textbf{F1/RL} &
  \multicolumn{1}{c|}{\cellcolor[HTML]{EFEFEF}86.51} &
  19.84 &
  \multicolumn{1}{c|}{\cellcolor[HTML]{EFEFEF}87.22} &
  19.47 &
  \multicolumn{1}{c|}{\cellcolor[HTML]{E7E6E6}87.41} &
  \cellcolor[HTML]{E7E6E6}23.55 &
  \multicolumn{1}{c|}{\cellcolor[HTML]{E7E6E6}87.71} &
  23.51 &
  \multicolumn{2}{c|}{\cellcolor[HTML]{E7E6E6}87.66} &
  24.08 \\ \hline
\rowcolor[HTML]{FFFFFF} 
\multicolumn{1}{|c|}{\cellcolor[HTML]{70AD47}} &
  \textbf{P/R1} &
  \multicolumn{1}{c|}{\cellcolor[HTML]{FFFFFF}\textbf{85.98}} &
  32.13 &
  \multicolumn{1}{c|}{\cellcolor[HTML]{FFFFFF}\textbf{86.89}} &
  \textbf{37.87} &
  \multicolumn{1}{c|}{\cellcolor[HTML]{FFFFFF}87.41} &
  37.21 &
  \multicolumn{1}{c|}{\cellcolor[HTML]{FFFFFF}87.85} &
  37.48 &
  \multicolumn{2}{c|}{\cellcolor[HTML]{FFFFFF}87.76} &
  37.59 \\ 
\rowcolor[HTML]{E7E6E6} 
\multicolumn{1}{|c|}{\cellcolor[HTML]{70AD47}} &
  \cellcolor[HTML]{EFEFEF}\textbf{R/R2} &
  \multicolumn{1}{c|}{\cellcolor[HTML]{E7E6E6}88.17} &
  13.45 &
  \multicolumn{1}{c|}{\cellcolor[HTML]{E7E6E6}88.06} &
  {\ul\textbf{17.07}} &
  \multicolumn{1}{c|}{\cellcolor[HTML]{EFEFEF}87.56} &
  \cellcolor[HTML]{EFEFEF}13.57 &
  \multicolumn{1}{c|}{\cellcolor[HTML]{EFEFEF}86.71} &
  13.36 &
  \multicolumn{2}{c|}{\cellcolor[HTML]{E7E6E6}87.25} &
  13.89 \\ 
\rowcolor[HTML]{FFFFFF} 
\multicolumn{1}{|c|}{\multirow{-3}{*}{\cellcolor[HTML]{70AD47}\textbf{Llama-2-13b-chat}}} &
  \textbf{F1/RL} &
  \multicolumn{1}{c|}{\cellcolor[HTML]{FFFFFF}87.05} &
  21.58 &
  \multicolumn{1}{c|}{\cellcolor[HTML]{FFFFFF}\textbf{87.46}} &
  \textbf{25.99} &
  \multicolumn{1}{c|}{\cellcolor[HTML]{FFFFFF}87.47} &
  24.01 &
  \multicolumn{1}{c|}{\cellcolor[HTML]{FFFFFF}87.39} &
  24.15 &
  \multicolumn{2}{c|}{\cellcolor[HTML]{FFFFFF}87.45} &
  24.01 \\ \hline
\rowcolor[HTML]{FFFFFF} 
\multicolumn{1}{|c|}{\cellcolor[HTML]{FF9900}} &
  \textbf{P/R1} &
  \multicolumn{1}{c|}{\cellcolor[HTML]{FFFFFF}85.95} &
  29.44 &
  \multicolumn{1}{c|}{\cellcolor[HTML]{FFFFFF}86.81} &
  35.31 &
  \multicolumn{1}{c|}{\cellcolor[HTML]{FFFFFF}87.39} &
  37.08 &
  \multicolumn{1}{c|}{\cellcolor[HTML]{FFFFFF}87.92} &
  \cellcolor[HTML]{FFFFFF}\textbf{37.76} &
  \multicolumn{2}{c|}{\cellcolor[HTML]{FFFFFF}87.86} &
  {\ul\textbf{38.19}} \\  
\rowcolor[HTML]{EFEFEF} 
\multicolumn{1}{|c|}{\cellcolor[HTML]{FF9900}} &
  \textbf{R/R2} &
  \multicolumn{1}{c|}{\cellcolor[HTML]{EFEFEF}\textbf{88.53}} &
  11.02 &
  \multicolumn{1}{c|}{\cellcolor[HTML]{EFEFEF}88.07} &
  15.12 &
  \multicolumn{1}{c|}{\cellcolor[HTML]{EFEFEF}87.96} &
  \cellcolor[HTML]{E7E6E6}14.19 &
  \multicolumn{1}{c|}{\cellcolor[HTML]{E7E6E6}87.56} &
  \cellcolor[HTML]{E7E6E6}\textbf{14.82} &
  \multicolumn{2}{c|}{\cellcolor[HTML]{EFEFEF}{87.21}} &
  14.28 \\ 
\rowcolor[HTML]{FFFFFF} 
\multicolumn{1}{|c|}{\multirow{-3}{*}{\cellcolor[HTML]{FF9900}\textbf{Llama-2-70b-chat}}} &
  \textbf{F1/RL} &
  \multicolumn{1}{c|}{\cellcolor[HTML]{FFFFFF}\textbf{87.21}} &
  18.39 &
  \multicolumn{1}{c|}{\cellcolor[HTML]{FFFFFF}87.42} &
  25.29 &
  \multicolumn{1}{c|}{\cellcolor[HTML]{FFFFFF}\textbf{87.66}} &
  24.03 &
  \multicolumn{1}{c|}{{\ul\textbf{87.72}}} &
  \textbf{25.08} &
  \multicolumn{2}{c|}{\cellcolor[HTML]{FFFFFF}87.52} &
  {\ul\textbf{24.56}} \\ \hline
\rowcolor[HTML]{FFFFFF} 
\multicolumn{1}{|c|}{\cellcolor[HTML]{4472C4}} &
  \textbf{P/R1} &
  \multicolumn{1}{c|}{\cellcolor[HTML]{FFFFFF}85.89} &
  \textbf{37.02} &
  \multicolumn{1}{c|}{\cellcolor[HTML]{FFFFFF}86.57} &
  37.44 &
  \multicolumn{1}{c|}{\cellcolor[HTML]{FFFFFF}86.82} &
 \textbf{37.44} &
  \multicolumn{1}{c|}{\cellcolor[HTML]{FFFFFF}86.91} &
  36.78 &
  \multicolumn{2}{c|}{\cellcolor[HTML]{FFFFFF}87.01} &
  36.12 \\ 
\rowcolor[HTML]{E7E6E6} 
\multicolumn{1}{|c|}{\cellcolor[HTML]{4472C4}} &
  \cellcolor[HTML]{EFEFEF}\textbf{R/R2} &
  \multicolumn{1}{c|}{\cellcolor[HTML]{EFEFEF}88.05} &
  \cellcolor[HTML]{EFEFEF}\textbf{16.5} &
  \multicolumn{1}{c|}{\cellcolor[HTML]{E7E6E6}87.88} &
  15.14 &
  \multicolumn{1}{c|}{{\ul\textbf{ 88.38}}} &
  \textbf{16.42} &
  \multicolumn{1}{c|}{\cellcolor[HTML]{E7E6E6}87.83} &
  14.56 &
  \multicolumn{2}{c|}{\cellcolor[HTML]{E7E6E6}88.15} &
  15.34 \\ 
 
\multicolumn{1}{|c|}{\multirow{-3}{*}{\cellcolor[HTML]{4472C4}\textbf{\begin{tabular}[c]{@{}c@{}}Mistral-7B-\\ Instruct-v0.1\end{tabular}}}} &
  \textbf{F1/RL} &
  \multicolumn{1}{c|}{\cellcolor[HTML]{FFFFFF}86.94} &
  \textbf{24.35} &
  \multicolumn{1}{c|}{\cellcolor[HTML]{FFFFFF}87.21} &
  24.08 &
  \multicolumn{1}{c|}{\cellcolor[HTML]{FFFFFF}87.58} &
  \textbf{24.53} &
  \multicolumn{1}{c|}{\cellcolor[HTML]{FFFFFF}87.36} &
  23.85 &
  \multicolumn{2}{c|}{\cellcolor[HTML]{FFFFFF}87.56} &
  23.51 \\ \hline
\rowcolor[HTML]{E7E6E6} 
\multicolumn{1}{|c|}{\cellcolor[HTML]{EA9999}} &
  \cellcolor[HTML]{EFEFEF}\textbf{P/R1} &
  \multicolumn{1}{c|}{\cellcolor[HTML]{EFEFEF}85.38} &
  \cellcolor[HTML]{EFEFEF}33.88 &
  \multicolumn{1}{c|}{\cellcolor[HTML]{E7E6E6}86.33} &
  37.18 &
  \multicolumn{1}{c|}{\cellcolor[HTML]{E7E6E6}86.96} &
  35.54 &
  \multicolumn{1}{c|}{\cellcolor[HTML]{E7E6E6}86.91} &
  35.48 &
  \multicolumn{2}{c|}{\cellcolor[HTML]{E7E6E6}87.05} &
  35.96 \\ 
\rowcolor[HTML]{FFFFFF} 
\multicolumn{1}{|c|}{\cellcolor[HTML]{EA9999}} &
  \textbf{R/R2} &
  \multicolumn{1}{c|}{\cellcolor[HTML]{FFFFFF}88.25} &
  13.94 &
  \multicolumn{1}{c|}{\cellcolor[HTML]{FFFFFF}88.06} &
  15.13 &
  \multicolumn{1}{c|}{\cellcolor[HTML]{FFFFFF}87.25} &
  13.04 &
  \multicolumn{1}{c|}{\cellcolor[HTML]{FFFFFF}86.95} &
  12.28 &
  \multicolumn{2}{c|}{\cellcolor[HTML]{FFFFFF}87.88} &
  13.66 \\ 

\multicolumn{1}{|c|}{\multirow{-3}{*}{\cellcolor[HTML]{EA9999}\textbf{\begin{tabular}[c]{@{}c@{}}Mixtral-8x7B-\\ Instruct-v0.1\end{tabular}}}} &
  \cellcolor[HTML]{E7E6E6}\textbf{F1/RL} &
  \multicolumn{1}{c|}{\cellcolor[HTML]{E7E6E6}86.77} &
  \cellcolor[HTML]{E7E6E6}22.12 &
  \multicolumn{1}{c|}{\cellcolor[HTML]{E7E6E6}87.17} &
  24.93 &
  \multicolumn{1}{c|}{\cellcolor[HTML]{E7E6E6}87.09} &
  22.79 &
  \multicolumn{1}{c|}{\cellcolor[HTML]{E7E6E6}86.91} &
  22.02 &
  \multicolumn{2}{c|}{\cellcolor[HTML]{E7E6E6}87.44} &
  22.6 \\ \hline
\end{tabular}%
%}
\end{adjustbox}
\caption{Performance of LLMs on CNN/DM dataset using ZSL.}
\label{tab:table5}
\end{table}

    \textbf{NewsRoom results with ZSL:} The NewsRoom dataset contains articles that have summaries extracted from the original articles and other articles have abstractive summaries that can be seen as headlines for the original articles \cite{bib67}. In our study, only documents with abstractive summaries from the NewsRoom dataset were evaluated. This decision is motivated by several factors. Firstly, abstractive summarization more closely aligns with the goals of this research, which aims to generate concise, human-like summaries that capture the essence of the original text. Secondly, the evaluation of abstractive summaries provides a more rigorous test of the capabilities of LLMs in generating coherent and informative content. Results of NewsRoom dataset across the various LLMs and different prompts are shown in Table\ref{tab:table6}. Subsequently, a detailed description of the results is provided and analyzed from different dimensions.

\begin{table}[h]
\centering
\begin{adjustbox}{max width=\textwidth} 
\renewcommand{\arraystretch}{1.5}
%\label{tab:table5}
%\resizebox{\columnwidth}{!}{%
\begin{tabular}{|cc|cc|cc|cc|cc|clc|}
\hline
\multicolumn{2}{|c|}{Prompts} &
  \multicolumn{2}{c|}{\cellcolor[HTML]{C96969}\textbf{prompt\#1}} &
  \multicolumn{2}{c|}{\cellcolor[HTML]{C96969}\textbf{prompt\#2}} &
  \multicolumn{2}{c|}{\cellcolor[HTML]{C96969}\textbf{prompt\#3}} &
  \multicolumn{2}{c|}{\cellcolor[HTML]{C96969}\textbf{prompt\#4}} &
  \multicolumn{3}{c|}{\cellcolor[HTML]{C96969}\textbf{prompt\#5}} \\ \hline 
\multicolumn{1}{|c|}{\textbf{Models}} &
  Metrics &
  \multicolumn{1}{c|}{\textbf{Bert}} &
  \textbf{Rouge} &
  \multicolumn{1}{c|}{\textbf{Bert}} &
  \textbf{Rouge} &
  \multicolumn{1}{c|}{\textbf{Bert}} &
  \textbf{Rouge} &
  \multicolumn{1}{c|}{\textbf{Bert}} &
  \textbf{Rouge} &
  \multicolumn{2}{c|}{\textbf{Bert}} &
  \textbf{Rouge} \\ \hline
\rowcolor[HTML]{FFFFFF} 
\multicolumn{1}{|c|}{\cellcolor[HTML]{C27BA0}} &
  \cellcolor[HTML]{FFFFFF}\textbf{P/R1} &
  \multicolumn{1}{c|}{{\ul \textbf{86.99}}} &
  {\ul \textbf{25.38}} &
  \multicolumn{1}{c|}{\cellcolor[HTML]{FFFFFF}86.72} &
  24.28 &
  \multicolumn{1}{c|}{\cellcolor[HTML]{FFFFFF}85.63} &
  21.85 &
  \multicolumn{1}{c|}{\cellcolor[HTML]{FFFFFF}85.93} &
  22.89 &
  \multicolumn{2}{c|}{\cellcolor[HTML]{FFFFFF}86.77} &
  35.73 \\   
\rowcolor[HTML]{E7E6E6} 
\multicolumn{1}{|c|}{\cellcolor[HTML]{C27BA0}} &
  \cellcolor[HTML]{EFEFEF}\textbf{R/R2} &
  \multicolumn{1}{c|}{{\ul \textbf{87.67}}} &
  {\ul \textbf{8.82}} &
  \multicolumn{1}{c|}{\cellcolor[HTML]{E7E6E6}87.71} &
  8.14 &
  \multicolumn{1}{c|}{\cellcolor[HTML]{E7E6E6}87.75} &
  6.99 &
  \multicolumn{1}{c|}{\cellcolor[HTML]{E7E6E6}87.74} &
  7.594 &
  \multicolumn{2}{c|}{\cellcolor[HTML]{E7E6E6}87.61} &
  13.32 \\   
\rowcolor[HTML]{FFFFFF} 
\multicolumn{1}{|c|}{\multirow{-3}{*}{\cellcolor[HTML]{C27BA0}\textbf{gemma-7b-it}}} &
  \cellcolor[HTML]{FFFFFF}\textbf{F1/RL} &
  \multicolumn{1}{c|}{{\ul \textbf{87.31}}} &
  {\ul \textbf{19.96}} &
  \multicolumn{1}{c|}{\cellcolor[HTML]{FFFFFF}87.19} &
  18.81 &
  \multicolumn{1}{c|}{\cellcolor[HTML]{FFFFFF}86.65} &
  16.44 &
  \multicolumn{1}{c|}{\cellcolor[HTML]{FFFFFF}86.81} &
  17.18 &
  \multicolumn{2}{c|}{\cellcolor[HTML]{FFFFFF}87.16} &
  23.81 \\ \hline
\rowcolor[HTML]{E7E6E6} 
\multicolumn{1}{|c|}{\cellcolor[HTML]{E6B8AF}} &
  \textbf{P/R1} &
  \multicolumn{1}{c|}{\cellcolor[HTML]{E7E6E6}85.19} &
  \textit{\textbf{22.78}} &
  \multicolumn{1}{c|}{\cellcolor[HTML]{E7E6E6}84.71} &
  21.69 &
  \multicolumn{1}{c|}{\cellcolor[HTML]{E7E6E6}84.51} &
  20.13 &
  \multicolumn{1}{c|}{\cellcolor[HTML]{E7E6E6}85.38} &
  22.15 &
  \multicolumn{2}{c|}{\cellcolor[HTML]{E7E6E6}84.84} &
  37.64 \\   
\rowcolor[HTML]{FFFFFF} 
\multicolumn{1}{|c|}{\cellcolor[HTML]{E6B8AF}} &
  \cellcolor[HTML]{FFFFFF}\textbf{R/R2} &
  \multicolumn{1}{c|}{\cellcolor[HTML]{FFFFFF}88.23} &
  7.81 &
  \multicolumn{1}{c|}{\cellcolor[HTML]{FFFFFF}88.29} &
  7.45 &
  \multicolumn{1}{c|}{\cellcolor[HTML]{FFFFFF}88.23} &
  6.717 &
  \multicolumn{1}{c|}{\cellcolor[HTML]{FFFFFF}88.04} &
  6.87 &
  \multicolumn{2}{c|}{\cellcolor[HTML]{FFFFFF}87.97} &
  14.06 \\   
\rowcolor[HTML]{E7E6E6} 
\multicolumn{1}{|c|}{\multirow{-3}{*}{\cellcolor[HTML]{E6B8AF}\textbf{Llama-2-7b-chat}}} &
  \textbf{F1/RL} &
  \multicolumn{1}{c|}{\cellcolor[HTML]{E7E6E6}\textit{\textbf{86.67}}} &
  17.07 &
  \multicolumn{1}{c|}{\cellcolor[HTML]{E7E6E6}86.44} &
  15.97 &
  \multicolumn{1}{c|}{\cellcolor[HTML]{E7E6E6}86.31} &
  14.73 &
  \multicolumn{1}{c|}{\cellcolor[HTML]{E7E6E6}86.67} &
  16.15 &
  \multicolumn{2}{c|}{\cellcolor[HTML]{E7E6E6}86.35} &
  24.08 \\ \hline
\rowcolor[HTML]{FFFFFF} 
\multicolumn{1}{|c|}{\cellcolor[HTML]{70AD47}} &
  \cellcolor[HTML]{FFFFFF}\textbf{P/R1} &
  \multicolumn{1}{c|}{\cellcolor[HTML]{FFFFFF}85.42} &
  \textit{\textbf{24.02}} &
  \multicolumn{1}{c|}{\cellcolor[HTML]{FFFFFF}85.07} &
  22.99 &
  \multicolumn{1}{c|}{\cellcolor[HTML]{FFFFFF}84.88} &
  21.38 &
  \multicolumn{1}{c|}{\cellcolor[HTML]{FFFFFF}85.69} &
  21.76 &
  \multicolumn{2}{c|}{\cellcolor[HTML]{FFFFFF}85.19} &
  37.59 \\   
\rowcolor[HTML]{E7E6E6} 
\multicolumn{1}{|c|}{\cellcolor[HTML]{70AD47}} &
  \textbf{R/R2} &
  \multicolumn{1}{c|}{\cellcolor[HTML]{E7E6E6}88.34} &
  8.19 &
  \multicolumn{1}{c|}{\cellcolor[HTML]{E7E6E6}88.41} &
  8.054 &
  \multicolumn{1}{c|}{\cellcolor[HTML]{E7E6E6}88.41} &
  7.019 &
  \multicolumn{1}{c|}{\cellcolor[HTML]{E7E6E6}87.72} &
  6.78 &
  \multicolumn{2}{c|}{\cellcolor[HTML]{E7E6E6}88.17} &
  13.89 \\   
\rowcolor[HTML]{FFFFFF} 
\multicolumn{1}{|c|}{\multirow{-3}{*}{\cellcolor[HTML]{70AD47}\textbf{Llama-2-13b-chat}}} &
  \cellcolor[HTML]{FFFFFF}\textbf{F1/RL} &
  \multicolumn{1}{c|}{\cellcolor[HTML]{FFFFFF}\textit{\textbf{86.83}}} &
  17.89 &
  \multicolumn{1}{c|}{\cellcolor[HTML]{FFFFFF}86.68} &
  17.17 &
  \multicolumn{1}{c|}{\cellcolor[HTML]{FFFFFF}86.58} &
  15.61 &
  \multicolumn{1}{c|}{\cellcolor[HTML]{FFFFFF}86.68} &
  16.26 &
  \multicolumn{2}{c|}{\cellcolor[HTML]{FFFFFF}86.63} &
  24.01 \\ \hline
\rowcolor[HTML]{E7E6E6} 
\multicolumn{1}{|c|}{\cellcolor[HTML]{FF9900}} &
  \textbf{P/R1} &
  \multicolumn{1}{c|}{\cellcolor[HTML]{E7E6E6}85.71} &
  \textit{\textbf{23.71}} &
  \multicolumn{1}{c|}{\cellcolor[HTML]{E7E6E6}85.37} &
  23.47 &
  \multicolumn{1}{c|}{\cellcolor[HTML]{E7E6E6}84.75} &
  21.17 &
  \multicolumn{1}{c|}{\cellcolor[HTML]{E7E6E6}85.4} &
  21.81 &
  \multicolumn{2}{c|}{\cellcolor[HTML]{E7E6E6}85.61} &
  38.19 \\   
\rowcolor[HTML]{FFFFFF} 
\multicolumn{1}{|c|}{\cellcolor[HTML]{FF9900}} &
  \textbf{R/R2} &
  \multicolumn{1}{c|}{\cellcolor[HTML]{FFFFFF}88.31} &
  8.254 &
  \multicolumn{1}{c|}{\cellcolor[HTML]{FFFFFF}88.39} &
  8.038 &
  \multicolumn{1}{c|}{\cellcolor[HTML]{FFFFFF}88.25} &
  7.16 &
  \multicolumn{1}{c|}{\cellcolor[HTML]{FFFFFF}88.11} &
  7.17 &
  \multicolumn{2}{c|}{\cellcolor[HTML]{FFFFFF}88.06} &
  14.28 \\   
\rowcolor[HTML]{E7E6E6} 
\multicolumn{1}{|c|}{\multirow{-3}{*}{\cellcolor[HTML]{FF9900}\textbf{Llama-2-70b-chat}}} &
  \textbf{F1/RL} &
  \multicolumn{1}{c|}{\cellcolor[HTML]{E7E6E6}\textit{\textbf{86.97}}} &
  18.19 &
  \multicolumn{1}{c|}{\cellcolor[HTML]{E7E6E6}86.84} &
  17.51 &
  \multicolumn{1}{c|}{\cellcolor[HTML]{E7E6E6}86.44} &
  15.59 &
  \multicolumn{1}{c|}{\cellcolor[HTML]{E7E6E6}86.71} &
  16.03 &
  \multicolumn{2}{c|}{\cellcolor[HTML]{E7E6E6}86.79} &
  24.56 \\ \hline
\rowcolor[HTML]{FFFFFF} 
\multicolumn{1}{|c|}{\cellcolor[HTML]{4472C4}} &
  \textbf{P/R1} &
  \multicolumn{1}{c|}{\cellcolor[HTML]{FFFFFF}85.4} &
  \textit{\textbf{23.59}} &
  \multicolumn{1}{c|}{\cellcolor[HTML]{FFFFFF}84.42} &
  20.59 &
  \multicolumn{1}{c|}{\cellcolor[HTML]{FFFFFF}83.43} &
  17.33 &
  \multicolumn{1}{c|}{\cellcolor[HTML]{FFFFFF}83.56} &
  17.44 &
  \multicolumn{2}{c|}{\cellcolor[HTML]{FFFFFF}85.46} &
  36.12 \\   
\rowcolor[HTML]{E7E6E6} 
\multicolumn{1}{|c|}{\cellcolor[HTML]{4472C4}} &
  \textbf{R/R2} &
  \multicolumn{1}{c|}{\cellcolor[HTML]{E7E6E6}88.17} &
  9.102 &
  \multicolumn{1}{c|}{\cellcolor[HTML]{E7E6E6}88.09} &
  7.52 &
  \multicolumn{1}{c|}{\cellcolor[HTML]{E7E6E6}88.01} &
  6.009 &
  \multicolumn{1}{c|}{\cellcolor[HTML]{E7E6E6}88.56} &
  6.024 &
  \multicolumn{2}{c|}{\cellcolor[HTML]{E7E6E6}87.87} &
  15.34 \\   

\multicolumn{1}{|c|}{\multirow{-3}{*}{\cellcolor[HTML]{4472C4}\textbf{\begin{tabular}[c]{@{}c@{}}Mistral-7B-\\ Instruct-v0.1\end{tabular}}}} &
  \textbf{F1/RL} &
  \multicolumn{1}{c|}{\cellcolor[HTML]{FFFFFF}\textit{\textbf{86.74}}} &
  18.34 &
  \multicolumn{1}{c|}{\cellcolor[HTML]{FFFFFF}86.18} &
  15.61 &
  \multicolumn{1}{c|}{\cellcolor[HTML]{FFFFFF}85.63} &
  12.84 &
  \multicolumn{1}{c|}{\cellcolor[HTML]{FFFFFF}85.69} &
  12.95 &
  \multicolumn{2}{c|}{\cellcolor[HTML]{FFFFFF}86.62} &
  23.51 \\ \hline
\rowcolor[HTML]{E7E6E6} 
\multicolumn{1}{|c|}{\cellcolor[HTML]{EA9999}} &
  \textbf{P/R1} &
  \multicolumn{1}{c|}{\cellcolor[HTML]{E7E6E6}85.39} &
  \textit{\textbf{23.32}} &
  \multicolumn{1}{c|}{\cellcolor[HTML]{E7E6E6}84.57} &
  21.62 &
  \multicolumn{1}{c|}{\cellcolor[HTML]{E7E6E6}83.96} &
  19.12 &
  \multicolumn{1}{c|}{\cellcolor[HTML]{E7E6E6}83.98} &
  19.52 &
  \multicolumn{2}{c|}{\cellcolor[HTML]{E7E6E6}84.89} &
  35.96 \\   
\rowcolor[HTML]{FFFFFF} 
\multicolumn{1}{|c|}{\cellcolor[HTML]{EA9999}} &
  \textbf{R/R2} &
  \multicolumn{1}{c|}{\cellcolor[HTML]{FFFFFF}88.11} &
  7.72 &
  \multicolumn{1}{c|}{\cellcolor[HTML]{FFFFFF}88.17} &
  7.061 &
  \multicolumn{1}{c|}{\cellcolor[HTML]{FFFFFF}88.07} &
  6.365 &
  \multicolumn{1}{c|}{\cellcolor[HTML]{FFFFFF}88.86} &
  7.52 &
  \multicolumn{2}{c|}{\cellcolor[HTML]{FFFFFF}87.79} &
  13.66 \\   

\multicolumn{1}{|c|}{\multirow{-3}{*}{\cellcolor[HTML]{EA9999}\textbf{\begin{tabular}[c]{@{}c@{}}Mixtral-8x7B-\\ Instruct-v0.1\end{tabular}}}} &
  \textbf{F1/RL} &
  \multicolumn{1}{c|}{\cellcolor[HTML]{E7E6E6}\textit{\textbf{86.71}}} &
  17.21 &
  \multicolumn{1}{c|}{\cellcolor[HTML]{E7E6E6}86.32} &
  15.83 &
  \multicolumn{1}{c|}{\cellcolor[HTML]{E7E6E6}85.95} &
  13.97 &
  \multicolumn{1}{c|}{\cellcolor[HTML]{E7E6E6}86.01} &
  14.23 &
  \multicolumn{2}{c|}{\cellcolor[HTML]{E7E6E6}86.29} &
  22.6 \\ \hline
\end{tabular}%
\end{adjustbox}
%}
\caption{Performance of LLMs on NewsRoom dataset using ZSL.}
\label{tab:table6}
\end{table}

For BERTScore analysis of the results, gemma-7b-it consistently performs well in BERTScore across all prompts, with the highest Precision (86.99), Recall (87.67), and F1 (87.31) using prompt\#1. Llama-2-70b-chat achieves a high BERTScore F1 across prompts, with a peak F1 score of 86.97 using prompt\#1. Mistral-7B-Instruct-v0.1 demonstrates strong performance in BERTScore Recall (88.17) using prompt\#1 and Precision (85.37) across other prompts.

In terms of Rouge metrics, gemma-7b-it scores highest in ROUGE-1 (25.38), ROUGE-2 (8.82), and ROUGE-L (19.96) using prompt\#1, indicating strong summarization performance. Llama-2-13b-chat and Llama-2-70b-chat show competitive ROUGE scores, with notable performance in ROUGE-1 and ROUGE-L using various prompts. Mistral-7B-Instruct-v0.1 achieves the highest ROUGE-1 score (23.59) with prompt\#1, suggesting effective summarization capabilities for this specific prompt.

Regarding prompt effectiveness, prompt\#1 and prompt\#2 generally lead to higher BERTScore and ROUGE scores across most models. This implies these prompts are clear and direct, which may help the models generate more accurate and focused summaries. Prompts\#3 to Prompt\#5 show slightly lower performance, which may be attributed to their varying lengths and instructions. While all prompts are fundamentally similar, the slight variations in wording could influence how models interpret and respond to the summarization task.

The overall performance of the employed models on the NewsRoom dataset shows that gemma-7b-it leads in BERTScore metrics with prompt\#1, highlighting its proficiency in producing high-quality summaries. Llama-2-70b-chat consistently performs well across different prompts, showcasing its robustness and adaptability. Mistral-7B-Instruct-v0.1 shows strong performance in both BERTScore and ROUGE metrics, particularly with prompt\#1, making it a versatile model for summarization tasks.

\textbf{SAMSum results with ZSL:} The SAMSum dataset, comprising abstractive dialogue summaries, presents unique challenges for automated summarization due to its conversational nature. Table \ref{tab:table7}  provides a detailed comparison of models using various prompts to evaluate their effectiveness in generating concise and coherent summaries of the conversations. Notably, the models achieve high ROUGE scores, surpassing those typically obtained on news article datasets, highlighting the distinctiveness of dialogue summarization.

\begin{table}[]
\centering
\begin{adjustbox}{max width=\textwidth} 
\renewcommand{\arraystretch}{1.5}
%\resizebox{\columnwidth}{!}{%
\begin{tabular}{|cc|cc|cc|cc|cc|cc|}

\hline
\multicolumn{2}{|c|}{Prompts} &
  \multicolumn{2}{c|}{\cellcolor[HTML]{C96969}\textbf{prompt\#1}} &
  \multicolumn{2}{c|}{\cellcolor[HTML]{C96969}\textbf{prompt\#2}} &
  \multicolumn{2}{c|}{\cellcolor[HTML]{C96969}\textbf{prompt\#3}} &
  \multicolumn{2}{c|}{\cellcolor[HTML]{C96969}\textbf{prompt\#4}} &
  \multicolumn{2}{c|}{\cellcolor[HTML]{C96969}\textbf{prompt\#5}} \\ \hline
\multicolumn{1}{|c|}{\textbf{Models}} &
  Metrics &
  \multicolumn{1}{c|}{\textbf{Bert}} &
  \textbf{Rouge} &
  \multicolumn{1}{c|}{\textbf{Bert}} &
  \textbf{Rouge} &
  \multicolumn{1}{c|}{\textbf{Bert}} &
  \textbf{Rouge} &
  \multicolumn{1}{c|}{\textbf{Bert}} &
  \textbf{Rouge} &
  \multicolumn{1}{c|}{\textbf{Bert}} &
  \textbf{Rouge} \\ \hline
\rowcolor[HTML]{FFFFFF} 
\multicolumn{1}{|c|}{\cellcolor[HTML]{C27BA0}} &
  \textbf{P/R1} &
  \multicolumn{1}{c|}{\cellcolor[HTML]{FFFFFF}{\ul \textbf{86.24}}} &
  {\ul \textbf{33.16}} &
  \multicolumn{1}{c|}{\cellcolor[HTML]{FFFFFF}88.21} &
  34.31 &
  \multicolumn{1}{c|}{\cellcolor[HTML]{FFFFFF}86.41} &
  32.05 &
  \multicolumn{1}{c|}{\cellcolor[HTML]{FFFFFF}\textbf{88.76}} &
  35.23 &
  \multicolumn{1}{c|}{\cellcolor[HTML]{FFFFFF}86.29} &
  28.71 \\   
\rowcolor[HTML]{EFEFEF} 
\multicolumn{1}{|c|}{\cellcolor[HTML]{C27BA0}} &
  \textbf{R/R2} &
  \multicolumn{1}{c|}{\cellcolor[HTML]{EFEFEF}{\ul \textbf{91.05}}} &
  {\ul \textbf{11.48}} &
  \multicolumn{1}{c|}{\cellcolor[HTML]{EFEFEF}90.1} &
  11.7 &
  \multicolumn{1}{c|}{\cellcolor[HTML]{EFEFEF}90.68} &
  10.29 &
  \multicolumn{1}{c|}{\cellcolor[HTML]{EFEFEF}90.62} &
  12.14 &
  \multicolumn{1}{c|}{\cellcolor[HTML]{EFEFEF}89.45} &
  7.9 \\   
\rowcolor[HTML]{FFFFFF} 
\multicolumn{1}{|c|}{\multirow{-3}{*}{\cellcolor[HTML]{C27BA0}\textbf{gemma-7b-it}}} &
  \textbf{F1/RL} &
  \multicolumn{1}{c|}{\cellcolor[HTML]{FFFFFF}{\ul \textbf{88.58}}} &
  {\ul \textbf{25.25}} &
  \multicolumn{1}{c|}{\cellcolor[HTML]{FFFFFF}89.02} &
  26.02 &
  \multicolumn{1}{c|}{\cellcolor[HTML]{FFFFFF}88.48} &
  24.29 &
  \multicolumn{1}{c|}{\cellcolor[HTML]{FFFFFF}89.66} &
  26.72 &
  \multicolumn{1}{c|}{\cellcolor[HTML]{FFFFFF}87.81} &
  20.97 \\ \hline
\rowcolor[HTML]{EFEFEF} 
\multicolumn{1}{|c|}{\cellcolor[HTML]{E6B8AF}} &
  \textbf{P/R1} &
  \multicolumn{1}{c|}{\cellcolor[HTML]{EFEFEF}85.12} &
  \textit{\textbf{32.72}} &
  \multicolumn{1}{c|}{\cellcolor[HTML]{EFEFEF}89.48} &
  37.75 &
  \multicolumn{1}{c|}{\cellcolor[HTML]{EFEFEF}86.94} &
  34.22 &
  \multicolumn{1}{c|}{\cellcolor[HTML]{EFEFEF}88.76} &
  37.73 &
  \multicolumn{1}{c|}{\cellcolor[HTML]{EFEFEF}86.14} &
  32.34 \\   
\rowcolor[HTML]{FFFFFF} 
\multicolumn{1}{|c|}{\cellcolor[HTML]{E6B8AF}} &
  \textbf{R/R2} &
  \multicolumn{1}{c|}{\cellcolor[HTML]{FFFFFF}90.51} &
  10.8 &
  \multicolumn{1}{c|}{\cellcolor[HTML]{FFFFFF}90.64} &
  13.35 &
  \multicolumn{1}{c|}{\cellcolor[HTML]{FFFFFF}90.55} &
  11.31 &
  \multicolumn{1}{c|}{\cellcolor[HTML]{FFFFFF}90.62} &
  13.47 &
  \multicolumn{1}{c|}{\cellcolor[HTML]{FFFFFF}90.51} &
  10.66 \\   
\rowcolor[HTML]{EFEFEF} 
\multicolumn{1}{|c|}{\multirow{-3}{*}{\cellcolor[HTML]{E6B8AF}\textbf{Llama-2-7b-chat}}} &
  \textbf{F1/RL} &
  \multicolumn{1}{c|}{\cellcolor[HTML]{EFEFEF}\textit{\textbf{87.72}}} &
  24.86 &
  \multicolumn{1}{c|}{\cellcolor[HTML]{EFEFEF}90.04} &
  29.9 &
  \multicolumn{1}{c|}{\cellcolor[HTML]{EFEFEF}88.69} &
  26.09 &
  \multicolumn{1}{c|}{\cellcolor[HTML]{EFEFEF}89.66} &
  30 &
  \multicolumn{1}{c|}{\cellcolor[HTML]{EFEFEF}87.81} &
  24.51 \\ \hline
\rowcolor[HTML]{FFFFFF} 
\multicolumn{1}{|c|}{\cellcolor[HTML]{70AD47}} &
  \textbf{P/R1} &
  \multicolumn{1}{c|}{\cellcolor[HTML]{FFFFFF}86.09} &
  \textit{\textbf{34.48}} &
  \multicolumn{1}{c|}{\cellcolor[HTML]{FFFFFF}89.23} &
  \cellcolor[HTML]{FFFFFF}\textit{\textbf{38.96}} &
  \multicolumn{1}{c|}{\cellcolor[HTML]{FFFFFF}88.29} &
  36.66 &
  \multicolumn{1}{c|}{\cellcolor[HTML]{FFFFFF}89.16} &
  {\ul \textbf{39.35}} &
  \multicolumn{1}{c|}{\cellcolor[HTML]{FFFFFF}87.34} &
  32.23 \\   
\rowcolor[HTML]{EFEFEF} 
\multicolumn{1}{|c|}{\cellcolor[HTML]{70AD47}} &
  \textbf{R/R2} &
  \multicolumn{1}{c|}{\cellcolor[HTML]{EFEFEF}90.55} &
  12.41 &
  \multicolumn{1}{c|}{\cellcolor[HTML]{EFEFEF}90.72} &
  14.22 &
  \multicolumn{1}{c|}{\cellcolor[HTML]{EFEFEF}91.27} &
  12.41 &
  \multicolumn{1}{c|}{\cellcolor[HTML]{EFEFEF}90.78} &
  14.61 &
  \multicolumn{1}{c|}{\cellcolor[HTML]{EFEFEF}90.13} &
  10.45 \\   
\rowcolor[HTML]{FFFFFF} 
\multicolumn{1}{|c|}{\multirow{-3}{*}{\cellcolor[HTML]{70AD47}\textbf{Llama-2-13b-chat}}} &
  \textbf{F1/RL} &
  \multicolumn{1}{c|}{\cellcolor[HTML]{FFFFFF}\textit{\textbf{88.26}}} &
  26.7 &
  \multicolumn{1}{c|}{\cellcolor[HTML]{FFFFFF}89.95} &
  30.47 &
  \multicolumn{1}{c|}{\cellcolor[HTML]{FFFFFF}89.74} &
  27.68 &
  \multicolumn{1}{c|}{\cellcolor[HTML]{FFFFFF}89.88} &
  30.7 &
  \multicolumn{1}{c|}{\cellcolor[HTML]{FFFFFF}88.69} &
  24.09 \\ \hline
\rowcolor[HTML]{EFEFEF} 
\multicolumn{1}{|c|}{\cellcolor[HTML]{FF9900}} &
  \textbf{P/R1} &
  \multicolumn{1}{c|}{\cellcolor[HTML]{EFEFEF}87.02} &
  \cellcolor[HTML]{EFEFEF}\textit{\textbf{35.11}} &
  \multicolumn{1}{c|}{\cellcolor[HTML]{EFEFEF}89.09} &
  38.31 &
  \multicolumn{1}{c|}{\cellcolor[HTML]{EFEFEF}88.56} &
  \cellcolor[HTML]{EFEFEF}\textit{\textbf{37.4}} &
  \multicolumn{1}{c|}{\cellcolor[HTML]{EFEFEF}89.09} &
  38.82 &
  \multicolumn{1}{c|}{\cellcolor[HTML]{EFEFEF}87.16} &
  30.14 \\   
\rowcolor[HTML]{FFFFFF} 
\multicolumn{1}{|c|}{\cellcolor[HTML]{FF9900}} &
  \textbf{R/R2} &
  \multicolumn{1}{c|}{\cellcolor[HTML]{FFFFFF}91.14} &
  12.32 &
  \multicolumn{1}{c|}{\cellcolor[HTML]{FFFFFF}90.14} &
  14.21 &
  \multicolumn{1}{c|}{\cellcolor[HTML]{FFFFFF}91.34} &
  13.47 &
  \multicolumn{1}{c|}{\cellcolor[HTML]{FFFFFF}90.89} &
   {\ul \textbf{14.97}} &
  \multicolumn{1}{c|}{\cellcolor[HTML]{FFFFFF}88.77} &
  8.59 \\   
\rowcolor[HTML]{EFEFEF} 
\multicolumn{1}{|c|}{\multirow{-3}{*}{\cellcolor[HTML]{FF9900}\textbf{Llama-2-70b-chat}}} &
  \textbf{F1/RL} &
  \multicolumn{1}{c|}{\cellcolor[HTML]{EFEFEF}\textit{\textbf{89.02}}} &
  26.98 &
  \multicolumn{1}{c|}{\cellcolor[HTML]{EFEFEF}89.59} &
  30.2 &
  \multicolumn{1}{c|}{\cellcolor[HTML]{EFEFEF}\textit{\textbf{89.91}}} &
  29.03 &
  \multicolumn{1}{c|}{\cellcolor[HTML]{EFEFEF}89.96} &
  {\ul \textbf{30.59}} &
  \multicolumn{1}{c|}{\cellcolor[HTML]{EFEFEF}87.93} &
  22.04 \\ \hline
\rowcolor[HTML]{FFFFFF} 
\multicolumn{1}{|c|}{\cellcolor[HTML]{4472C4}} &
  \textbf{P/R1} &
  \multicolumn{1}{c|}{\cellcolor[HTML]{FFFFFF}87.01} &
  \textit{\textbf{34.24}} &
  \multicolumn{1}{c|}{\cellcolor[HTML]{FFFFFF}89.11} &
  38.56 &
  \multicolumn{1}{c|}{\cellcolor[HTML]{FFFFFF}87.64} &
  34.85 &
  \multicolumn{1}{c|}{\cellcolor[HTML]{FFFFFF}89.07} &
  38.55 &
  \multicolumn{1}{c|}{\cellcolor[HTML]{FFFFFF}86.87} &
  \cellcolor[HTML]{FFFFFF}\textbf{32.56} \\   
\rowcolor[HTML]{EFEFEF} 
\multicolumn{1}{|c|}{\cellcolor[HTML]{4472C4}} &
  \textbf{R/R2} &
  \multicolumn{1}{c|}{\cellcolor[HTML]{EFEFEF}91.51} &
  11.91 &
  \multicolumn{1}{c|}{\cellcolor[HTML]{EFEFEF}90.92} &
  13.94 &
  \multicolumn{1}{c|}{\cellcolor[HTML]{EFEFEF}91.12} &
  12.03 &
  \multicolumn{1}{c|}{\cellcolor[HTML]{EFEFEF}90.71} &
  14.31 &
  \multicolumn{1}{c|}{\cellcolor[HTML]{EFEFEF}91.03} &
  10.79 \\   

\multicolumn{1}{|c|}{\multirow{-3}{*}{\cellcolor[HTML]{4472C4}\textbf{\begin{tabular}[c]{@{}c@{}}Mistral-7B-\\ Instruct-v0.1\end{tabular}}}} &
  \textbf{F1/RL} &
  \multicolumn{1}{c|}{\cellcolor[HTML]{FFFFFF}\textit{\textbf{89.19}}} &
  26.51 &
  \multicolumn{1}{c|}{\cellcolor[HTML]{FFFFFF}89.98} &
  30.26 &
  \multicolumn{1}{c|}{\cellcolor[HTML]{FFFFFF}89.33} &
  26.64 &
  \multicolumn{1}{c|}{\cellcolor[HTML]{FFFFFF}89.86} &
  30.38 &
  \multicolumn{1}{c|}{\cellcolor[HTML]{FFFFFF}\textit{\textbf{88.88}}} &
  24.67 \\ \hline
\rowcolor[HTML]{EFEFEF} 
\multicolumn{1}{|c|}{\cellcolor[HTML]{EA9999}} &
  \textbf{P/R1} &
  \multicolumn{1}{c|}{\cellcolor[HTML]{EFEFEF}86.83} &
  \textit{\textbf{33.67}} &
  \multicolumn{1}{c|}{\cellcolor[HTML]{EFEFEF}89.31} &
  38.81 &
  \multicolumn{1}{c|}{\cellcolor[HTML]{EFEFEF}87.65} &
  35.07 &
  \multicolumn{1}{c|}{{\ul \textbf{89.38}}} &
  37.97 &
  \multicolumn{1}{c|}{\cellcolor[HTML]{EFEFEF}86.25} &
  31.53 \\   
\rowcolor[HTML]{FFFFFF} 
\multicolumn{1}{|c|}{\cellcolor[HTML]{EA9999}} &
  \textbf{R/R2} &
  \multicolumn{1}{c|}{\cellcolor[HTML]{FFFFFF}91.18} &
  10.78 &
  \multicolumn{1}{c|}{{\ul \textbf{91.48}}} &
  13.89 &
  \multicolumn{1}{c|}{\cellcolor[HTML]{FFFFFF}91.19} &
  11.27 &
  \multicolumn{1}{c|}{\cellcolor[HTML]{FFFFFF}91.14} &
  13.58 &
  \multicolumn{1}{c|}{\cellcolor[HTML]{FFFFFF}90.32} &
  9.61 \\   
\multicolumn{1}{|c|}{\multirow{-3}{*}{\cellcolor[HTML]{EA9999}\textbf{\begin{tabular}[c]{@{}c@{}}Mixtral-8x7B-\\ Instruct-v0.1\end{tabular}}}} &
  \textbf{F1/RL} &
  \multicolumn{1}{c|}{\cellcolor[HTML]{EFEFEF}\textit{\textbf{88.93}}} &
  25.28 &
  \multicolumn{1}{c|}{{\ul \textbf{90.37}}} &
  30.16 &
  \multicolumn{1}{c|}{\cellcolor[HTML]{EFEFEF}89.37} &
  26.85 &
  \multicolumn{1}{c|}{\cellcolor[HTML]{EFEFEF}\textit{\textbf{90.23}}} &
  29.6 &
  \multicolumn{1}{c|}{\cellcolor[HTML]{EFEFEF}88.22} &
  23.16 \\ \hline
\end{tabular}%
%}
\end{adjustbox}
\caption{Performance of LLMs on the SAMSum dataset with ZSL.}
\label{tab:table7}
\end{table}

According to the results, BERTScore and ROUGE metrics align well in most cases, with high BERT F1 scores in some prompts with most models. Often these high scores correspond to high ROUGE scores. Mixtral-8x7B-Instruct-v0.1 gets the highest value of BERT F1 with prompts\#2 and\#4, and Llama-2-13b-chat achieves the best Rouge-1 score with also prompts\#2 and\#4. Thus, there is a consistency between BERT and Rouge scores. This consistency reinforces the reliability of these metrics in evaluating summarization quality.

In terms of prompt effectiveness, prompt\#1 and prompt\#2, which are simpler and more direct, tend to result in higher ROUGE scores, suggesting that straightforward instructions may be more effective for summarization tasks. Prompts \#3 to \#5, which are more descriptive, show varied performance across different models. This indicates that while detailed prompts can be useful, they might not always yield the best results.

Finally, the overall performance of the models across different prompts shows that \textbf{\underline{Mistral-7B-Instruct-v0.1}} frequently achieves top scores across different prompts, indicating its robustness and adaptability in summarizing dialogues. \textbf{\underline{Llama-2-13b-chat}} also performs exceptionally well, especially with more direct prompts like Prompt\#2 and Prompt\#5, suggesting it is well-suited for concise summarization tasks.

\textbf{Arxiv results with ZSL:} ArXiv dataset is one of the popular datasets for representing text summarization with scientific papers. The dataset represents a pair of articles and their abstract. It is suitable for abstractive text summarization since the abstract is written by humans making it more coherent and comprehensive. The ArXiv dataset, introduced for testing abstractive summarization models on longer-form scientific documents, poses unique challenges compared to summarizing shorter documents. The challenge of long input documents caused LLMs' context length may be problematic in our model. So, our experiments are done on the documents of the dataset after trimming the documents to be suitable to context length of the employed models. Our analysis highlights how different prompts and models perform on this dataset, revealing insights into their strengths and weaknesses. The results of the are provided in Table \ref{tab:table8}.

% \begin{flushleft}
 Regarding prompt effectiveness, prompt\#2 consistently results in higher ROUGE scores across models, especially for \textbf{Llama-2-13b-chat} and \textbf{Mistral-7B-Instruct-v0.1}, suggesting it effectively guides the models to capture critical information. prompt\#3 shows strong performance, particularly in F1/RL scores, indicating that framing the task explicitly as summarizing helps models generate more accurate abstracts. prompt\#4 balances between BERT and ROUGE scores, showing it helps models produce concise and relevant summaries. In general, the result across multiple prompts highlights that while BERT scores are relatively consistent across models, ROUGE scores show more variation, particularly with prompt\#2 and prompt\#3. This highlights the different capabilities of models in capturing the essence of the source text when dealing with long text.

The overall performance of the models shows \textbf{Llama-2-7b-chat} achieves the highest scores for prompt\#2 and overall robust performance, indicating its capability in handling abstractive summarization tasks ef fectively. \textbf{Mistral-7B-Instruct-v0.1} demonstrates consistently high F1/RL scores, particularly with prompt\#3, making it a strong contender for generating precise and comprehensive abstracts. \textbf{Gemma-7b-it} generally exhibits lower performance, suggesting it might require further tuning or may not be as well-suited for long-form document summarization compared to other models. In conclusion, the variability in performance across different metrics and prompts highlights the complexity of the task. Also maybe the reason for these difference in results caused by the trimming process of the input document, which results in some models can’t understand the context well after trimming.

\begin{table}[]
\centering

\begin{adjustbox}{max width=\textwidth} 
\renewcommand{\arraystretch}{1.5}
%\resizebox{\columnwidth}{!}{%
\begin{tabular}{|cc|cc|cc|cc|cc|}
\hline
\multicolumn{2}{|c|}{Prompts} &
  \multicolumn{2}{c|}{\cellcolor[HTML]{C96969}\textbf{prompt\#1}} &
  \multicolumn{2}{c|}{\cellcolor[HTML]{C96969}\textbf{prompt\#2}} &
  \multicolumn{2}{c|}{\cellcolor[HTML]{C96969}\textbf{prompt\#3}} &
  \multicolumn{2}{c|}{\cellcolor[HTML]{C96969}\textbf{prompt\#4}} \\ \hline
\multicolumn{1}{|c|}{\textbf{Models}} &
  Metrics &
  \multicolumn{1}{c|}{\textbf{Bert}} &
  \textbf{Rouge} &
  \multicolumn{1}{c|}{\textbf{Bert}} &
  \textbf{Rouge} &
  \multicolumn{1}{c|}{\textbf{Bert}} &
  \textbf{Rouge} &
  \multicolumn{1}{c|}{\textbf{Bert}} &
  \textbf{Rouge} \\ \hline
  
\multicolumn{1}{|c|}{\cellcolor[HTML]{C27BA0}} &
  \cellcolor[HTML]{FFFFFF}\textbf{P/R1} &
  \multicolumn{1}{c|}{\cellcolor[HTML]{FFFFFF}{ {84.27}}} &
  \cellcolor[HTML]{FFFFFF}{ {25.72}} &
  \multicolumn{1}{c|}{\cellcolor[HTML]{FFFFFF}84.51} &
  \cellcolor[HTML]{FFFFFF}40.05 &
  \multicolumn{1}{c|}{\cellcolor[HTML]{FFFFFF}86.14} &
  \cellcolor[HTML]{FFFFFF}24.73 &
  \multicolumn{1}{c|}{\cellcolor[HTML]{FFFFFF}86.09} &
  \cellcolor[HTML]{FFFFFF}22.52 \\   
  
\multicolumn{1}{|c|}{\cellcolor[HTML]{C27BA0}} &
  \cellcolor[HTML]{EFEFEF}\textbf{R/R2} &
  \multicolumn{1}{c|}{\cellcolor[HTML]{EFEFEF}{ {77.81}}} &
  \cellcolor[HTML]{EFEFEF}{\ {15.76}} &
  \multicolumn{1}{c|}{\cellcolor[HTML]{EFEFEF}79.46} &
  \cellcolor[HTML]{EFEFEF}25.64 &
  \multicolumn{1}{c|}{\cellcolor[HTML]{EFEFEF}78.69} &
  \cellcolor[HTML]{EFEFEF}13.45 &
  \multicolumn{1}{c|}{\cellcolor[HTML]{EFEFEF}78.45} &
  \cellcolor[HTML]{EFEFEF}12.04 \\ 
  
\multicolumn{1}{|c|}{\multirow{-3}{*}{\cellcolor[HTML]{C27BA0}\textbf{gemma-7b-it}}} &
  \cellcolor[HTML]{FFFFFF}\textbf{F1/RL} &
  \multicolumn{1}{c|}{\cellcolor[HTML]{FFFFFF}{ {80.89}}} &
  \cellcolor[HTML]{FFFFFF}{ {18.61}} &
  \multicolumn{1}{c|}{\cellcolor[HTML]{FFFFFF}81.89} &
  \cellcolor[HTML]{FFFFFF}27.81 &
  \multicolumn{1}{c|}{\cellcolor[HTML]{FFFFFF}82.24} &
  \cellcolor[HTML]{FFFFFF}18.05 &
  \multicolumn{1}{c|}{\cellcolor[HTML]{FFFFFF}82.09} &
  \cellcolor[HTML]{FFFFFF}16.59 \\ \hline
  %4
\multicolumn{1}{|c|}{\cellcolor[HTML]{E6B8AF}} &
  \cellcolor[HTML]{EFEFEF}\textbf{P/R1} &
  \multicolumn{1}{c|}{\cellcolor[HTML]{EFEFEF}83.86} &
  \cellcolor[HTML]{EFEFEF}\textit{\textbf{39.61}} &
  \multicolumn{1}{c|}{\cellcolor[HTML]{EFEFEF}83.92} &
  {\ul\textbf{49.74}} &
  \multicolumn{1}{c|}{{\textbf{86.86}}} &
  \cellcolor[HTML]{EFEFEF}\textit{\textbf{38.25}} &
  \multicolumn{1}{c|}{\cellcolor[HTML]{EFEFEF}84.73} &
  \cellcolor[HTML]{EFEFEF}\textit{\textbf{36.19}} \\   
  %5
\multicolumn{1}{|c|}{\cellcolor[HTML]{E6B8AF}} &
  \cellcolor[HTML]{FFFFFF}\textbf{R/R2} &
  \multicolumn{1}{c|}{\cellcolor[HTML]{FFFFFF}79.56} &
  \cellcolor[HTML]{FFFFFF}21.52 &
  \multicolumn{1}{c|}{{\ul \textbf{80.41}}} &
 {\ul \textbf{32.47}} &
  \multicolumn{1}{c|}{\cellcolor[HTML]{FFFFFF}80.09} &
  \cellcolor[HTML]{FFFFFF}19.43 &
  \multicolumn{1}{c|}{\cellcolor[HTML]{FFFFFF}79.79} &
  \cellcolor[HTML]{FFFFFF}18.84 \\
  %6
\multicolumn{1}{|c|}{\multirow{-3}{*}{\cellcolor[HTML]{E6B8AF}\textbf{Llama-2-7b-chat}}} &
  \cellcolor[HTML]{EFEFEF}\textbf{F1/RL} &
  \multicolumn{1}{c|}{\cellcolor[HTML]{EFEFEF}\textit{\textbf{81.64}}} &
  \cellcolor[HTML]{EFEFEF}27.03 &
  \multicolumn{1}{c|}{\cellcolor[HTML]{EFEFEF}82.12} &
  {\ul \textbf{33.98}} &
  \multicolumn{1}{c|}{{\ul \textbf{83.33}}} &
  \cellcolor[HTML]{E7E6E6}24.25 &
  \multicolumn{1}{c|}{\cellcolor[HTML]{EFEFEF}82.17} &
  \cellcolor[HTML]{EFEFEF}24.13 \\ \hline
\multicolumn{1}{|c|}{\cellcolor[HTML]{70AD47}} &
  \cellcolor[HTML]{FFFFFF}\textbf{P/R1} &
  \multicolumn{1}{c|}{\cellcolor[HTML]{FFFFFF}84.84} &
  \cellcolor[HTML]{FFFFFF}\textit{\textbf{35.47}} &
  \multicolumn{1}{c|}{\cellcolor[HTML]{FFFFFF}84.21} &
  \cellcolor[HTML]{FFFFFF}\textit{\textbf{47.92}} &
  \multicolumn{1}{c|}{\cellcolor[HTML]{FFFFFF}84.94} &
  \cellcolor[HTML]{FFFFFF}37.87 &
  \multicolumn{1}{c|}{\cellcolor[HTML]{FFFFFF}85.89} &
  \cellcolor[HTML]{FFFFFF}{ \textbf{23.33}} \\   
\multicolumn{1}{|c|}{\cellcolor[HTML]{70AD47}} &
  \cellcolor[HTML]{EFEFEF}\textbf{R/R2} &
  \multicolumn{1}{c|}{\cellcolor[HTML]{EFEFEF}79.14} &
  \cellcolor[HTML]{EFEFEF}20.63 &
  \multicolumn{1}{c|}{\cellcolor[HTML]{EFEFEF}79.96} &
  \cellcolor[HTML]{EFEFEF}29.73 &
  \multicolumn{1}{c|}{\cellcolor[HTML]{EFEFEF}79.95} &
  \cellcolor[HTML]{EFEFEF}19.35 &
  \multicolumn{1}{c|}{\cellcolor[HTML]{EFEFEF}78.48} &
  \cellcolor[HTML]{EFEFEF}12.01 \\   
\multicolumn{1}{|c|}{\multirow{-3}{*}{\cellcolor[HTML]{70AD47}\textbf{Llama-2-13b-chat}}} &
  \cellcolor[HTML]{FFFFFF}\textbf{F1/RL} &
  \multicolumn{1}{c|}{\cellcolor[HTML]{FFFFFF}\textit{\textbf{81.88}}} &
  \cellcolor[HTML]{FFFFFF}24.48 &
  \multicolumn{1}{c|}{\cellcolor[HTML]{FFFFFF}82.02} &
  \cellcolor[HTML]{FFFFFF}32.43 &
  \multicolumn{1}{c|}{\cellcolor[HTML]{FFFFFF}82.36} &
  \cellcolor[HTML]{FFFFFF}23.93 &
  \multicolumn{1}{c|}{\cellcolor[HTML]{FFFFFF}82.01} &
  \cellcolor[HTML]{FFFFFF}15.65 \\ \hline
\multicolumn{1}{|c|}{\cellcolor[HTML]{FF9900}} &
  \cellcolor[HTML]{EFEFEF}\textbf{P/R1} &
  \multicolumn{1}{c|}{\cellcolor[HTML]{EFEFEF}84.39} &
  \cellcolor[HTML]{EFEFEF}\textit{\textbf{39.31}} &
  \multicolumn{1}{c|}{\cellcolor[HTML]{EFEFEF}84.26} &
  \cellcolor[HTML]{EFEFEF}35.48 &
  \multicolumn{1}{c|}{\cellcolor[HTML]{EFEFEF}85.33} &
  \cellcolor[HTML]{EFEFEF}\textit{\textbf{35.29}} &
  \multicolumn{1}{c|}{\cellcolor[HTML]{EFEFEF}85.4} &
  \cellcolor[HTML]{EFEFEF}25.64 \\   
  
\multicolumn{1}{|c|}{\cellcolor[HTML]{FF9900}} &
  \cellcolor[HTML]{FFFFFF}\textbf{R/R2} &
  \multicolumn{1}{c|}{\cellcolor[HTML]{FFFFFF}78.34} &
  \cellcolor[HTML]{FFFFFF}25.09 &
  \multicolumn{1}{c|}{\cellcolor[HTML]{FFFFFF}79.25} &
  \cellcolor[HTML]{FFFFFF}17.61 &
  \multicolumn{1}{c|}{\cellcolor[HTML]{FFFFFF}79.94} &
  \cellcolor[HTML]{FFFFFF}18.3 &
  \multicolumn{1}{c|}{\cellcolor[HTML]{FFFFFF}78.41} &
  \cellcolor[HTML]{FFFFFF}12.56 \\
  
\multicolumn{1}{|c|}{\multirow{-3}{*}{\cellcolor[HTML]{FF9900}\textbf{Llama-2-70b-chat}}} &
  \cellcolor[HTML]{EFEFEF}\textbf{F1/RL} &
  \multicolumn{1}{c|}{\cellcolor[HTML]{EFEFEF}\textit{\textbf{81.24}}} &
  \cellcolor[HTML]{EFEFEF}27.38 &
  \multicolumn{1}{c|}{\cellcolor[HTML]{EFEFEF}81.67} &
  \cellcolor[HTML]{EFEFEF}21.65 &
  \multicolumn{1}{c|}{\cellcolor[HTML]{EFEFEF}\textit{\textbf{82.54}}} &
  \cellcolor[HTML]{EFEFEF}22.85 &
  \multicolumn{1}{c|}{\cellcolor[HTML]{EFEFEF}81.74} &
  \cellcolor[HTML]{EFEFEF}16.7 \\ \hline
  
\multicolumn{1}{|c|}{\cellcolor[HTML]{4472C4}} &
  \cellcolor[HTML]{FFFFFF}\textbf{P/R1} &
  \multicolumn{1}{c|}{\cellcolor[HTML]{FFFFFF}85.29} &
  \cellcolor[HTML]{FFFFFF}\textit{\textbf{31.89}} &
  \multicolumn{1}{c|}{\cellcolor[HTML]{FFFFFF}85.61} &
  \cellcolor[HTML]{FFFFFF}36.64 &
  \multicolumn{1}{c|}{\cellcolor[HTML]{FFFFFF}86.86} &
  \cellcolor[HTML]{FFFFFF}30.59 &
  \multicolumn{1}{c|}{\cellcolor[HTML]{FFFFFF}86.26} &
  \cellcolor[HTML]{FFFFFF}29.47 \\   
  
\multicolumn{1}{|c|}{\cellcolor[HTML]{4472C4}} &
  \cellcolor[HTML]{EFEFEF}\textbf{R/R2} &
  \multicolumn{1}{c|}{\cellcolor[HTML]{EFEFEF}79.14} &
  \cellcolor[HTML]{EFEFEF}20.81 &
  \multicolumn{1}{c|}{\cellcolor[HTML]{EFEFEF}79.38} &
  \cellcolor[HTML]{EFEFEF}25.77 &
  \multicolumn{1}{c|}{\cellcolor[HTML]{EFEFEF}80.09} &
  \cellcolor[HTML]{EFEFEF}18.49 &
  \multicolumn{1}{c|}{\cellcolor[HTML]{EFEFEF}79.74} &
  \cellcolor[HTML]{EFEFEF}17.98 \\
  
\multicolumn{1}{|c|}{\multirow{-3}{*}{\cellcolor[HTML]{4472C4}\textbf{\begin{tabular}[c]{@{}c@{}}Mistral-7B-\\ Instruct-v0.1\end{tabular}}}} &
  \cellcolor[HTML]{FFFFFF}\textbf{F1/RL} &
  \multicolumn{1}{c|}{\cellcolor[HTML]{FFFFFF}\textit{\textbf{82.08}}} &
  \cellcolor[HTML]{FFFFFF}25.18 &
  \multicolumn{1}{c|}{\cellcolor[HTML]{FFFFFF}\textit{\textbf{82.37}}} &
  \cellcolor[HTML]{FFFFFF}30.05 &
  \multicolumn{1}{c|}{\cellcolor[HTML]{FFFFFF}83.33} &
  \cellcolor[HTML]{FFFFFF}23.98 &
  \multicolumn{1}{c|}{\cellcolor[HTML]{FFFFFF}\textit{\textbf{82.85}}} &
  \cellcolor[HTML]{FFFFFF}23.15 \\ \hline
  
\multicolumn{1}{|c|}{\cellcolor[HTML]{EA9999}} &
  \cellcolor[HTML]{EFEFEF}\textbf{P/R1} &
  \multicolumn{1}{c|}{\cellcolor[HTML]{EFEFEF}84.89} &
  \cellcolor[HTML]{E7E6E6}\textit{\textbf{35.45}} &
  \multicolumn{1}{c|}{\cellcolor[HTML]{EFEFEF}84.73} &
  \cellcolor[HTML]{EFEFEF}37.43 &
  \multicolumn{1}{c|}{\cellcolor[HTML]{EFEFEF}85.23} &
  \cellcolor[HTML]{EFEFEF}34.34 &
  \multicolumn{1}{c|}{\cellcolor[HTML]{EFEFEF}\textbf{85.28}} &
  \cellcolor[HTML]{EFEFEF}31.18 \\ 
  
\multicolumn{1}{|c|}{\cellcolor[HTML]{EA9999}} &
  \cellcolor[HTML]{FFFFFF}\textbf{R/R2} &
  \multicolumn{1}{c|}{\cellcolor[HTML]{FFFFFF}79.29} &
  \cellcolor[HTML]{FFFFFF}19.92 &
  \multicolumn{1}{c|}{\cellcolor[HTML]{FFFFFF}79.74} &
  \cellcolor[HTML]{FFFFFF}21.84 &
  \multicolumn{1}{c|}{\cellcolor[HTML]{FFFFFF}79.95} &
  \cellcolor[HTML]{FFFFFF}18.23 &
  \multicolumn{1}{c|}{\cellcolor[HTML]{FFFFFF}79.34} &
  \cellcolor[HTML]{FFFFFF}16.63 \\ 
  
\multicolumn{1}{|c|}{\multirow{-3}{*}{\cellcolor[HTML]{EA9999}\textbf{\begin{tabular}[c]{@{}c@{}}Mixtral-8x7B-\\ Instruct-v0.1\end{tabular}}}} &
  \cellcolor[HTML]{EFEFEF}\textbf{F1/RL} &
  \multicolumn{1}{c|}{\cellcolor[HTML]{EFEFEF}\textit{\textbf{81.98}}} &
  \cellcolor[HTML]{EFEFEF}25.39 &
  \multicolumn{1}{c|}{\cellcolor[HTML]{EFEFEF}\textbf{82.15}} &
  \cellcolor[HTML]{EFEFEF}27.07 &
  \multicolumn{1}{c|}{\cellcolor[HTML]{EFEFEF}82.49} &
  \cellcolor[HTML]{EFEFEF}24.39 &
  \multicolumn{1}{c|}{\cellcolor[HTML]{EFEFEF}\textit{\textbf{82.19}}} &
  \cellcolor[HTML]{EFEFEF}22.06 \\ \hline
\end{tabular}%
%}
\end{adjustbox}
\caption{Performance of LLMs on the Arxiv dataset with ZSL.}
\label{tab:table8}
\end{table}

\subsubsection{ICL Results}
By providing a series of example documents and their corresponding summaries, models learn to generate accurate and concise summaries for new input documents. Using the format of the prompts in Figure \ref{fig:fig1}, ICL is applied to prompt LLMs for generating summaries of each dataset. The results of each data set are discussed as follows.

\textbf{\underline{CNN/DM results with ICL:}} Table \ref{tab:table9} illustrates the performance of the Models on the CNN/DM dataset using a different number of demonstrations in ICL. For BERT Metric, across all models, the Precision, Recall, and F1-Score are relatively consistent through the different number of demonstrations, with some variations when increasing the number of examples. In addition, adding more examples (from 1 to 3, 5 and 7) results in a slight improvement in these scores, indicating that providing more examples helps the models understand the summarization task better. In other hand, ROUGE metrics show more noticeable differences. \textbf{Mistral-7B-Instruct-v0.1}, \textbf{Mixtral-8x7B-Instruct-v0.1} and \textbf{Llama-2-70b-chat} models get the most significant improvements in ROUGE scores with some difference when increase number of demonstrations. Additionally, \textbf{Llama-2-70b-chat} model stands out as the best performer with 7-Shots setup where it achieves the highest scores across Rouge and BERT metrics and \textbf{Mixtral-8x7B-Instruct-v0.1} gets good results across Rouge and BERT metrics.  In contrast, \textbf{gemma-7b-it} shows the least improvement with additional examples, indicating a possible limitation in how it leverages extra context for summarization.

\begin{table}[h]
\centering
\begin{adjustbox}{max width=\textwidth} 
\renewcommand{\arraystretch}{1.5}
%\resizebox{\columnwidth}{!}{%
\begin{tabular}{|cc|cc|cc|cc|cc|}
\hline
\multicolumn{2}{|c|}{Prompts} &
  \multicolumn{2}{c|}{\cellcolor[HTML]{C96969}\textbf{1-Shot}} &
  \multicolumn{2}{c|}{\cellcolor[HTML]{C96969}\textbf{3-Shots}} &
  \multicolumn{2}{c|}{\cellcolor[HTML]{C96969}\textbf{5-Shots}} &
  \multicolumn{2}{c|}{\cellcolor[HTML]{C96969}\textbf{7-Shots}} \\ \hline
\multicolumn{1}{|c|}{\textbf{Models}} &
  Metrics &
  \multicolumn{1}{c|}{\textbf{Bert}} &
  \textbf{Rouge} &
  \multicolumn{1}{c|}{\textbf{Bert}} &
  \textbf{Rouge} &
  \multicolumn{1}{c|}{\textbf{Bert}} &
  \textbf{Rouge} &
  \multicolumn{1}{c|}{\textbf{Bert}} &
  \textbf{Rouge} \\ \hline
\rowcolor[HTML]{FFFFFF} 
\multicolumn{1}{|c|}{\cellcolor[HTML]{C27BA0}} &
  \textbf{P/R1} &
  \multicolumn{1}{c|}{\cellcolor[HTML]{FFFFFF}\textbf{84.04}} &
  \textbf{21.78} &
  \multicolumn{1}{c|}{\cellcolor[HTML]{FFFFFF}87.92} &
  33.7 &
  \multicolumn{1}{c|}{\cellcolor[HTML]{FFFFFF}85.4} &
  27.55 &
  \multicolumn{1}{c|}{\cellcolor[HTML]{FFFFFF}85.19} &
  28.36 \\  
\rowcolor[HTML]{EFEFEF} 
\multicolumn{1}{|c|}{\cellcolor[HTML]{C27BA0}} &
  \textbf{R/R2} &
  \multicolumn{1}{c|}{\cellcolor[HTML]{EFEFEF}\textbf{82.27}} &
  \textbf{7.75} &
  \multicolumn{1}{c|}{\cellcolor[HTML]{EFEFEF}86.43} &
  12.37 &
  \multicolumn{1}{c|}{\cellcolor[HTML]{EFEFEF}85.59} &
  9.36 &
  \multicolumn{1}{c|}{\cellcolor[HTML]{EFEFEF}87.3} &
  10.68 \\  
\rowcolor[HTML]{FFFFFF} 
\multicolumn{1}{|c|}{\multirow{-3}{*}{\cellcolor[HTML]{C27BA0}\textbf{gemma-7b-it}}} &
  \textbf{F1/RL} &
  \multicolumn{1}{c|}{\cellcolor[HTML]{FFFFFF}\textbf{83.14}} &
  \textbf{12.6} &
  \multicolumn{1}{c|}{\cellcolor[HTML]{FFFFFF}87.16} &
  22.06 &
  \multicolumn{1}{c|}{\cellcolor[HTML]{FFFFFF}85.47} &
  17.96 &
  \multicolumn{1}{c|}{\cellcolor[HTML]{FFFFFF}86.22} &
  18.55 \\ \hline
\rowcolor[HTML]{EFEFEF} 
\multicolumn{1}{|c|}{\cellcolor[HTML]{E6B8AF}} &
  \textbf{P/R1} &
  \multicolumn{1}{c|}{\cellcolor[HTML]{EFEFEF}86.91} &
  \textit{\textbf{35.92}} &
  \multicolumn{1}{c|}{\cellcolor[HTML]{EFEFEF}86.14} &
  \textit{\textbf{32.58}} &
  \multicolumn{1}{c|}{\cellcolor[HTML]{EFEFEF}87.27} &
  \textit{\textbf{34.35}} &
  \multicolumn{1}{c|}{\cellcolor[HTML]{EFEFEF}88.19} &
  \textit{\textbf{39.45}} \\  
\rowcolor[HTML]{FFFFFF} 
\multicolumn{1}{|c|}{\cellcolor[HTML]{E6B8AF}} &
  \textbf{R/R2} &
  \multicolumn{1}{c|}{\cellcolor[HTML]{FFFFFF}87.52} &
  13.55 &
  \multicolumn{1}{c|}{\cellcolor[HTML]{FFFFFF}88.08} &
  12.77 &
  \multicolumn{1}{c|}{\cellcolor[HTML]{FFFFFF}87.41} &
  14.62 &
  \multicolumn{1}{c|}{\cellcolor[HTML]{FFFFFF}88.59} &
  16.49 \\  
\rowcolor[HTML]{EFEFEF} 
\multicolumn{1}{|c|}{\multirow{-3}{*}{\cellcolor[HTML]{E6B8AF}\textbf{Llama-2-7b-chat}}} &
  \textbf{F1/RL} &
  \multicolumn{1}{c|}{\cellcolor[HTML]{EFEFEF}\textbf{87.2}} &
  23.01 &
  \multicolumn{1}{c|}{\cellcolor[HTML]{EFEFEF}87.09} &
  \textbf{20.85} &
  \multicolumn{1}{c|}{\cellcolor[HTML]{EFEFEF}\textit{\textbf{87.29}}} &
  23.16 &
  \multicolumn{1}{c|}{\cellcolor[HTML]{EFEFEF}88.37} &
  26.25 \\ \hline
\rowcolor[HTML]{FFFFFF} 
\multicolumn{1}{|c|}{\cellcolor[HTML]{70AD47}} &
  \textbf{P/R1} &
  \multicolumn{1}{c|}{\cellcolor[HTML]{FFFFFF}87.54} &
  \textit{\textbf{36.31}} &
  \multicolumn{1}{c|}{\cellcolor[HTML]{FFFFFF}86.15} &
  \textit{\textbf{32.37}} &
  \multicolumn{1}{c|}{\cellcolor[HTML]{FFFFFF}87.41} &
  33.95 &
  \multicolumn{1}{c|}{\cellcolor[HTML]{FFFFFF}88.63} &
  \textbf{39.43} \\  
\rowcolor[HTML]{EFEFEF} 
\multicolumn{1}{|c|}{\cellcolor[HTML]{70AD47}} &
  \textbf{R/R2} &
  \multicolumn{1}{c|}{\cellcolor[HTML]{EFEFEF}87.84} &
  13.71 &
  \multicolumn{1}{c|}{\cellcolor[HTML]{EFEFEF}87.94} &
  12.63 &
  \multicolumn{1}{c|}{\cellcolor[HTML]{EFEFEF}87.16} &
  14.24 &
  \multicolumn{1}{c|}{\cellcolor[HTML]{EFEFEF}89.12} &
  16.48 \\  
\rowcolor[HTML]{FFFFFF} 
\multicolumn{1}{|c|}{\multirow{-3}{*}{\cellcolor[HTML]{70AD47}\textbf{Llama-2-13b-chat}}} &
  \textbf{F1/RL} &
  \multicolumn{1}{c|}{\cellcolor[HTML]{FFFFFF}\textbf{87.67}} &
  23.11 &
  \multicolumn{1}{c|}{\cellcolor[HTML]{FFFFFF}87.02} &
  20.7 &
  \multicolumn{1}{c|}{\cellcolor[HTML]{FFFFFF}87.24} &
  22.62 &
  \multicolumn{1}{c|}{\cellcolor[HTML]{FFFFFF}88.86} &
  26.22 \\ \hline
\rowcolor[HTML]{EFEFEF} 
\multicolumn{1}{|c|}{\cellcolor[HTML]{FF9900}} &
  \textbf{P/R1} &
  \multicolumn{1}{c|}{\cellcolor[HTML]{EFEFEF}\textbf{87.63}} &
  \textbf{38.26} &
  \multicolumn{1}{c|}{\cellcolor[HTML]{EFEFEF}85.31} &
  29.8 &
  \multicolumn{1}{c|}{\cellcolor[HTML]{EFEFEF}87.09} &
  \textbf{33.11} &
  \multicolumn{1}{c|}{\cellcolor[HTML]{EFEFEF}{\ul \textbf{88.97}}} &
  {\ul \textbf{40.98}} \\  
\rowcolor[HTML]{FFFFFF} 
\multicolumn{1}{|c|}{\cellcolor[HTML]{FF9900}} &
  \textbf{R/R2} &
  \multicolumn{1}{c|}{\cellcolor[HTML]{FFFFFF}\textbf{87.68}} &
  \textbf{14.51} &
  \multicolumn{1}{c|}{\cellcolor[HTML]{FFFFFF}88.03} &
  11.91 &
  \multicolumn{1}{c|}{\cellcolor[HTML]{FFFFFF}87.62} &
  14.11 &
  \multicolumn{1}{c|}{\cellcolor[HTML]{FFFFFF}88.45} &
  {\ul \textbf{17.23}} \\  
\rowcolor[HTML]{EFEFEF} 
\multicolumn{1}{|c|}{\multirow{-3}{*}{\cellcolor[HTML]{FF9900}\textbf{Llama-2-70b-chat}}} &
  \textbf{F1/RL} &
  \multicolumn{1}{c|}{\cellcolor[HTML]{EFEFEF}\textbf{87.65}} &
  \textbf{24.22} &
  \multicolumn{1}{c|}{\cellcolor[HTML]{EFEFEF}86.63} &
  19.23 &
  \multicolumn{1}{c|}{\cellcolor[HTML]{EFEFEF}\textbf{87.31}} &
  22.32 &
  \multicolumn{1}{c|}{\cellcolor[HTML]{EFEFEF}{\ul \textbf{88.7}}} &
  {\ul \textbf{27.52}} \\ \hline
\rowcolor[HTML]{FFFFFF} 
\multicolumn{1}{|c|}{\cellcolor[HTML]{4472C4}} &
  \textbf{P/R1} &
  \multicolumn{1}{c|}{\cellcolor[HTML]{FFFFFF}85.87} &
  \textbf{32.47} &
  \multicolumn{1}{c|}{\cellcolor[HTML]{FFFFFF}87.4} &
  36.02 &
  \multicolumn{1}{c|}{\cellcolor[HTML]{FFFFFF}87.45} &
  35.9 &
  \multicolumn{1}{c|}{\cellcolor[HTML]{FFFFFF}87.55} &
  34.41 \\  
\rowcolor[HTML]{EFEFEF} 
\multicolumn{1}{|c|}{\cellcolor[HTML]{4472C4}} &
  \textbf{R/R2} &
  \multicolumn{1}{c|}{\cellcolor[HTML]{EFEFEF}88.26} &
  13.59 &
  \multicolumn{1}{c|}{\cellcolor[HTML]{EFEFEF}87.6} &
  15.7 &
  \multicolumn{1}{c|}{\cellcolor[HTML]{EFEFEF}88.21} &
  14.79 &
  \multicolumn{1}{c|}{\cellcolor[HTML]{EFEFEF}87.22} &
  14.52 \\  

\multicolumn{1}{|c|}{\multirow{-3}{*}{\cellcolor[HTML]{4472C4}\textbf{\begin{tabular}[c]{@{}c@{}}Mistral-7B-\\ Instruct-v0.1\end{tabular}}}} &
  \textbf{F1/RL} &
  \multicolumn{1}{c|}{\cellcolor[HTML]{FFFFFF}\textbf{87.02}} &
  21.19 &
  \multicolumn{1}{c|}{\cellcolor[HTML]{FFFFFF}\textbf{87.47}} &
  23.8 &
  \multicolumn{1}{c|}{\cellcolor[HTML]{FFFFFF}87.81} &
  23.7 &
  \multicolumn{1}{c|}{\cellcolor[HTML]{FFFFFF}\textbf{87.36}} &
  23.26 \\ \hline
\rowcolor[HTML]{EFEFEF} 
\multicolumn{1}{|c|}{\cellcolor[HTML]{EA9999}} &
  \textbf{P/R1} &
  \multicolumn{1}{c|}{\cellcolor[HTML]{EFEFEF}86.75} &
  \textbf{34.34} &
  \multicolumn{1}{c|}{\cellcolor[HTML]{EFEFEF}\textbf{88.03}} &
  \textbf{37.53} &
  \multicolumn{1}{c|}{\cellcolor[HTML]{EFEFEF}\textbf{87.59}} &
  \textbf{37.12} &
  \multicolumn{1}{c|}{\cellcolor[HTML]{EFEFEF}\textbf{87.37}} &
  36.6 \\  
\rowcolor[HTML]{FFFFFF} 
\multicolumn{1}{|c|}{\cellcolor[HTML]{EA9999}} &
  \textbf{R/R2} &
  \multicolumn{1}{c|}{\cellcolor[HTML]{FFFFFF}86.31} &
  12.74 &
  \multicolumn{1}{c|}{\cellcolor[HTML]{FFFFFF}\textbf{88.9}} &
  \textbf{16.01} &
  \multicolumn{1}{c|}{\cellcolor[HTML]{FFFFFF}\textbf{88.95}} &
  \textbf{15.49} &
  \multicolumn{1}{c|}{\cellcolor[HTML]{FFFFFF}{\ul \textbf{89.52}}} &
  15.74 \\  

\multicolumn{1}{|c|}{\multirow{-3}{*}{\cellcolor[HTML]{EA9999}\textbf{\begin{tabular}[c]{@{}c@{}}Mixtral-8x7B-\\ Instruct-v0.1\end{tabular}}}} &
  \textbf{F1/RL} &
  \multicolumn{1}{c|}{\cellcolor[HTML]{EFEFEF}\textbf{86.51}} &
  21.57 &
  \multicolumn{1}{c|}{\cellcolor[HTML]{EFEFEF}\textbf{88.44}} &
  \textbf{24.79} &
  \multicolumn{1}{c|}{\cellcolor[HTML]{EFEFEF}\textbf{88.24}} &
  \textbf{24.58} &
  \multicolumn{1}{c|}{\cellcolor[HTML]{EFEFEF}\textit{\textbf{88.41}}} &
  24.25 \\ \hline
\end{tabular}%
%}
\end{adjustbox}
\caption{Performance of LLMs on CNN/DM dataset, with ICL.}
\label{tab:table9}
\end{table}

{\ul\textbf{NewsRoom results with ICL:} }Table \ref{tab:table10}  shows the results of employing ICL with NewsRoom dataset. According to the results, \textbf{Mixtral-8x7B-Instruct-v0.1} model learned the context of the data better than others and provided the model with 5 examples to get some little performance in both BERTScore and Rouge Score. Rouge metrics show more variability than BERT metrics. The three Llama models show good performances in 7-shots but slightly lower in other examples compared to \textbf{Mixtral-8x7B-Instruct-v0.1}. In conclusion, the results show that an increase in demonstrations generally in most models enhances the performance.

\begin{table}[]
\centering
\begin{adjustbox}{max width=\textwidth} 
\renewcommand{\arraystretch}{1.5}
%\resizebox{\columnwidth}{!}{%
\begin{tabular}{|cc|cc|cc|cc|cc|}
\hline
\multicolumn{2}{|c|}{Prompts} &
  \multicolumn{2}{c|}{\cellcolor[HTML]{C96969}\textbf{1-Shot}} &
  \multicolumn{2}{c|}{\cellcolor[HTML]{C96969}\textbf{3-Shots}} &
  \multicolumn{2}{c|}{\cellcolor[HTML]{C96969}\textbf{5-Shots}} &
  \multicolumn{2}{c|}{\cellcolor[HTML]{C96969}\textbf{7-Shots}} \\ \hline
\multicolumn{1}{|c|}{\textbf{Models}} &
  Metrics &
  \multicolumn{1}{c|}{\textbf{Bert}} &
  \textbf{Rouge} &
  \multicolumn{1}{c|}{\textbf{Bert}} &
  \textbf{Rouge} &
  \multicolumn{1}{c|}{\textbf{Bert}} &
  \textbf{Rouge} &
  \multicolumn{1}{c|}{\textbf{Bert}} &
  \textbf{Rouge} \\ \hline
\rowcolor[HTML]{FFFFFF} 
\multicolumn{1}{|c|}{\cellcolor[HTML]{C27BA0}} &
  \textbf{P/R1} &
  \multicolumn{1}{c|}{\cellcolor[HTML]{FFFFFF}84.39} &
  13.11 &
  \multicolumn{1}{c|}{\cellcolor[HTML]{FFFFFF}86.4} &
  20.28 &
  \multicolumn{1}{c|}{\cellcolor[HTML]{FFFFFF}85.35} &
  18 &
  \multicolumn{1}{c|}{\cellcolor[HTML]{FFFFFF}80.68} &
  10.71 \\   
\rowcolor[HTML]{EFEFEF} 
\multicolumn{1}{|c|}{\cellcolor[HTML]{C27BA0}} &
  \textbf{R/R2} &
  \multicolumn{1}{c|}{\cellcolor[HTML]{EFEFEF}84.05} &
  3.01 &
  \multicolumn{1}{c|}{\cellcolor[HTML]{EFEFEF}86.6} &
  5.81 &
  \multicolumn{1}{c|}{\cellcolor[HTML]{EFEFEF}87.06} &
  5.64 &
  \multicolumn{1}{c|}{\cellcolor[HTML]{EFEFEF}86.66} &
  3.71 \\   
\rowcolor[HTML]{FFFFFF} 
\multicolumn{1}{|c|}{\multirow{-3}{*}{\cellcolor[HTML]{C27BA0}\textbf{gemma-7b-it}}} &
  \textbf{F1/RL} &
  \multicolumn{1}{c|}{\cellcolor[HTML]{FFFFFF}84.2} &
  11.13 &
  \multicolumn{1}{c|}{\cellcolor[HTML]{FFFFFF}86.48} &
  15.9 &
  \multicolumn{1}{c|}{\cellcolor[HTML]{FFFFFF}86.18} &
  13.76 &
  \multicolumn{1}{c|}{\cellcolor[HTML]{FFFFFF}83.53} &
  8.14 \\ \hline
\rowcolor[HTML]{EFEFEF} 
\multicolumn{1}{|c|}{\cellcolor[HTML]{E6B8AF}} &
  \textbf{P/R1} &
  \multicolumn{1}{c|}{\cellcolor[HTML]{EFEFEF}82.35} &
  \textit{16.56} &
  \multicolumn{1}{c|}{\cellcolor[HTML]{EFEFEF}85.05} &
  21.58 &
  \multicolumn{1}{c|}{\cellcolor[HTML]{EFEFEF}\textbf{86.55}} &
  20.86 &
  \multicolumn{1}{c|}{\cellcolor[HTML]{EFEFEF}86.25} &
  \textbf{26.64} \\   
\rowcolor[HTML]{FFFFFF} 
\multicolumn{1}{|c|}{\cellcolor[HTML]{E6B8AF}} &
  \textbf{R/R2} &
  \multicolumn{1}{c|}{\cellcolor[HTML]{FFFFFF}85.54} &
  4.99 &
  \multicolumn{1}{c|}{\cellcolor[HTML]{FFFFFF}88.24} &
  7.09 &
  \multicolumn{1}{c|}{\cellcolor[HTML]{FFFFFF}85.75} &
  6.16 &
  \multicolumn{1}{c|}{\cellcolor[HTML]{FFFFFF}88.36} &
  \textbf{10.43} \\   
\rowcolor[HTML]{EFEFEF} 
\multicolumn{1}{|c|}{\multirow{-3}{*}{\cellcolor[HTML]{E6B8AF}\textbf{Llama-2-7b-chat}}} &
  \textbf{F1/RL} &
  \multicolumn{1}{c|}{\cellcolor[HTML]{EFEFEF}83.9} &
  13.58 &
  \multicolumn{1}{c|}{\cellcolor[HTML]{EFEFEF}86.59} &
  16.11 &
  \multicolumn{1}{c|}{\cellcolor[HTML]{EFEFEF}86.13} &
  17.24 &
  \multicolumn{1}{c|}{\cellcolor[HTML]{EFEFEF}87.27} &
  \textbf{21.19} \\ \hline
\rowcolor[HTML]{FFFFFF} 
\multicolumn{1}{|c|}{\cellcolor[HTML]{70AD47}} &
  \textbf{P/R1} &
  \multicolumn{1}{c|}{\cellcolor[HTML]{FFFFFF}82.25} &
  \textit{16.81} &
  \multicolumn{1}{c|}{\cellcolor[HTML]{FFFFFF}84.74} &
  \textit{\textbf{22}} &
  \multicolumn{1}{c|}{\cellcolor[HTML]{FFFFFF}83.56} &
  23.01 &
  \multicolumn{1}{c|}{\cellcolor[HTML]{FFFFFF}86.88} &
  26.3 \\   
\rowcolor[HTML]{EFEFEF} 
\multicolumn{1}{|c|}{\cellcolor[HTML]{70AD47}} &
  \textbf{R/R2} &
  \multicolumn{1}{c|}{\cellcolor[HTML]{EFEFEF}85.46} &
  4.95 &
  \multicolumn{1}{c|}{\cellcolor[HTML]{EFEFEF}87.76} &
  7.17 &
  \multicolumn{1}{c|}{\cellcolor[HTML]{EFEFEF}84.75} &
  9.32 &
  \multicolumn{1}{c|}{\cellcolor[HTML]{EFEFEF}88.75} &
  10.01 \\   
\rowcolor[HTML]{FFFFFF} 
\multicolumn{1}{|c|}{\multirow{-3}{*}{\cellcolor[HTML]{70AD47}\textbf{Llama-2-13b-chat}}} &
  \textbf{F1/RL} &
  \multicolumn{1}{c|}{\cellcolor[HTML]{FFFFFF}83.81} &
  13.44 &
  \multicolumn{1}{c|}{\cellcolor[HTML]{FFFFFF}86.21} &
  16.35 &
  \multicolumn{1}{c|}{\cellcolor[HTML]{FFFFFF}84.13} &
  18.33 &
  \multicolumn{1}{c|}{\cellcolor[HTML]{FFFFFF}87.78} &
  20.77 \\ \hline
\rowcolor[HTML]{EFEFEF} 
\multicolumn{1}{|c|}{\cellcolor[HTML]{FF9900}} &
  \textbf{P/R1} &
  \multicolumn{1}{c|}{\cellcolor[HTML]{EFEFEF}82.65} &
  17.32 &
  \multicolumn{1}{c|}{\cellcolor[HTML]{EFEFEF}85.26} &
  20.92 &
  \multicolumn{1}{c|}{\cellcolor[HTML]{EFEFEF}84.5} &
  22.62 &
  \multicolumn{1}{c|}{\cellcolor[HTML]{EFEFEF}87.27} &
  26.04 \\   
\rowcolor[HTML]{FFFFFF} 
\multicolumn{1}{|c|}{\cellcolor[HTML]{FF9900}} &
  \textbf{R/R2} &
  \multicolumn{1}{c|}{\cellcolor[HTML]{FFFFFF}85.82} &
  5.23 &
  \multicolumn{1}{c|}{\cellcolor[HTML]{FFFFFF}88.05} &
  6.62 &
  \multicolumn{1}{c|}{\cellcolor[HTML]{FFFFFF}86.3} &
  7.65 &
  \multicolumn{1}{c|}{\cellcolor[HTML]{FFFFFF}88.49} &
  9.23 \\   
\rowcolor[HTML]{EFEFEF} 
\multicolumn{1}{|c|}{\multirow{-3}{*}{\cellcolor[HTML]{FF9900}\textbf{Llama-2-70b-chat}}} &
  \textbf{F1/RL} &
  \multicolumn{1}{c|}{\cellcolor[HTML]{EFEFEF}84.19} &
  13.83 &
  \multicolumn{1}{c|}{\cellcolor[HTML]{EFEFEF}86.61} &
  15.62 &
  \multicolumn{1}{c|}{\cellcolor[HTML]{EFEFEF}85.36} &
  17.49 &
  \multicolumn{1}{c|}{\cellcolor[HTML]{EFEFEF}87.86} &
  20.18 \\ \hline
\rowcolor[HTML]{FFFFFF} 
\multicolumn{1}{|c|}{\cellcolor[HTML]{4472C4}} &
  \textbf{P/R1} &
  \multicolumn{1}{c|}{\cellcolor[HTML]{FFFFFF}\textbf{84.96}} &
  19.1 &
  \multicolumn{1}{c|}{\cellcolor[HTML]{FFFFFF}85.75} &
  23.38 &
  \multicolumn{1}{c|}{\cellcolor[HTML]{FFFFFF}86.17} &
  23.57 &
  \multicolumn{1}{c|}{\cellcolor[HTML]{FFFFFF}87.18} &
  22.84 \\   
\rowcolor[HTML]{EFEFEF} 
\multicolumn{1}{|c|}{\cellcolor[HTML]{4472C4}} &
  \textbf{R/R2} &
  \multicolumn{1}{c|}{\cellcolor[HTML]{EFEFEF}\textbf{86.63}} &
  6.61 &
  \multicolumn{1}{c|}{\cellcolor[HTML]{EFEFEF}87.39} &
  8.31 &
  \multicolumn{1}{c|}{\cellcolor[HTML]{EFEFEF}87.81} &
  8.75 &
  \multicolumn{1}{c|}{\cellcolor[HTML]{EFEFEF}86.02} &
  8.35 \\   

\multicolumn{1}{|c|}{\multirow{-3}{*}{\cellcolor[HTML]{4472C4}\textbf{\begin{tabular}[c]{@{}c@{}}Mistral-7B-\\ Instruct-v0.1\end{tabular}}}} &
  \textbf{F1/RL} &
  \multicolumn{1}{c|}{\cellcolor[HTML]{FFFFFF}\textbf{85.75}} &
  15.02 &
  \multicolumn{1}{c|}{\cellcolor[HTML]{FFFFFF}86.54} &
  17.86 &
  \multicolumn{1}{c|}{\cellcolor[HTML]{FFFFFF}86.96} &
  18.4 &
  \multicolumn{1}{c|}{\cellcolor[HTML]{FFFFFF}86.57} &
  19.07 \\ \hline
\rowcolor[HTML]{EFEFEF} 
\multicolumn{1}{|c|}{\cellcolor[HTML]{EA9999}} &
  \textbf{P/R1} &
  \multicolumn{1}{c|}{\cellcolor[HTML]{EFEFEF}84.76} &
  \textbf{20.61} &
  \multicolumn{1}{c|}{\cellcolor[HTML]{EFEFEF}\textbf{86.33}} &
  \textbf{23.73} &
  \multicolumn{1}{c|}{\cellcolor[HTML]{EFEFEF}85.87} &
  { \textbf{24.28}} &
  \multicolumn{1}{c|}{\cellcolor[HTML]{EFEFEF}{ \textbf{83.96}}} &
  21.05 \\   
\rowcolor[HTML]{FFFFFF} 
\multicolumn{1}{|c|}{\cellcolor[HTML]{EA9999}} &
  \textbf{R/R2} &
  \multicolumn{1}{c|}{\cellcolor[HTML]{FFFFFF}86.28} &
  \textbf{6.19} &
  \multicolumn{1}{c|}{\cellcolor[HTML]{FFFFFF}\textbf{88.58}} &
  \textbf{7.96} &
  \multicolumn{1}{c|}{\cellcolor[HTML]{FFFFFF}\textbf{88.4}} &
  { \textbf{9.15}} &
  \multicolumn{1}{c|}{\cellcolor[HTML]{FFFFFF}\textbf{87.46}} &
  7.99 \\   

\multicolumn{1}{|c|}{\multirow{-3}{*}{\cellcolor[HTML]{EA9999}\textbf{\begin{tabular}[c]{@{}c@{}}Mixtral-8x7B-\\ Instruct-v0.1\end{tabular}}}} &
  \textbf{F1/RL} &
  \multicolumn{1}{c|}{\cellcolor[HTML]{EFEFEF}85.5} &
  \textbf{16.27} &
  \multicolumn{1}{c|}{\cellcolor[HTML]{EFEFEF}\textbf{87.42}} &
  \textbf{18.03} &
  \multicolumn{1}{c|}{\cellcolor[HTML]{EFEFEF}\textbf{87.09}} &
  { \textbf{18.34}} &
  \multicolumn{1}{c|}{\cellcolor[HTML]{EFEFEF}{ \textit{\textbf{85.64}}}} &
  16.28 \\ \hline
\end{tabular}%
%}
\end{adjustbox}
\caption{Performance of LLMs on the NewsRoom dataset with ICL. }
\label{tab:table10}
\end{table}

{\ul\textbf{SAMSum results With ICL:}} when working with ICL in SAMSum dataset, Table \ref{tab:table11}  showed that \textbf{Mixtral-8x7B-Instruct-v0.1} got the best results in terms of BERTScore and Rouge score with seven examples to learn from which means it learned the context of the dialogues better than others. The more examples given to the model the more the result improved, Table \ref{tab:table11} shows that the results improved when increased the demonstrations in most models.

\begin{table}[]
\centering
\begin{adjustbox}{max width=\textwidth} 
\renewcommand{\arraystretch}{1.5}
%\resizebox{\columnwidth}{!}{%
\begin{tabular}{|cc|cc|cc|cc|cc|}
\hline
\multicolumn{2}{|c|}{Prompts} &
  \multicolumn{2}{c|}{\cellcolor[HTML]{C96969}\textbf{1-Shot}} &
  \multicolumn{2}{c|}{\cellcolor[HTML]{C96969}\textbf{3-Shots}} &
  \multicolumn{2}{c|}{\cellcolor[HTML]{C96969}\textbf{5-Shots}} &
  \multicolumn{2}{c|}{\cellcolor[HTML]{C96969}\textbf{7-Shots}} \\ \hline
\multicolumn{1}{|c|}{\textbf{Models}} &
  Metrics &
  \multicolumn{1}{c|}{\textbf{Bert}} &
  \textbf{Rouge} &
  \multicolumn{1}{c|}{\textbf{Bert}} &
  \textbf{Rouge} &
  \multicolumn{1}{c|}{\textbf{Bert}} &
  \textbf{Rouge} &
  \multicolumn{1}{c|}{\textbf{Bert}} &
  \textbf{Rouge} \\ \hline
\rowcolor[HTML]{FFFFFF} 
\multicolumn{1}{|c|}{\cellcolor[HTML]{C27BA0}} &
  \textbf{P/R1} &
  \multicolumn{1}{c|}{\cellcolor[HTML]{FFFFFF}84.35} &
  8.58 &
  \multicolumn{1}{c|}{\cellcolor[HTML]{FFFFFF}89.44} &
  34.21 &
  \multicolumn{1}{c|}{\cellcolor[HTML]{FFFFFF}90.31} &
  36.09 &
  \multicolumn{1}{c|}{\cellcolor[HTML]{FFFFFF}84.2} &
  23.34 \\   
\rowcolor[HTML]{EFEFEF} 
\multicolumn{1}{|c|}{\cellcolor[HTML]{C27BA0}} &
  \textbf{R/R2} &
  \multicolumn{1}{c|}{\cellcolor[HTML]{EFEFEF}85.6} &
  3.43 &
  \multicolumn{1}{c|}{\cellcolor[HTML]{EFEFEF}89.17} &
  11.64 &
  \multicolumn{1}{c|}{\cellcolor[HTML]{EFEFEF}89.71} &
  12.48 &
  \multicolumn{1}{c|}{\cellcolor[HTML]{EFEFEF}89.57} &
  8.14 \\   
\rowcolor[HTML]{FFFFFF} 
\multicolumn{1}{|c|}{\multirow{-3}{*}{\cellcolor[HTML]{C27BA0}\textbf{gemma-7b-it}}} &
  \textbf{F1/RL} &
  \multicolumn{1}{c|}{\cellcolor[HTML]{FFFFFF}84.97} &
  6.71 &
  \multicolumn{1}{c|}{\cellcolor[HTML]{FFFFFF}89.29} &
  27.08 &
  \multicolumn{1}{c|}{\cellcolor[HTML]{FFFFFF}89.99} &
  28.78 &
  \multicolumn{1}{c|}{\cellcolor[HTML]{FFFFFF}86.77} &
  17.44 \\ \hline
\rowcolor[HTML]{EFEFEF} 
\multicolumn{1}{|c|}{\cellcolor[HTML]{E6B8AF}} &
  \textbf{P/R1} &
  \multicolumn{1}{c|}{\cellcolor[HTML]{EFEFEF}89.32} &
  \textit{39.06} &
  \multicolumn{1}{c|}{\cellcolor[HTML]{EFEFEF}88.2} &
  35.08 &
  \multicolumn{1}{c|}{\cellcolor[HTML]{EFEFEF}88.44} &
  36.49 &
  \multicolumn{1}{c|}{\cellcolor[HTML]{EFEFEF}90.67} &
  42.49 \\   
\rowcolor[HTML]{FFFFFF} 
\multicolumn{1}{|c|}{\cellcolor[HTML]{E6B8AF}} &
  \textbf{R/R2} &
  \multicolumn{1}{c|}{\cellcolor[HTML]{FFFFFF}90.43} &
  14.06 &
  \multicolumn{1}{c|}{\cellcolor[HTML]{FFFFFF}91.51} &
  13.46 &
  \multicolumn{1}{c|}{\cellcolor[HTML]{FFFFFF}90.39} &
  13.54 &
  \multicolumn{1}{c|}{\cellcolor[HTML]{FFFFFF}91.74} &
  17.75 \\   
\rowcolor[HTML]{EFEFEF} 
\multicolumn{1}{|c|}{\multirow{-3}{*}{\cellcolor[HTML]{E6B8AF}\textbf{Llama-2-7b-chat}}} &
  \textbf{F1/RL} &
  \multicolumn{1}{c|}{\cellcolor[HTML]{EFEFEF}89.85} &
  31.02 &
  \multicolumn{1}{c|}{\cellcolor[HTML]{EFEFEF}89.81} &
  27.01 &
  \multicolumn{1}{c|}{\cellcolor[HTML]{EFEFEF}89.39} &
  28.15 &
  \multicolumn{1}{c|}{\cellcolor[HTML]{EFEFEF}91.19} &
  34.19 \\ \hline
\rowcolor[HTML]{FFFFFF} 
\multicolumn{1}{|c|}{\cellcolor[HTML]{70AD47}} &
  \textbf{P/R1} &
  \multicolumn{1}{c|}{\cellcolor[HTML]{FFFFFF}89.25} &
  \textit{39.06} &
  \multicolumn{1}{c|}{\cellcolor[HTML]{FFFFFF}88.89} &
  \textit{\textbf{39.44}} &
  \multicolumn{1}{c|}{\cellcolor[HTML]{FFFFFF}88.72} &
  39.26 &
  \multicolumn{1}{c|}{\cellcolor[HTML]{FFFFFF}90.08} &
  42.47 \\   
\rowcolor[HTML]{EFEFEF} 
\multicolumn{1}{|c|}{\cellcolor[HTML]{70AD47}} &
  \textbf{R/R2} &
  \multicolumn{1}{c|}{\cellcolor[HTML]{EFEFEF}90.32} &
  14.12 &
  \multicolumn{1}{c|}{\cellcolor[HTML]{EFEFEF}91.31} &
  15.93 &
  \multicolumn{1}{c|}{\cellcolor[HTML]{EFEFEF}90.61} &
  15.54 &
  \multicolumn{1}{c|}{\cellcolor[HTML]{EFEFEF}91.25} &
  17.66 \\   
\rowcolor[HTML]{FFFFFF} 
\multicolumn{1}{|c|}{\multirow{-3}{*}{\cellcolor[HTML]{70AD47}\textbf{Llama-2-13b-chat}}} &
  \textbf{F1/RL} &
  \multicolumn{1}{c|}{\cellcolor[HTML]{FFFFFF}89.77} &
  31.12 &
  \multicolumn{1}{c|}{\cellcolor[HTML]{FFFFFF}90.06} &
  30.79 &
  \multicolumn{1}{c|}{\cellcolor[HTML]{FFFFFF}89.64} &
  30.54 &
  \multicolumn{1}{c|}{\cellcolor[HTML]{FFFFFF}90.64} &
  34.04 \\ \hline
\rowcolor[HTML]{EFEFEF} 
\multicolumn{1}{|c|}{\cellcolor[HTML]{FF9900}} &
  \textbf{P/R1} &
  \multicolumn{1}{c|}{\cellcolor[HTML]{EFEFEF}89.8} &
  39.32 &
  \multicolumn{1}{c|}{\cellcolor[HTML]{EFEFEF}88.08} &
  36.41 &
  \multicolumn{1}{c|}{\cellcolor[HTML]{EFEFEF}88.86} &
  37.86 &
  \multicolumn{1}{c|}{\cellcolor[HTML]{EFEFEF}90.77} &
  44.05 \\   
\rowcolor[HTML]{FFFFFF} 
\multicolumn{1}{|c|}{\cellcolor[HTML]{FF9900}} &
  \textbf{R/R2} &
  \multicolumn{1}{c|}{\cellcolor[HTML]{FFFFFF}90.76} &
  13.74 &
  \multicolumn{1}{c|}{\cellcolor[HTML]{FFFFFF}91.87} &
  14.68 &
  \multicolumn{1}{c|}{\cellcolor[HTML]{FFFFFF}91.23} &
  14.39 &
  \multicolumn{1}{c|}{\cellcolor[HTML]{FFFFFF}91.66} &
  18.59 \\   
\rowcolor[HTML]{EFEFEF} 
\multicolumn{1}{|c|}{\multirow{-3}{*}{\cellcolor[HTML]{FF9900}\textbf{Llama-2-70b-chat}}} &
  \textbf{F1/RL} &
  \multicolumn{1}{c|}{\cellcolor[HTML]{EFEFEF}90.26} &
  31.01 &
  \multicolumn{1}{c|}{\cellcolor[HTML]{EFEFEF}89.92} &
  28.24 &
  \multicolumn{1}{c|}{\cellcolor[HTML]{EFEFEF}90.01} &
  29.49 &
  \multicolumn{1}{c|}{\cellcolor[HTML]{EFEFEF}91.2} &
  35.72 \\ \hline
\rowcolor[HTML]{FFFFFF} 
\multicolumn{1}{|c|}{\cellcolor[HTML]{4472C4}} &
  \textbf{P/R1} &
  \multicolumn{1}{c|}{\cellcolor[HTML]{FFFFFF}\textbf{89.49}} &
  40.83 &
  \multicolumn{1}{c|}{\cellcolor[HTML]{FFFFFF}89.99} &
  42.31 &
  \multicolumn{1}{c|}{\cellcolor[HTML]{FFFFFF}90.47} &
  42.02 &
  \multicolumn{1}{c|}{\cellcolor[HTML]{FFFFFF}90.35} &
  44.81 \\   
\rowcolor[HTML]{EFEFEF} 
\multicolumn{1}{|c|}{\cellcolor[HTML]{4472C4}} &
  \textbf{R/R2} &
  \multicolumn{1}{c|}{\cellcolor[HTML]{EFEFEF}\textbf{91.57}} &
  16.49 &
  \multicolumn{1}{c|}{\cellcolor[HTML]{EFEFEF}91.54} &
  18.28 &
  \multicolumn{1}{c|}{\cellcolor[HTML]{EFEFEF}91.14} &
  17.62 &
  \multicolumn{1}{c|}{\cellcolor[HTML]{EFEFEF}91.77} &
  20.49 \\   

\multicolumn{1}{|c|}{\multirow{-3}{*}{\cellcolor[HTML]{4472C4}\textbf{\begin{tabular}[c]{@{}c@{}}Mistral-7B-\\ Instruct-v0.1\end{tabular}}}} &
  \textbf{F1/RL} &
  \multicolumn{1}{c|}{\cellcolor[HTML]{FFFFFF}\textbf{90.51}} &
  32.21 &
  \multicolumn{1}{c|}{\cellcolor[HTML]{FFFFFF}90.73} &
  34 &
  \multicolumn{1}{c|}{\cellcolor[HTML]{FFFFFF}90.79} &
  33.68 &
  \multicolumn{1}{c|}{\cellcolor[HTML]{FFFFFF}91.04} &
  36.02 \\ \hline

\multicolumn{1}{|c|}{\cellcolor[HTML]{EA9999}} &
  \textbf{P/R1} &
  \multicolumn{1}{c|}{\cellcolor[HTML]{EFEFEF}\textbf{90.39}} &
  \textbf{41.61} &
  \multicolumn{1}{c|}{\cellcolor[HTML]{EFEFEF}\textbf{90.39}} &
  \textbf{43.21} &
  \multicolumn{1}{c|}{\cellcolor[HTML]{EFEFEF}{\ul \textbf{91.03}}} &
  \textbf{44.37} &
  \multicolumn{1}{c|}{\cellcolor[HTML]{EFEFEF}\textbf{91.01}} &
  {\ul \textbf{45.86}} \\   
\rowcolor[HTML]{FFFFFF} 
\multicolumn{1}{|c|}{\cellcolor[HTML]{EA9999}} &
  \textbf{R/R2} &
  \multicolumn{1}{c|}{\cellcolor[HTML]{FFFFFF}\textbf{92.05}} &
  \textbf{16.21} &
  \multicolumn{1}{c|}{\cellcolor[HTML]{FFFFFF}\textbf{91.88}} &
  \textbf{18.45} &
  \multicolumn{1}{c|}{\cellcolor[HTML]{FFFFFF}\textbf{92.11}} &
  \textbf{20} &
  \multicolumn{1}{c|}{\cellcolor[HTML]{FFFFFF}{\ul \textbf{92.22}}} &
  {\ul \textbf{20.78}} \\   

\multicolumn{1}{|c|}{\multirow{-3}{*}{\cellcolor[HTML]{EA9999}\textbf{\begin{tabular}[c]{@{}c@{}}Mixtral-8x7B-\\ Instruct-v0.1\end{tabular}}}} &
  \textbf{F1/RL} &
  \multicolumn{1}{c|}{\cellcolor[HTML]{EFEFEF}\textbf{91.2}} &
  \textbf{32.55} &
  \multicolumn{1}{c|}{\cellcolor[HTML]{EFEFEF}\textbf{91.11}} &
  \textbf{34.55} &
  \multicolumn{1}{c|}{\cellcolor[HTML]{EFEFEF}\textbf{91.55}} &
  \textbf{35.76} &
  \multicolumn{1}{c|}{\cellcolor[HTML]{EFEFEF}{\ul \textit{\textbf{91.6}}}} &
  {\ul \textbf{36.96}} \\ \hline
\end{tabular}%
%}
\end{adjustbox}
\caption{Performance of LLMs on the SAMSum dataset with ICL.}
\label{tab:table11}
\end{table}

{\ul\textbf{ArXiv results with ICL:}} according to the results shown in Table \ref{tab:table12}  \textbf{Llama} models outperform other models in the conducted experiments in terms of Rouge and BERT metrics with different number of in-context demonstrations. The results show an improvement of the performance accompanied by increasing the number of demonstrations in most cases . As mentioned before the results on ArXiv dataset are conducted after trimming the input documents due to context length so this interprets the variations of the results from one model to another.

\begin{table}[]
\centering
\begin{adjustbox}{max width=\textwidth} 
\renewcommand{\arraystretch}{1.5}
%\resizebox{\columnwidth}{!}{%
\begin{tabular}{|cc|cc|cc|cc|cc|}
\hline
\multicolumn{2}{|c|}{Prompts} &
  \multicolumn{2}{c|}{\cellcolor[HTML]{C96969}\textbf{1-Shot}} &
  \multicolumn{2}{c|}{\cellcolor[HTML]{C96969}\textbf{3-Shots}} &
  \multicolumn{2}{c|}{\cellcolor[HTML]{C96969}\textbf{5-Shots}} &
  \multicolumn{2}{c|}{\cellcolor[HTML]{C96969}\textbf{7-Shots}} \\ \hline
\multicolumn{1}{|c|}{\textbf{Models}} &
  Metrics &
  \multicolumn{1}{c|}{\textbf{Bert}} &
  \textbf{Rouge} &
  \multicolumn{1}{c|}{\textbf{Bert}} &
  \textbf{Rouge} &
  \multicolumn{1}{c|}{\textbf{Bert}} &
  \textbf{Rouge} &
  \multicolumn{1}{c|}{\textbf{Bert}} &
  \textbf{Rouge} \\ \hline
\rowcolor[HTML]{FFFFFF} 
\multicolumn{1}{|c|}{\cellcolor[HTML]{C27BA0}} &
  \textbf{P/R1} &
  \multicolumn{1}{c|}{\cellcolor[HTML]{FFFFFF}81.52} &
  22.43 &
  \multicolumn{1}{c|}{\cellcolor[HTML]{FFFFFF}83.72} &
  30.1 &
  \multicolumn{1}{c|}{\cellcolor[HTML]{FFFFFF}85.4} &
  35.21 &
  \multicolumn{1}{c|}{\cellcolor[HTML]{FFFFFF}83.24} &
  29.58 \\  
\rowcolor[HTML]{EFEFEF} 
\multicolumn{1}{|c|}{\cellcolor[HTML]{C27BA0}} &
  \textbf{R/R2} &
  \multicolumn{1}{c|}{\cellcolor[HTML]{EFEFEF}78.52} &
  2.56 &
  \multicolumn{1}{c|}{\cellcolor[HTML]{EFEFEF}81.52} &
  \multicolumn{1}{r|}{\cellcolor[HTML]{EFEFEF}7.46} &
  \multicolumn{1}{c|}{\cellcolor[HTML]{EFEFEF}82.28} &
  10.65 &
  \multicolumn{1}{c|}{\cellcolor[HTML]{EFEFEF}81.2} &
  7.12 \\  
\rowcolor[HTML]{FFFFFF} 
\multicolumn{1}{|c|}{\multirow{-3}{*}{\cellcolor[HTML]{C27BA0}\textbf{gemma-7b-it}}} &
  \textbf{F1/RL} &
  \multicolumn{1}{c|}{\cellcolor[HTML]{FFFFFF}79.98} &
  13.51 &
  \multicolumn{1}{c|}{\cellcolor[HTML]{FFFFFF}82.59} &
  17.2 &
  \multicolumn{1}{c|}{\cellcolor[HTML]{FFFFFF}83.79} &
  20.59 &
  \multicolumn{1}{c|}{\cellcolor[HTML]{FFFFFF}82.18} &
  17.73 \\ \hline
\rowcolor[HTML]{EFEFEF} 
\multicolumn{1}{|c|}{\cellcolor[HTML]{E6B8AF}} &
  \textbf{P/R1} &
  \multicolumn{1}{c|}{\cellcolor[HTML]{EFEFEF}85.72} &
  40.82 &
  \multicolumn{1}{c|}{\cellcolor[HTML]{EFEFEF}86.68} &
  42.16 &
  \multicolumn{1}{c|}{\cellcolor[HTML]{EFEFEF}86.13} &
  {\ul \textbf{48.29}} &
  \multicolumn{1}{c|}{\cellcolor[HTML]{EFEFEF}86.88} &
  39.84 \\  
\rowcolor[HTML]{FFFFFF} 
\multicolumn{1}{|c|}{\cellcolor[HTML]{E6B8AF}} &
  \textbf{R/R2} &
  \multicolumn{1}{c|}{\cellcolor[HTML]{FFFFFF}82.08} &
  13.68 &
  \multicolumn{1}{c|}{\cellcolor[HTML]{FFFFFF}\textbf{84.22}} &
  14.99 &
  \multicolumn{1}{c|}{\cellcolor[HTML]{FFFFFF}\textbf{83.85}} &
  {\ul \textbf{17.81}} &
  \multicolumn{1}{c|}{\cellcolor[HTML]{FFFFFF}{\ul \textbf{83.97}}} &
  14.32 \\  
\rowcolor[HTML]{EFEFEF} 
\multicolumn{1}{|c|}{\multirow{-3}{*}{\cellcolor[HTML]{E6B8AF}\textbf{Llama-2-7b-chat}}} &
  \textbf{F1/RL} &
  \multicolumn{1}{c|}{\cellcolor[HTML]{EFEFEF}\textbf{83.84}} &
  22.61 &
  \multicolumn{1}{c|}{\cellcolor[HTML]{EFEFEF}\textbf{85.41}} &
  23.95 &
  \multicolumn{1}{c|}{\cellcolor[HTML]{EFEFEF}84.97} &
  {\ul \textbf{22.45}} &
  \multicolumn{1}{c|}{\cellcolor[HTML]{EFEFEF}85.38} &
  23.29 \\ \hline
\rowcolor[HTML]{FFFFFF} 
\multicolumn{1}{|c|}{\cellcolor[HTML]{70AD47}} &
  \textbf{P/R1} &
  \multicolumn{1}{c|}{\cellcolor[HTML]{FFFFFF}\textbf{85.83}} &
  40.7 &
  \multicolumn{1}{c|}{\cellcolor[HTML]{FFFFFF}\textbf{86.71}} &
  \textbf{42.09} &
  \multicolumn{1}{c|}{\cellcolor[HTML]{FFFFFF}\textbf{86.7}} &
  47.85 &
  \multicolumn{1}{c|}{\cellcolor[HTML]{FFFFFF}{\ul \textbf{87.02}}} &
  39.96 \\  
\rowcolor[HTML]{EFEFEF} 
\multicolumn{1}{|c|}{\cellcolor[HTML]{70AD47}} &
  \textbf{R/R2} &
  \multicolumn{1}{c|}{\cellcolor[HTML]{EFEFEF}81.76} &
  13.66 &
  \multicolumn{1}{c|}{\cellcolor[HTML]{EFEFEF}84.08} &
  \textbf{15.07} &
  \multicolumn{1}{c|}{\cellcolor[HTML]{EFEFEF}83.64} &
  19.42 &
  \multicolumn{1}{c|}{\cellcolor[HTML]{EFEFEF}83.87} &
  14.47 \\  
\rowcolor[HTML]{FFFFFF} 
\multicolumn{1}{|c|}{\multirow{-3}{*}{\cellcolor[HTML]{70AD47}\textbf{Llama-2-13b-chat}}} &
  \textbf{F1/RL} &
  \multicolumn{1}{c|}{\cellcolor[HTML]{FFFFFF}83.73} &
  22.68 &
  \multicolumn{1}{c|}{\cellcolor[HTML]{FFFFFF}85.35} &
  \textbf{24} &
  \multicolumn{1}{c|}{\cellcolor[HTML]{FFFFFF}\textbf{85.14}} &
  22.14 &
  \multicolumn{1}{c|}{\cellcolor[HTML]{FFFFFF}{\ul \textbf{85.39}}} &
  23.43 \\ \hline
\rowcolor[HTML]{EFEFEF} 
\multicolumn{1}{|c|}{\cellcolor[HTML]{FF9900}} &
  \textbf{P/R1} &
  \multicolumn{1}{c|}{\cellcolor[HTML]{EFEFEF}85.93} &
  \textbf{41.03} &
  \multicolumn{1}{c|}{\cellcolor[HTML]{EFEFEF}86.7} &
  41.67 &
  \multicolumn{1}{c|}{\cellcolor[HTML]{EFEFEF}85.73} &
  39.41 &
  \multicolumn{1}{c|}{\cellcolor[HTML]{EFEFEF}86.53} &
  \textbf{40.39} \\  
\rowcolor[HTML]{FFFFFF} 
\multicolumn{1}{|c|}{\cellcolor[HTML]{FF9900}} &
  \textbf{R/R2} &
  \multicolumn{1}{c|}{\cellcolor[HTML]{FFFFFF}81.86} &
  \textbf{13.67} &
  \multicolumn{1}{c|}{\cellcolor[HTML]{FFFFFF}82.81} &
  14.71 &
  \multicolumn{1}{c|}{\cellcolor[HTML]{FFFFFF}82.66} &
  15.73 &
  \multicolumn{1}{c|}{\cellcolor[HTML]{FFFFFF}83.48} &
  \textbf{14.07} \\  
\rowcolor[HTML]{EFEFEF} 
\multicolumn{1}{|c|}{\multirow{-3}{*}{\cellcolor[HTML]{FF9900}\textbf{Llama-2-70b-chat}}} &
  \textbf{F1/RL} &
  \multicolumn{1}{c|}{\cellcolor[HTML]{EFEFEF}83.83} &
  \textbf{22.8} &
  \multicolumn{1}{c|}{\cellcolor[HTML]{EFEFEF}84.69} &
  23.83 &
  \multicolumn{1}{c|}{\cellcolor[HTML]{EFEFEF}84.17} &
  18.58 &
  \multicolumn{1}{c|}{\cellcolor[HTML]{EFEFEF}84.96} &
  \textbf{23.01} \\ \hline
\rowcolor[HTML]{FFFFFF} 
\multicolumn{1}{|c|}{\cellcolor[HTML]{4472C4}} &
  \textbf{P/R1} &
  \multicolumn{1}{c|}{\cellcolor[HTML]{FFFFFF}\textbf{84.04}} &
  37.1 &
  \multicolumn{1}{c|}{\cellcolor[HTML]{FFFFFF}84.79} &
  37.49 &
  \multicolumn{1}{c|}{\cellcolor[HTML]{FFFFFF}85.59} &
  38.88 &
  \multicolumn{1}{c|}{\cellcolor[HTML]{FFFFFF}85.17} &
  36.95 \\  
\rowcolor[HTML]{EFEFEF} 
\multicolumn{1}{|c|}{\cellcolor[HTML]{4472C4}} &
  \textbf{R/R2} &
  \multicolumn{1}{c|}{\cellcolor[HTML]{EFEFEF}\textbf{82.14}} &
  12.82 &
  \multicolumn{1}{c|}{\cellcolor[HTML]{EFEFEF}82.83} &
  12.68 &
  \multicolumn{1}{c|}{\cellcolor[HTML]{EFEFEF}82.06} &
  12.94 &
  \multicolumn{1}{c|}{\cellcolor[HTML]{EFEFEF}81.62} &
  12.7 \\  

\multicolumn{1}{|c|}{\multirow{-3}{*}{\cellcolor[HTML]{4472C4}\textbf{\begin{tabular}[c]{@{}c@{}}Mistral-7B-\\ Instruct-v0.1\end{tabular}}}} &
  \textbf{F1/RL} &
  \multicolumn{1}{c|}{\cellcolor[HTML]{FFFFFF}\textbf{83.05}} &
  20.03 &
  \multicolumn{1}{c|}{\cellcolor[HTML]{FFFFFF}83.76} &
  20.55 &
  \multicolumn{1}{c|}{\cellcolor[HTML]{FFFFFF}83.77} &
  21.47 &
  \multicolumn{1}{c|}{\cellcolor[HTML]{FFFFFF}83.31} &
  20.6 \\ \hline
\rowcolor[HTML]{EFEFEF} 
\multicolumn{1}{|c|}{\cellcolor[HTML]{EA9999}} &
  \textbf{P/R1} &
  \multicolumn{1}{c|}{\cellcolor[HTML]{EFEFEF}84.49} &
  39.33 &
  \multicolumn{1}{c|}{\cellcolor[HTML]{EFEFEF}85.65} &
  40.42 &
  \multicolumn{1}{c|}{\cellcolor[HTML]{EFEFEF}86.62} &
  40.14 &
  \multicolumn{1}{c|}{\cellcolor[HTML]{EFEFEF}85.87} &
  39.54 \\  
\rowcolor[HTML]{FFFFFF} 
\multicolumn{1}{|c|}{\cellcolor[HTML]{EA9999}} &
  \textbf{R/R2} &
  \multicolumn{1}{c|}{\cellcolor[HTML]{FFFFFF}80.89} &
  13.05 &
  \multicolumn{1}{c|}{\cellcolor[HTML]{FFFFFF}83.57} &
  13.97 &
  \multicolumn{1}{c|}{\cellcolor[HTML]{FFFFFF}83.51} &
  13.94 &
  \multicolumn{1}{c|}{\cellcolor[HTML]{FFFFFF}84.38} &
  13.37 \\  

\multicolumn{1}{|c|}{\multirow{-3}{*}{\cellcolor[HTML]{EA9999}\textbf{\begin{tabular}[c]{@{}c@{}}Mixtral-8x7B-\\ Instruct-v0.1\end{tabular}}}} &
  \textbf{F1/RL} &
  \multicolumn{1}{c|}{\cellcolor[HTML]{EFEFEF}82.64} &
  21.79 &
  \multicolumn{1}{c|}{\cellcolor[HTML]{EFEFEF}84.58} &
  22.27 &
  \multicolumn{1}{c|}{\cellcolor[HTML]{EFEFEF}85.02} &
  22.92 &
  \multicolumn{1}{c|}{\cellcolor[HTML]{EFEFEF}\textit{85.09}} &
  21.83 \\ \hline
\end{tabular}%
\end{adjustbox}
%}
\caption{Performance of LLMs on the Arxiv dataset with ICL.}
\label{tab:table12}
\end{table}

Lastly, according to the results ICL enhanced the performance of LLMs in most cases compared to ZSL by providing crucial contextual examples that improve the models' understanding and generation capabilities. In our experiments, the significant improvement across all models underscores the importance of providing sufficient context examples to the models to learn the context of the input documents. The consistent improvements observed with increased in-context examples highlight the value of ICL in leveraging the strengths of LLMs. This makes them more effective for various natural language processing tasks including text summarization as we see in the previous results.

\subsection{Comparisons With State-of-the-art Models}
\label{subsec:comparison }

To situate LLMs within the broader landscape of text summarization, we compare their results with earlier state-of-the-art neural models across datasets Tables \ref{tab:CNN-NewsRoom} and \ref{tab:arxiv_Samsum}. We focus on the highest-performing LLMs in our study in each dataset and contrast them with earlier existing models.

On news summarization (CNN/DM, NEWSROOM) in Table \ref{tab:CNN-NewsRoom}, LLMs like Llama-2-70b-chat underperform  models such as PEGASUS but eliminate dataset-specific training costs, as seen in Gemma-7b-it’s NEWSROOM results.

In table \ref{tab:arxiv_Samsum} For dialogue summarization , Mixtral-8x7B trails specialized models like SICK, suggesting LLMs require further optimization for conversational coherence. On scientific documents , Llama-2-7b-chat achieves a high ROUGE-1 (49.74) but struggles with coherence (ROUGE-L: 33.98), reflecting sensitivity to prompts and context truncation.

These results highlight LLMs’ potential for multi-domain adaptability but underscore the need for tailored prompts and hybrid architectures to balance semantic flexibility with syntactic precision.

\begin{table}[ht]
\centering

\begin{tabular}{llccc}
\toprule
\textbf{Dataset} & \textbf{Method} & \textbf{R-1} & \textbf{R-2} & \textbf{R-L} \\
\midrule
\multirow{5}{*}{\textbf{CNN/DM}} 
 & Hie-BART\cite{bib33} & 44.35 & 21.37 & 41.05 \\
 & PEGASUS\cite{bib34} & 47.97 & 24.18 & 44.88 \\
 & Llama-2-70b-chat & 40.98 & 17.23 & 27.52 \\
\cmidrule(lr){1-5}

\multirow{2}{*}{\textbf{NEWSROOM}} 
 & PEGASUS\cite{bib34} & 45.07 & 33.39 & 41.28 \\
 & Gemma-7b-it & 25.38 & 8.52 & 19.96 \\
 
\bottomrule
\end{tabular}
\caption{Comparison of traditional summarization models and LLMs across CNN/DM and SAMSum datasets.}
\label{tab:CNN-NewsRoom}
\end{table}

\begin{table}[ht]
\centering

\begin{tabular}{llccc}
\toprule
\textbf{Dataset} & \textbf{Method} & \textbf{R-1} & \textbf{R-2} & \textbf{R-L} \\
\midrule

\multirow{4}{*}{\textbf{SAMSum}} 
 & SICK\cite{bib14} & 53.73 & 28.81 & 49.5 \\
 & PEGASUS\cite{bib34} & 54.37 & 29.88 & 45.89 \\
 & Mixtral-8x7B & 45.86 & 20.78 & 36.96 \\
 
\cmidrule(lr){1-5}
\multirow{4}{*}{\textbf{ArXiv}} 
 & PRIMERA\cite{bib9} & 47.60 & 20.80 & 42.60 \\
 & PEGASUS\cite{bib34} & 44.67 & 17.18 & 25.7 \\
 & Llama-2-7b-chat & 49.74 & 32.47 & 33.98 \\
\bottomrule
\end{tabular}
\caption{Comparison of traditional summarization models and LLMs on NEWSROOM and ArXiv datasets.}
\label{tab:arxiv_Samsum}
\end{table}

\subsection{Summarization with Chunking Strategy}

To evaluate the impact of proposed chunking strategy, we conducted experiments on the ArXiv dataset. Table \ref{tab:chunking_comparison} reports the ROUGE-1 and BERTScore (F-1) results before and after chunking. 

The results demonstrate that chunking improves the performance of certain models—most notably \textbf{Llama-2-70b-chat} and \textbf{Mixtral-8x7B-Instruct-v0.1}—in both lexical (ROUGE-1) and semantic (BERTScore) metrics. For others, such as \textbf{Llama-2-13b-chat} and \textbf{gemma-7b-it}, chunking did not yield consistent gains, suggesting that model architecture and input sensitivity may influence its effectiveness.

\begin{table}[ht]
\centering
\renewcommand{\arraystretch}{1.5}
\begin{tabular}{|c|cc|cc|}
\hline
\textbf{Model} & \multicolumn{2}{c|}{\textbf{ROUGE-1}} & \multicolumn{2}{c|}{\textbf{BERTScore(F-1)}} \\ \hline
               & Before Chunking & After Chunking & Before Chunking & After Chunking \\ \hline
gemma-7b-it               & 40.05 & 39.91 & 84.51 & 83.60 \\ \hline
Llama-2-7b-chat           & 49.74 & 50.30 & 83.92 & 84.06 \\ \hline
Llama-2-13b-chat          & 47.92 & 47.01 & 84.21 & 82.04 \\ \hline
Llama-2-70b-chat          & 35.48 & 38.12 & 84.26 & 86.19 \\ \hline
Mistral-7b-instruct-v0.1       & 36.64 & 36.90 & 85.61 & 85.80 \\ \hline
Mixtral-8x7B-Instruct-v0.1  & 37.43 & 40.02 & 84.73 & 87.16 \\ \hline
\end{tabular}
\caption{Comparison of summarization performance on the ArXiv dataset before and after implementing the chunking strategy using zero-shot learning, evaluated using ROUGE-1 and F-1 BERTScore metrics.}
\label{tab:chunking_comparison}
\end{table}
\section{Discussion}
\label{sec5}
\subsection{Different tasks with different prompts and In-context examples}

Upon analyzing the datasets utilized in our research, it becomes evident that they represent distinct tasks and perspectives. For instance, CNNDM and NewsRoom datasets focus on news summarization. Conversely, ArXiv dataset centers on summarizing scientific articles, and SAMSum represents dialogue summarization. 
Given this diversity across the datasets, varied prompts are employed for each dataset, even within the same dataset, to assess the outputs comprehensively. Our objective is to gain deeper insights into the most effective strategies for producing high-quality summaries that align with reference summaries and human evaluations.

\begin{figure}[h]
    \centering
    \includegraphics[width=1\textwidth]{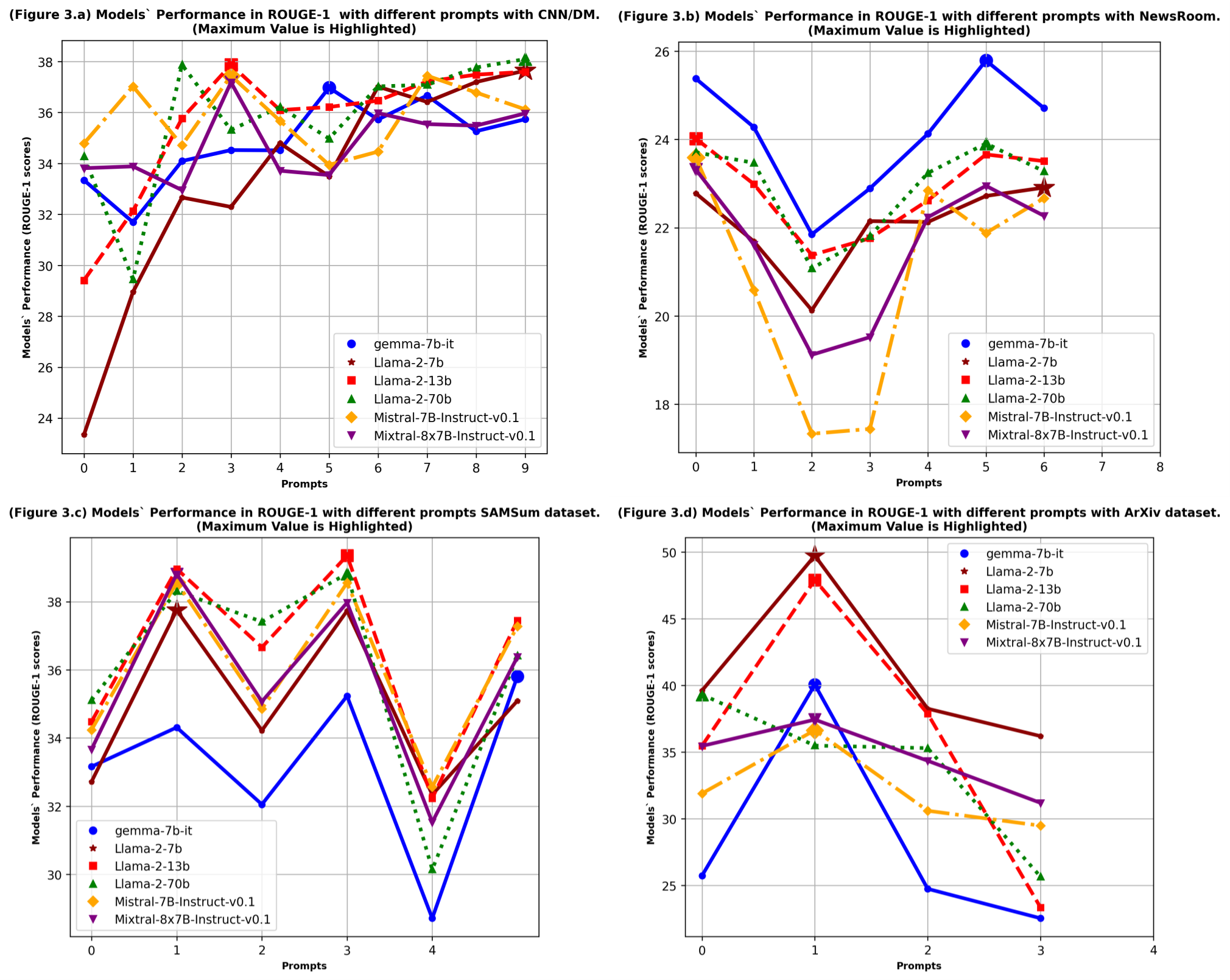}
    \caption{Change in Models' performance in ROUGE-1 across datasets using multiple prompts in zero-shot prompting.}
    \label{fig:fig3}
\end{figure}

Across the datasets used, Figure \ref{fig:fig3} shows how altering the prompts affects the LLMs performance in ZSL. Each subplot in the figure represents the performance of LLMs on a specific dataset, with several prompts serving as input motivators for generating summaries. Through careful analysis, noticeable variations are observed in model performance across different prompts within the same dataset. Certain prompts elicit stronger model responses, leading to improved performance metrics, while others may result in suboptimal outcomes. This highlights the importance of prompt selection in guiding LLMs' behavior.

Furthermore, our findings highlight the adaptability and versatility of LLMs in responding to varying prompts across different datasets. Prompt diversity enables LLMs to generalize across tasks and datasets, showcasing their capability to understand and execute tasks across multiple domains without explicit training. In addition, ICL is deployed to see how the models learn the context from the provided examples. Results show that there is an improvement in performance of models when working with more context examples compared to ZSL.

\subsection{LLMs in Summarizing long documents.}

The results displayed in Table \ref{tab:chunking_comparison} indicate that for some models, such as \textbf{Llama-2-7b-chat}, \textbf{Llama-2-70b-chat}, and \textbf{Mixtral-8x7B-Instruct-v0.1}, chunking yielded improved ROUGE-1 scores, suggesting better preservation of critical contextual details with reference summaries. Similarly, improvements in BERTScore were observed, confirming enhanced semantic alignment with reference summaries. However, for certain models like \textbf{Mistral-7b-Instruct-v0.1}, \textbf{gemma-7b-it}, and \textbf{Llama-2-13b-chat}, the gains were minimal or even slightly negative, indicating that the benefits of chunking can vary depending on model architecture and inherent handling of long contexts.

Our evaluation revealed that breaking long documents into manageable chunks helps retain contextual information. The process yields more precise intermediate summaries and enhances the final output. However, this multi-stage process results in increased computational overhead and inference time, particularly when many chunks are being processed. Furthermore, synthesizing these summaries could also lead to inconsistencies in tone and detail, which may impact the overall quality.

\subsection{Comparison of Generated Summary Lengths with Reference Summaries}

Another analysis is conducted regarding the results summaries generated from the employed models to get more insights about the results. Hence, the lengths of the generated summaries are compared to the lengths of the reference summaries from all the datasets used in our study. Table \ref{tab:table13} offers a summary of the average lengths of the summaries produced by each model when given different prompts together with the reference summary lengths for every dataset. The average length is calculated by taking the average lengths of the resulting summaries across all different used prompts. The average lengths of the generated summaries show that the LLMs do not consistently generate summaries that are comparable in length to the reference summaries.
The results show that most models generate summaries that are longer when compared to reference summaries.

\begin{table}[h]
\centering

\renewcommand{\arraystretch}{1.75}
%\resizebox{\columnwidth}{!}{%
\begin{tabular}{|c|c|c|c|c|}
\hline
     & \multicolumn{4}{|c|}{\textbf{Datasets}}  \\
     \hline
\textbf{Models} & 
\textbf{CNN/DM} & 
\textbf{NewsRoom} & 
\textbf{SAMSum} & 
\textbf{ArXiv} \\
\hline
\textbf{gemma-7b-it}& 
\textbf{53.584} & 
\textbf{27.672} & 
44.722 & 
143.31 \\
\hline
\textbf{Llama-2-7b-chat} & 
83.456 & 
56.8  & 
47.534  & 
289.3575       \\
\hline
\textbf{Llama-2-13b-chat} & 
65.706  & 
47.67 & 
\textbf{39.308}  & 
\textbf{208.825}\\
\hline
\textbf{Llama-2-70b-chat}  & 
71.864 & 
44.832 & 
40.45 & 
200.4975   \\
\hline
\textbf{Mistral-7B-Instruct-v0.1}   & 
108.056         & 
60.46             & 
48.098          & 
168.57         \\
\hline
\textbf{Mixtral-8x7B-Instruct-v0.1} & 
83.704          & 
45.07             & 
46.528          & 
204.71         \\
\hline
\textbf{Reference Length (\#Words) }         & 
\textbf{52}              & 
\textbf{26}                & 
\textbf{22}              & 
\textbf{220}   \\
\hline
\end{tabular}%
%}
\caption{Average generated summary lengths of LLMs compared to reference summary lengths.}
\label{tab:table13}
\end{table}

\begin{figure}[H]
    \centering
    \includegraphics[width=1\textwidth]{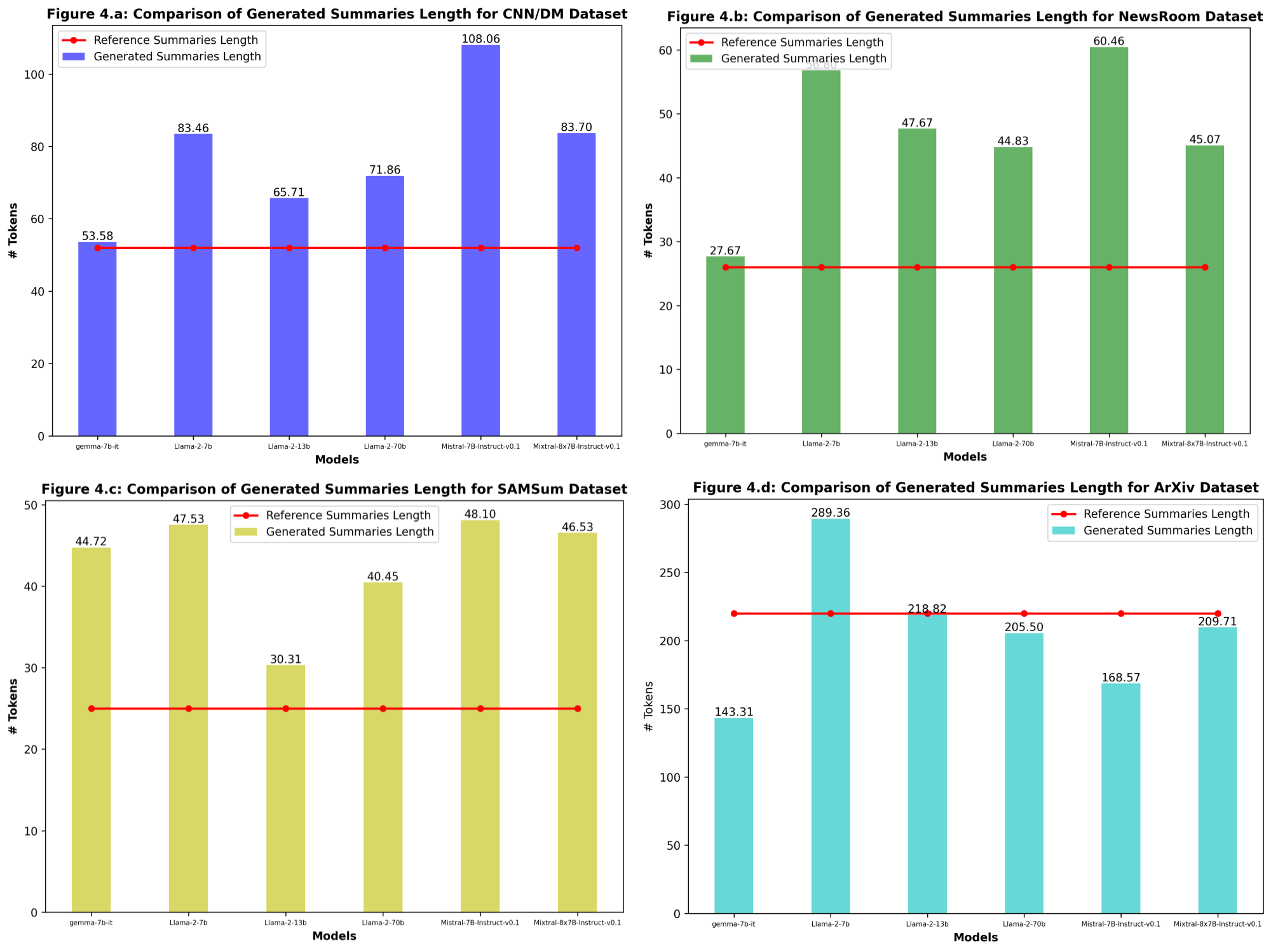}
    \caption{Comparison of Generated Summaries Length Across Different Models and Datasets using ZSL.}
    \label{fig:fig4}
\end{figure}

According to the results, some observations can be depicted. The variation in summary lengths across different models suggests that the inherent complexity and content structure of each dataset affect how models generate summaries. For instance, the ArXiv dataset, with its longer reference summaries, sees significant variation in generated summary lengths, reflecting the models' different strategies in handling detailed scientific and long texts. Another observation is that LLMs occasionally generate detailed responses to be more meaningful unless the input prompt constrains the model to generate fixed length output. As a result of that, most of the generated summaries have length longer than the reference summaries. Figure \ref{fig:fig4} visually provides a detailed comparison of the generated summary lengths according with the reference summary length of each dataset.

Overall, these findings collectively imply that the strengths and weaknesses of various models vary according to the dataset characteristics, in addition to model's length, number of parameters and prompt building as described previously. For instance, Llama-2-13b-chat excels in producing summaries that are nearly the reference length for datasets like SAMSum and ArXiv whereas gemma-7b-it consistently generates succinct summaries for the majority of datasets. 

However \textbf{Mistral-7B-Instruct-v0.1} shows a tendency to produce longer summaries which may risk verbosity in an attempt to capture more detail. Consequently the characteristics of the particular dataset and the intended ratio of detail to conciseness should be taken into account when choosing a model for a summarization task. In conclusion there is a trade-off between verbosity and the efficiency of the generated summaries when using LLMs for ATS and this trade-off affects how well these summaries are evaluated in comparison to reference summaries.

\subsection{Variation in LLM Performance: Insights and Contributing Factors}

 The complexity of the ATS tasks is illustrated by the variations in performance observed between the LLMs when running on various datasets. Significant variations in performance metrics were observed according to the results. Larger-parameter LLMs (e.g., \textbf{Llama-2-70b-chat} and \textbf{Mistral-8x7B-Instruct-v0.1}) outperform smaller ones on certain datasets (e.g., CNN/Daily Mail and ArXiv), suggesting that model capacity plays a crucial role in generating accurate summaries. However, this trend does not apply to all metrics and datasets, which suggests that the observed variations cannot be fully explained by the models' parameters alone \cite{bib47, bib76}.

Another factor influencing the performance of LLMs is the prompt design. The words included in prompts and prompt's organization have a significant influence on how well-written the summaries turn out \cite{bib60}. For instance, brief summary-specific prompts tend to elicit more accurate and relevant summaries than open-ended questions. This suggests that LLMs performance can be enhanced by employing particular prompts created for the summarization task.

ICL also affects the model’s results as well. As demonstrated by the results, adding examples from one to seven to the prompts generally improved the performance metrics in most modes, underscoring the significance of giving contextual examples to help the LLMs create summaries \cite{bib46}.

Despite these insights, we conclude that no single factor can be definitively identified as the sole determinant of variation in performance. Many combinations produce different outcomes due to the intricate relationship between prompt design, prompt size, prompting techniques and models' parameters. Thus, performance variability can be attributed to various factors, and it is not possible to say for sure that a single strategy will always yield better results. Future research should continue to explore these dimensions to better understand and optimize LLMs for ATS.

\subsection{Discussion of Metric Usability}

In this study, we evaluate the performance of the output summaries using two metrics: ROUGE and BERTScore. These metrics were selected due to their ability to measure both surface-level overlap (ROUGE) and semantic similarity (BERTScore) \cite{bib2,bib39}. We explore the strengths and limitations of these metrics based on our experience in this study, as described below.

ROUGE is a simple and effective metric for measuring n-gram overlap in summaries, making it a long-standing gold standard \cite{bib2}. BERTScore, on the other hand, leverages contextual embeddings to capture deeper semantic relationships \cite{bib39}. Together, these metrics offer a balanced assessment of both lexical and semantic quality.

Both ROUGE and BERTScore have inherent limitations that must be acknowledged. ROUGE tends to favor shorter summaries and may penalize models that generate longer but equally valid summaries. Additionally, its reliance on exact word matches makes it less effective for capturing paraphrased content or synonyms. BERTScore, while more semantically aware, can still struggle with out-of-domain data if the pre-trained model lacks exposure to relevant contexts. \cite{bib79,bib80}

During our experiments, we observed that interpreting metric scores required careful consideration of dataset characteristics and task requirements. For example, achieving high ROUGE scores on SAMSum (dialogue summarization) often meant generating overly verbose summaries that sacrificed conciseness. Similarly, optimizing for BERTScore on ArXiv sometimes led to summaries that were technically accurate but lacked clarity or readability. These challenges underscore the importance of balancing quantitative metrics with qualitative assessments when evaluating summarization systems.

Given the advancements in LLMs and NLP, future research should prioritize on evaluation frameworks that combine ROUGE's lexical precision with BERTScore's semantic nuance. Such metrics are essential for assessing LLM-generated summaries, as they need to balance syntactic fidelity and semantic abstraction to account for the generative nature of these models.

\subsection{Qualitative Analysis of Generated Summaries}
 
The examination of the summaries generated by the top-performing models on various datasets allowed us to perform qualitative analysis. For the utilized datasets Table \ref{tab:table16} displays the corresponding generated summaries along with random samples of the reference summaries. The key elements of LLM-generated summaries are shown in these examples:
\begin{itemize}
\item \textbf{Content Coverage and Informative Value:} Although the generated summaries may not fully align with the reference summaries at the lexical level, they often capture the core meaning effectively. For example in Table \ref{tab:table16} the produced summaries are paraphrased from the reference summaries but remain informative and contextually accurate.

\item \textbf{Coherence and Structure:} The summaries exhibit a well-structured narrative flow. Although some stylistic variations exist compared to the references. This suggests that LLMs can generate coherent and contextually appropriate summaries even when lexical differences arise.

\item \textbf{Error Patterns:} Several significant issues with the generated summaries are revealed by our error analysis. For example, the NewsRoom example clearly demonstrates how important details can be misrepresented or left out, leading to factual inconsistencies. Furthermore, although core meanings are typically maintained, overgeneralization and paraphrasing variability can occasionally lead to a lack of specificity and clarity, as demonstrated in the CNNDM example (e.g., "more than 200 people" becoming "a large crowd"). Furthermore, these examples, along with other generated examples from different models and datasets, show recurrent error patterns that compromise the accuracy and readability of the summaries.

\end{itemize}

Overall, our analysis suggests that while LLM-generated summaries may not always achieve exact lexical matching with reference summaries, they frequently preserve the intended meaning and provide concise, informative content. This highlights the potential of LLMs to generate summaries that are both contextually relevant and useful, even when their wording diverges from human-written references.

\subsection{Inference Time Analysis}

In addition to assessing summarization quality, we measured the average inference time required for each model to generate summaries across all datasets. Inference time was recorded by capturing the time immediately before and after the generation call, ensuring consistency across experiments. Table \ref{tab:inference_times} summarizes the average inference times.

\begin{table}[ht]
\centering

\begin{tabular}{|c|c|c|c|c|}
\hline
\textbf{Model} & \textbf{CNN/DM} & \textbf{NewsRoom} & \textbf{ArXiv} & \textbf{SAMSum} \\ \hline
Llama2-7b-chat & 23.34 & 25.1 & 61.2 & 11.5 \\ \hline
Llama2-13b-chat & 25.7 & 30.3 & 68.8 & 14.9 \\ \hline
Llama2-70b-chat & 80.6 & 95.2 & 148.4 & 30.7 \\ \hline
Gemma-7b-it & 21.1 & 23.8 & 63.4 & 10.2 \\ \hline
Mistral-7b-instruct & 17.9 & 20.5 & 52.9 & 8.3 \\ \hline
Mixtral-8$\times$7b-instruct-v0.1 & 31.5 & 36.8 & 100.3 & 15.4 \\ \hline
\end{tabular}
\caption{Average inference time (seconds) per article across models and datasets.}
\label{tab:inference_times}
\end{table}

\begin{figure}[H]
    \centering
    \includegraphics[width=1\textwidth]{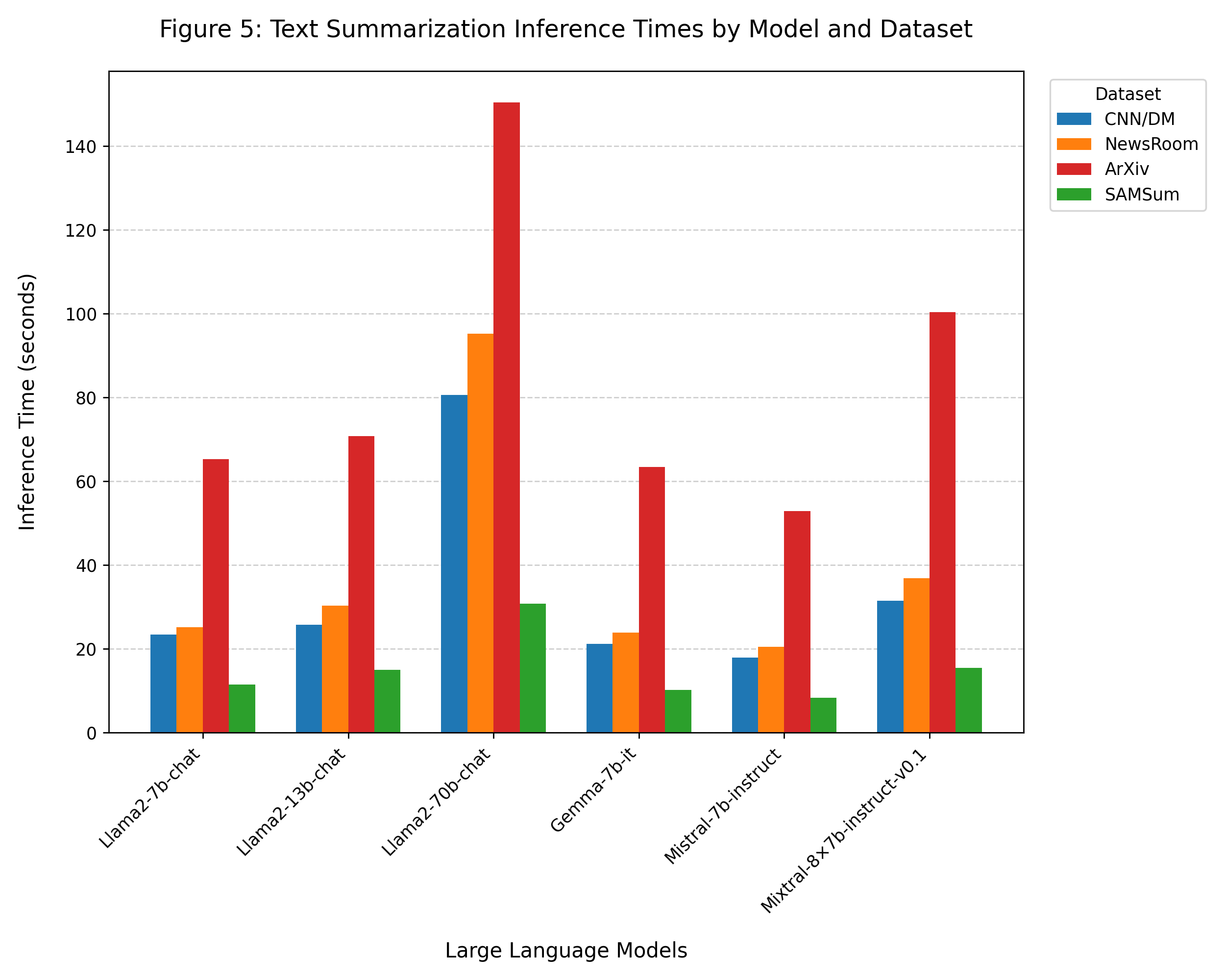}
    \caption{Text Summarization Inference Times by Model and Dataset.}
    \label{fig:fig5}
\end{figure}

Our analysis of the inference time revealed three important patterns. First, it's notable that larger models show noticeable slowdowns, as illustrated in Figure \ref{fig:fig5}. The increased inference time can be attributed to the larger model size, which inherently increases the computational latency. For example, Llama2-70b-chat takes longer to summarize each article than Llama2-7b-chat and Llama2-13b-chat on CNN/DM. Second, runtime variability is dominated by input length: short dialogs in SAMSum require less time than summarizing ArXiv papers because long-context encoding disproportionately increases memory bandwidth pressure. Third, these costs are reduced by architectural optimizations. For instance, even with higher parameter counts, the Mixtral-8x7Bs sparse Mixture-of-Experts (MoE) design performed better on ArXiv than Llama2-70B.

These results reveal the trade-off between summarization quality and computational efficiency. Some models produce higher-quality summaries but require significantly longer inference times, which is a critical factor in real-world applications. Moreover, variations in model architecture, such as parameter size and complexity, further influence these trade-offs.

\section{Conclusion}
\label{sec6}

Transforming vast amounts of unstructured text into concise, coherent summaries is a practical and challenging task. Which is critical for enabling streamlined knowledge transfer across disciplines. Considering these challenges, the advent of LLMs has created unprecedented opportunities in NLP to effectively summarize diverse textual content. In this study, we comprehensively evaluated locally hosted, open-source LLMs for text summarization across news, dialog, and scientific domains. Although LLMs offer the advantage of Zero-Shot Learning (ZSL) or Incontext Learning (ICL) in the summarization domain without task-specific training, our results demonstrate considerable performance variations across different domains. On news datasets (CNN/DM, NEWSROOM), while models such as PEGASUS achieve slightly higher scores, LLMs exhibit moderate semantic performance and provide a unified solution that can be used in a variety of domains without extensive training.In dialog summarization (SAMSum), LLMs such as Mixtral-8x7B deliver robust performance with minimal adjustments, although there is still potential to enhance syntactic precision compared to previous models such as SICK. For scientific texts, applying a chunking strategy helped mitigate context limitations, leading to more coherent and informative summaries, particularly for models like Llama-2-7b-chat.

The results of this study showed considerable variation in the performance of LLMs across different domains and models. The effects are being reflected of model architecture, scale, prompting strategy, prompt design, and training data diversity. 

% This variation suggests that no one LLM excels at all tasks, thereby requiring a deliberate choice based on the domain specifics and the available resources.

\section*{Statements and Declarations}

\subsection*{Competing interests}
 There is no competing interest.
 %======================================================
\section*{Statements and Declarations}

\subsection*{Competing interests}
 The authors have no competing interests to declare that are relevant to the content of this article.
 
 \subsection*{Funding}
No funding was received for conducting this study.

\subsection*{Availability of data and materials}
All used datasets, the generated summaries and results are available here \url{https://github.com/walid798/TextSummarizationCode}

%%===========================================================================================%%
%\printbibliography
%\bibliographystyle{sn-basic}
% \bibliographystyle{plainnat}
\bibliography{sn-bibliography}% common bib file

% \bibliography{sn-bibliography}
\appendix

\section{Qualitative Examples of Generated Summaries}
\begin{table}[htb]
\centering
\renewcommand{\arraystretch}{2.5}
\begin{adjustbox}{max width=\textwidth} 
\begin{tabular}{|>{\raggedright\arraybackslash}p{2.5cm}|>{\raggedright\arraybackslash}p{2.5cm}|>{\raggedright\arraybackslash}p{4.5cm}|>{\raggedright\arraybackslash}p{4.5cm}|}
\hline
\textbf{Dataset} & \textbf{Model} & \textbf{Reference Summary} & \textbf{Generated Summary} \\
\hline
\multicolumn{1}{|c|}{\textbf{CNNDM}} & 
\textbf{Llama-2-70B-chat} & 
More than 200 people queued up to get their hands on Aldi's spot prizes. The company issued golden tickets to the first 100 customers through doors. It was to celebrate the launch of a new store in Avlaston, Derbyshire. 
& 
The new Aldi store in Avlaston, Derbyshire, opened today and attracted a large crowd of eager customers who queued for hours to get their hands on prizes such as 21-inch TVs. The company gave away golden tickets to the first 100 people through the door, and the store is expected to create 29 jobs and support the local community. \\
\hline
\multicolumn{1}{|c|}{\textbf{NewsRoom}} & 
\textbf{Mixtral-8x7B-Instruct-v0.1} & 
Briefing note argues for 2018 further injection of democracy 2019 so the public or parliament are able to vote on the terms of Brexit. 
& 
The Labour party has not ruled out the possibility of another referendum on the European Union, but there is uncertainty over the timing and conditions. \\
\hline
\multicolumn{1}{|c|}{\textbf{SAMSum}} & 
\textbf{Mixtral-8x7B-Instruct-v0.1} & 
Amy gives Everett permission to give her number to Ralph. 
& 
Everett asked Amy if she would give her phone number to Ralph, and Amy agreed. \\
\hline
\multicolumn{1}{|c|}{\textbf{ArXiv}} & 
\textbf{Llama-2-7b} & 
accurate molecular crystal structure prediction is a fundamental goal in academic and industrial condensed matter research and polymorphism is arguably the biggest obstacle on the way. we tackle this challenge in the difficult case of the repeatedly studied , abundantly used aminoacid glycine that hosts still little - known phase transitions and we illustrate the current state of the field through this example .  we demonstrate that the combination of recent progress in structure search algorithms with the latest advances in the description of van der waals interactions in density functional theory , supported by data - mining analysis , enables a leap in predictive power : we resolve , without prior empirical input , all known phases of glycine , as well as the structure of the previously unresolved  phase after a decade of its experimental observation .  the search for the well - established  phase instead reveals the remaining challenges in exploring a polymorphic landscape . 
& 
Molecular polymorphism, the observation of different crystal structures composed of the same molecules, has been a significant challenge in crystal structure prediction (CSP). The development of accurate and reliable CSP methods is crucial for the discovery of new materials in solid-state chemistry, materials science, and pharmaceutical science. Recent advancements in density functional theory (DFT) have enabled the description of van der Waals (vdW) interactions, a critical component in CSP. This study combines state-of-the-art DFT with evolutionary CSP methods to perform a fully ab initio CSP search on glycine crystals without semi-empirical corrections. The results demonstrate the ability of ab initio CSP to predict the correct stability ordering and accurate pressure evolution of glycine polymorphs, passing the challenging blind test of exploring the phase space of polymorphic glycine. \\
\hline
\end{tabular}
\end{adjustbox}
\caption{Random samples of reference summaries from the datasets and generated summaries produced by different LLMs.}
\label{tab:table16}
\end{table}
\end{document}